\pdfoutput=1

\documentclass[11pt]{article}

\usepackage[preprint]{acl}

\usepackage{times}
\usepackage{latexsym}
\usepackage{setspace}

\usepackage[T1]{fontenc}

\usepackage[utf8]{inputenc}

\usepackage{microtype}
\usepackage{enumitem}
\setlist[itemize]{leftmargin=*}

\usepackage{inconsolata}

\usepackage{graphicx}
\usepackage{booktabs}
\usepackage{tabularx}
\usepackage{colortbl}
\usepackage{multirow}
\usepackage{amsmath,amssymb}
\usepackage{makecell}

\usepackage{array}
\newcolumntype{H}{>{\setbox0=\hbox\bgroup}c<{\egroup}@{}}
\usepackage{rotating}
\usepackage{pifont}
\usepackage[dvipsnames,svgnames]{xcolor}

\newcommand{\red}{\textcolor{red}}
\newcommand{\green}{\textcolor{DarkGreen}}

\definecolor{egreen}{RGB}{180, 255, 160}
\definecolor{ered}{RGB}{190, 240, 255}

\newcommand{\NAPData}{NAP}
\newcommand{\UnbiasedData}{PunnyPattern}
\newcommand{\SubstitutionData}{PunBreak}

\usepackage{tcolorbox}
\tcbuselibrary{skins}
\usepackage{float}
\usepackage{caption}
\usepackage{pdfpages}
\newfloat{qaexample}{htbp}{lop}
\floatname{qaexample}{Example}

\newtcolorbox{colorboxwithtitle}[2][]{
  colback=#1!10,
  colframe=#1,
  title=#2,
  fonttitle=\bfseries,
  coltitle=black,
  boxrule=0.8pt,
  arc=2mm,
  top=1mm,
  bottom=1mm,
  left=1mm,
  right=1mm,
}

\newtcolorbox{mergedcolorboxtop}[2][]{
  enhanced, 
  colback=#1!10,
  colframe=#1, 
  title=#2,
  fonttitle=\bfseries,
  coltitle=black,
  boxrule=0.8pt, 
  arc=2mm,       
  top=0.4mm,       
  bottom=0.4mm,    
  left=1mm,
  right=1mm,
  sharp corners=south, 
  bottomrule=0pt,    
  before skip=0pt,   
  after skip=0pt,    
}

\newtcolorbox{mergedcolorboxmid}[2][]{
  enhanced, 
  colback=#1!10,
  colframe=#1, 
  title=#2,
  fonttitle=\bfseries,
  coltitle=black,
  boxrule=0.8pt, 
  arc=2mm,       
  top=0.4mm,       
  bottom=0.4mm,    
  left=1mm,
  right=1mm,
  sharp corners=north,
  sharp corners=south,
  bottomrule=0pt,
  toprule=0pt,     
  before skip=0pt,   
  after skip=0pt,    
}

\newtcolorbox{mergedcolorboxbottom}[2][]{
  enhanced, 
  colback=#1!10,
  colframe=#1, 
  title=#2,
  fonttitle=\bfseries,
  coltitle=black,
  boxrule=0.8pt, 
  arc=2mm,       
  top=0.4mm,       
  bottom=0.4mm,    
  left=1mm,
  right=1mm,
  sharp corners=north, 
  toprule=0pt,     
  before skip=0pt,   
  after skip=0pt,    
}

\newtcolorbox{prompt}[1][]{ 
    colframe=black,            
    colback=white!0,          
    colbacktitle=gray!40,     
    arc=2mm,                  
    boxrule=1pt,              
    fonttitle=\bfseries\itshape\color{black},      
    center title,             
    enhanced,                 
    lower separated=false,
    boxsep=2pt,               
    #1                        
}
\newtcolorbox{promptop}[1][]{ 
    colframe=black,            
    colback=white!0,          
    colbacktitle=gray!40,     
    arc=2mm,                  
    boxrule=1pt,              
    fonttitle=\bfseries\itshape\color{black},      
    center title,             
    enhanced,                 
    lower separated=false, 
    sharp corners=south,
    after skip=0pt,
    boxsep=2pt,               
    #1                        
}
\newtcolorbox{promptbot}[1][]{ 
    colframe=black,            
    colback=white!0,          
    colbacktitle=gray!40,     
    arc=2mm,                  
    boxrule=1pt,              
    fonttitle=\bfseries\itshape\color{black},      
    center title,             
    enhanced,                 
    lower separated=false, 
    sharp corners=north, 
    toprule=0pt,     
    before skip=0pt,   
    boxsep=2pt,               
    #1                        
}

\graphicspath{{images/}}

\title{Pun Unintended: LLMs and the Illusion of Humor Understanding}

\author{
  Alessandro Zangari $^{\spadesuit}$ \quad
  \textbf{Matteo Marcuzzo} $^{\spadesuit}$ \quad
  \textbf{Andrea Albarelli} $^{\spadesuit}$ \\
  \textbf{Mohammad Taher Pilehvar} $^{\diamondsuit}$ \quad
  \textbf{Jose Camacho-Collados} $^{\diamondsuit}$ \\
  $^{\spadesuit}$Dept of Environmental Sciences, Informatics and Statistics, Ca' Foscari University of Venice \\
  \normalsize{\texttt{alessandro.zangari@unive.it, matteo.marcuzzo@unive.it, albarelli@unive.it}} \\
  $^{\diamondsuit}$School of Computer Science and Informatics, Cardiff University \\
  \normalsize{\texttt{pilehvarmt@cardiff.ac.uk, camachocolladosj@cardiff.ac.uk}}
}

\begin{document}
\maketitle
\begin{abstract}
Puns are a form of humorous wordplay that exploits polysemy and phonetic similarity. 
While LLMs have shown promise in detecting puns, we show in this paper that their understanding often remains shallow, lacking the nuanced grasp typical of human interpretation. 
By systematically analyzing and reformulating existing pun benchmarks, we demonstrate how subtle changes in puns are sufficient to mislead LLMs. 
Our contributions include comprehensive and nuanced pun detection benchmarks, human evaluation across recent LLMs, and an analysis of the robustness challenges these models face in processing puns.
\end{abstract}

\section{Introduction}\label{sec:intro}

Understanding linguistic nuances, such as hedges, idiomatic phrases, figures of speech, and metaphors, is crucial for effective communication \cite{boisson-etal-2024-metaphors,nuance_language}. This requires deep contextual and cultural awareness, presenting ongoing challenges for LLMs that struggle with subtle, multi-layered language \cite{liu-etal-2023-afraid,ghosh-srivastava-2022-epic,zhang-etal-2024-clamber}.
One notable example of nuanced language is the \emph{pun} (paronomasia), a form of wordplay generating rhetorical, often humorous, effects through polysemy and phonetic similarity.
Prominent in literature, poetry, and advertising \cite{miller-gurevych-2015-automatic}, puns depend on the intuitive recognition of dual meanings (literal vs. figurative), creating the characteristic \emph{pun effect} \cite{brown-pun,attardo_routledge_2017}.
{The ability to automatically recognize puns is relevant for digital humanities and NLP tasks, such as sentiment analysis, and machine translation, which struggle with the ambiguity and non-literal nature of puns \cite{Miller_Turkovic_2016}. 
In comparison to other rhetorical devices, such as sarcasm, metaphors, or jokes, puns are structurally simpler \cite{punbook,miller-etal-2017-semeval}, and easier to formalize \cite{sun-etal-2022-context}. These qualities make them a good candidate for assessing linguistic understanding in LLMs.}

\begin{figure}[t!]
  \includegraphics[width=\columnwidth]{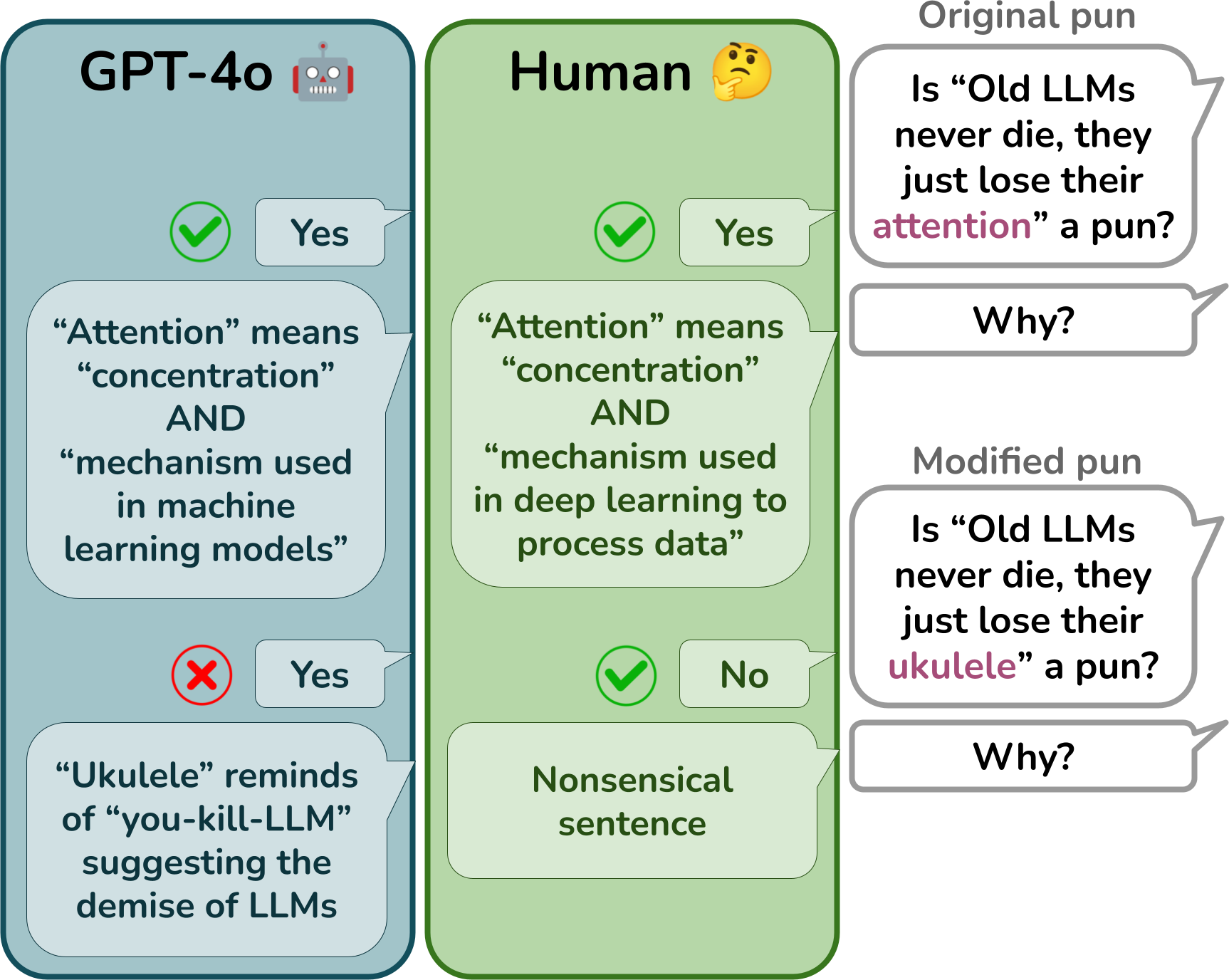}
  \caption{Example pun detection and explanation by GPT-4o vs. human. Subtle modifications to the pun (replacing the polysemous word, \textit{attention}, with a random one, \textit{ukulele}) are often sufficient to mislead LLMs.}
  \label{example:intro}
\end{figure}

While recent studies have examined pun generation \cite{mittal-etal-2022-ambipun,tian-etal-2022-unified}, detection, and explanation \cite{sun-etal-2022-expunations,xu-etal-2024-good,miller-etal-2017-semeval}, this line of research faces two key limitations. First, most studies have relied on a single dataset from SemEval \cite{miller-etal-2017-semeval}, where shallow cues and data leakage may inflate performance \cite{joker_corpus}. Second, evaluations are typically focused on binary classification or hard-to-assess free-text rationales. A systematic analysis of more structured rationales would help clarify the capabilities and limitations of LLMs in interpreting rhetorical ambiguity \cite{kim-2025-ambiguous,wiegreffe-etal-2021-measuring}.


Motivated by these limitations, we introduce two new collections of annotated short texts, \UnbiasedData\ and \SubstitutionData, designed specifically to evaluate the robustness of LLMs in pun detection (see Fig. \ref{example:intro}). These collections move beyond simple detection, using targeted substitutions and common language patterns to probe whether models can accurately recognize puns' context and structure, or if they rely on memorization and superficial cues. We publicly release both datasets\footnote{\url{https://github.com/alezanga/punintended}}.


We evaluate 7 open- and closed-weights LLMs on both identifying puns and providing supporting \emph{rationales}, then analyze the quality of these rationales, and report their impact on detection performance. 
In contrast to earlier studies \cite{sun-etal-2022-expunations,xu-etal-2024-good}, we design prompts that elicit semi-structured rationales, enabling more systematic evaluation via automatic metrics and detailed manual analysis guided by a new annotation protocol.
More specifically, we organize our discourse around the following research questions:
\begin{itemize}
\setlength\itemsep{0em}
\item RQ1: How well can LLMs detect puns on new and existing datasets?
\item RQ2: How robust are LLMs at detecting puns?
\item RQ3: To what extent can LLMs explain puns?
\end{itemize}

Our results show that most LLMs superficially associate puns with common language patterns, which makes it difficult for them to distinguish genuine puns from structurally similar sentences that contain no pun.
Automatic and manual error analyses further reveal that the models struggle to handle context and phonological properties effectively.

\section{Related Work}

SemEval-2017 Task 7 \cite{miller-etal-2017-semeval} targeted pun detection, location, and interpretation tasks by releasing a dataset of annotated puns and non-puns.
The participating systems were mostly based on traditional NLP techniques, with leading systems employing heuristics based on positional and semantic features, such as n-grams enhanced by Word2Vec embeddings \cite{mikolov2013efficient} and phonetic distance measurements. 
The reported results showed strong performance in binary detection, moderate in pun location, but notably weak in correct sense association.
SemEval-2021 \cite{meaney-etal-2021-semeval} later expanded this dataset by adding new jokes, although without additional annotations.


Building upon this foundation, the JOKER workshop \cite{joker_2023} introduced multilingual datasets for pun detection, location, interpretation, and translation, encompassing English, French, and Spanish.
The JOKER corpus \cite{joker_corpus} aimed to address earlier criticisms of imbalance and reliance on superficial cues by providing 
3,506 annotated short jokes in English and French, alongside 1,700 negative examples generated via word substitutions. 
Despite utilizing advanced LLMs such as GPT-3 \cite{gpt3} and BLOOM \cite{workshop2023bloom176bparameteropenaccessmultilingual},
a custom-trained T5 model \cite{RaffelT5} consistently outperformed others on all tasks, 
albeit with limited success (detection F1 < 0.6). 
Other studies have further explored non‑English puns \cite{porto_pun,chinese_2023,silva}.



Complementing these efforts, \citet{sun-etal-2022-expunations} investigated pun explanation by enhancing training data with rationales generated by T5 models.
They measured explanation quality using simulatability, defined as the accuracy difference when explanations are provided as additional input vs. when they are not. They found that high-quality human-annotated rationales can improve model performance, but automatically generating such explanations with language models remains challenging.
In a related work, \citet{sun-etal-2022-context} focus on pun generation by retrieving context and pun words. 
They propose a system that first retrieves a suitable pair of words, along with their senses, from given contextual words, and then employs a T5 model to generate a pun. However, the success rate of the generated puns was reported to be significantly lower than that achieved by humans.

To the best of our knowledge, \citet{xu-etal-2024-good} is the only study to evaluate recent LLMs on pun detection, explanation, and generation. They refined the SemEval dataset by removing duplicates and adding human-annotated explanations.
While the study included a manual evaluation of a small set of generated explanations, it lacks a comprehensive error analysis and detailed evaluation guidelines. Their results indicate that, although LLMs perform reasonably well on binary pun detection, they struggle with pun generation and explanation, often relying on memorization or biased shortcuts. Building on this work, our study conducts a robustness analysis targeting specific biases and examines common errors in model-generated rationales.




\section{Experimental Methodology}\label{sec:method}

The primary pun-related tasks previously discussed in the NLP literature relate to detection, location, and interpretation. 
\emph{Pun detection} is a binary classification task, categorizing texts as either containing a pun or not. \emph{Pun location} involves identifying the specific pair of words responsible for creating the double entendre, while \emph{pun interpretation} requires associating each identified pun word with its correct meaning or sense  \cite{joker_2023,jain_hira}. \\

\noindent \textbf{Preliminaries.} Structurally, a pun is composed of a \emph{pun word} ($w_p$), an \emph{alternative word} ($w_a$) and their respective \emph{senses} $s_p$ and $s_a$ \cite{sun-etal-2022-context}. For brevity, we will refer to $(w_p, w_a)$ as the \emph{pun pair}. Following previous literature \cite{miller-etal-2017-semeval,xu-etal-2024-good}, we focus on \emph{heterographic} and \emph{homographic} puns. 

In \emph{heterographic} puns (het-puns), only $w_p$ appears in the sentence, while $w_a$ and its sense are evoked through the context. For example, the het-pun ``\textit{I bought a boat because it was for \underline{sail} (\underline{sale})}'' creates a double meaning by playing on the homophones \textit{sail} (to navigate on a boat) and \textit{sale} (an occasion when items are sold).
In contrast, \emph{homographic} puns (hom-puns) feature a polysemous $w_p$, resulting in $w_p = w_a$. The hom-pun ``\textit{I was wondering why the ball was getting bigger; then it \underline{hit} (\underline{hit}) me}'' plays on the two senses of \textit{hit} (to be physically struck) and \textit{hit} (to suddenly realize).

The term \emph{rationale} has been used in the literature to denote various types of evidence justifying a decision, particularly in the form of natural language explanations \cite{herrewijnen-human-annotated-2024}. This is consistent with its usage in prior research on puns \cite{xu-etal-2024-good,sun-etal-2022-expunations}, although we do not limit it to unstructured or free-text explanations.

\subsection{Datasets}\label{sec:datasets}
Existing collections of puns annotated with the structure just described are limited. The SemEval 2017 dataset \cite{miller-etal-2017-semeval} provides 4,030 samples encompassing heterographic and homographic puns. Later research \cite{sun-etal-2022-expunations,xu-etal-2024-good} refined this dataset by adding sentiment annotations and removing incorrect examples. 
The English part of the JOKER corpus \cite{joker_corpus,joker_2023} 
extends SemEval and adds 632 new English puns collected from \texttt{PunOfTheDay.com}. By removing or replacing a single contextual word, these puns have been altered to create an equal number of non-puns.
Both datasets contain pun words and senses annotations. 


We utilize the dataset proposed by \citet{xu-etal-2024-good}, which is the most refined iteration of the SemEval 2017 dataset, along with a subset of the JOKER dataset. In the former, we corrected several typographical errors and 13 incorrect annotations, resulting in 2,589 samples. We split the dataset into training (1,071 samples), testing (1,341), and validation (177) sets, ensuring that no $w_p$ and $w_a$ in the training set appear in the test and validation sets. We refer to this as the \textit{PunEval} dataset.
Unlike the SemEval dataset, the full \textit{JOKER} corpus is not publicly available and was released only as part of the JOKER CLEF workshop \cite{joker_2023}. However, the authors kindly agreed to share the complete dataset with us upon request. Since most examples in this dataset originate from the SemEval dataset, we retained the subset of 632 puns and 632 non-puns created by replacing a single word in the original pun.
Statistics for all datasets are in Appendix §\ref{supp:data}.


\subsection{Comparison models}\label{sec:models}
We benchmark five recent instruction-tuned LLMs on pun understanding, including both open- and closed-weights models: GPT-4o-2024-08-06 (OpenAI) \cite{openai2024gpt4ocard}, Qwen2.5-72B (Alibaba) \cite{qwen2025qwen25technicalreport}, Llama3.3-70B (Meta) \cite{grattafiori2024llama3herdmodels}, Gemini2.0-Flash (Google DeepMind) \cite{geminiteam2025geminifamilyhighlycapable}, and Mistral3-24B (Mistral AI) \cite{jiang2023mistral7b}. We refer to these models as follows: GPT-4o, Qwen2.5, Llama3.3, Gemini2.0, and Mistral3.
Additionally, we employed two reasoning models from the DeepSeek family: DeepSeek-R1 (R1) and DeepSeek-R1-Distill-Llama-70B (R1-D) \cite{deepseekai2025deepseekr1incentivizingreasoningcapability}, the latter being a distilled version of the full R1, based on Llama-3.3-70B-Instruct. 

\subsection{Prompts}\label{sec:prompts}
Unlike in \citet{xu-etal-2024-good}, in this work we leverage the pun structure described in §\ref{sec:method} and instruct the model to respond in a semi-structured format, facilitating parsing and the evaluation of the rationales.


We adopt the following response format that is compatible with the rationale:
\texttt{(yes|no) [<$w_p$> <$w_a$> [<$s_p$> <$s_a$>]]}. In the configurations without rationales, the answer would simply be \textit{yes} or \textit{no} for the pun detection task.
The content within the angular brackets represents the rationale (or justification) for the answer. 
Inspired by the success of CoT prompting \cite{NEURIPS2022_zerocot,cot}, we explore an alternative prompting strategy to elicit reasoning from non-reasoning LLMs by asking them to \emph{think before answering}. Unlike previous cases where responses were restricted to a structured format, the reasoning prompts allow the model to provide free-text reasoning before delivering the final answer in the same structured format.
In what follows, we describe the prompt configurations used. Validation experiments with alternative prompts, along with the exact prompts we adopted, are detailed in Appendix §\ref{supp:prompts-valid}. \\

\noindent \textbf{Zero-Shot.} (\textsc{0s}) This prompt asks to provide a ``Yes'' or ``No'' answer on whether the text contains a pun, with no formal definition of a pun.

\noindent \textbf{Few-Shot.} (\textsc{fs}) A definition of pun is provided, including the concepts of $w_p$, $w_a$, $s_p$, and $s_a$, followed by six examples (three puns, three non-puns) drawn from the same set used by \citet{xu-etal-2024-good}. Examples are held constant across experiments.

\noindent \textbf{Words.} (\textsc{w}) Same as \emph{Few-Shot} prompt, but requires reporting the pun pair as justification.

\noindent \textbf{Words+Senses.} (\textsc{w+s}) This prompt mirrors the \emph{Words} prompt, but additionally requires reporting $s_p$ and $s_a$ after the pun pair.

\noindent \textbf{Reasoning.} (\textsc{R+}) This configuration uses the same prompts, but allows the generation of arbitrary text before the final answer. This prompt is applied only to the five non-reasoning LLMs.


\section{RQ1: Can LLMs Detect Puns?}\label{sec:rq1}
The first research question investigates LLMs' overall performance in binary pun detection. We thus evaluate all comparison LLMs described in §\ref{sec:models} using the prompting strategies outlined in §\ref{sec:prompts}.


\subsection{Data}
In addition to the PunEval and JOKER datasets described in §\ref{sec:datasets}, we annotate a new collection of 128 puns from various sources, including websites\footnote{\url{https://parade.com/1024249/marynliles/funny-puns}}, personal recollections, and original creations. The main reason for creating this new dataset was to avoid any potential contamination from existing datasets. Each pun is rephrased into a non-pun, similar to the JOKER dataset, but without the requirement of replacing a single word, resulting in a total of 256 instances. We use Mistral3 for this rephrasing process, and all generated samples are manually inspected and modified as needed. We refer to this as the \textit{Newly Annotated Puns} (\NAPData) dataset. A sample pun and its corresponding non-pun are shown in Example \ref{example:datarq1}, while general \NAPData\ statistics are included in Appendix §\ref{supp:data}. 
Finally, we also annotate each pun with its pun pair ($w_p$, $w_a$) and respective senses ($s_p$, $s_a$), using short definitions from online dictionaries\footnote{\url{https://dictionary.cambridge.org/dictionary}, \url{https://www.merriam-webster.com}}. This process is consistent with the methods employed in the SemEval and PunEval datasets, and these annotations will be used to address RQ3.

\begin{qaexample}[t!]
\small
\begin{mergedcolorboxtop}[egreen]{Original Pun}
What did the buffalo say to his son? \underline{Bison} (\emph{Bye son}). 
\end{mergedcolorboxtop}
\begin{mergedcolorboxbottom}[ered]{Rephrased pun}
What did the buffalo ask his son? \underline{I do not know}.
\end{mergedcolorboxbottom}
\caption{A pun and non-pun from the \NAPData\ dataset.}
\label{example:datarq1}
\end{qaexample}

\begin{table}[t!]
  \small
  \centering
  \begin{tabularx}{\columnwidth}{Xlccc}
    \toprule   
    \multirow{2}{*}{\textbf{M}} & \multirow{2}{*}{\textbf{Prompt}} & \multicolumn{3}{c}{\textbf{F1-score (\%)}} \\
    \cmidrule(lr){3-5}
     & & \textbf{\NAPData} & \textbf{JOKER} & \textbf{PunEval} \\
    \midrule
    
    \multirow{7}{*}{\begin{turn}{90}Gemini2.0\end{turn}}
    & \textsc{0s}  & 73.3 {\scriptsize $\pm$ 0.0} & 71.2 {\scriptsize $\pm$ 0.0} & 85.7 {\scriptsize $\pm$ 0.2} \\ 
    & \textsc{fs}  & 74.9 {\scriptsize $\pm$ 0.0} & 74.0 {\scriptsize $\pm$ 0.1} & 89.3 {\scriptsize $\pm$ 0.1} \\ 
    & \textsc{w}   & 76.2 {\scriptsize $\pm$ 0.4} & 74.4 {\scriptsize $\pm$ 0.1} & 88.3 {\scriptsize $\pm$ 0.1} \\ 
    & \textsc{w+s} & \textbf{77.9} {\scriptsize $\pm$ 0.3} & 74.4 {\scriptsize $\pm$ 0.0} & 89.1 {\scriptsize $\pm$ 0.2} \\ 
    & \textsc{R+fs}& 70.9 {\scriptsize $\pm$ 0.6} & 71.1 {\scriptsize $\pm$ 0.2} & 82.7 {\scriptsize $\pm$ 0.3} \\ 
    & \textsc{R+w} & 76.6 {\scriptsize $\pm$ 0.0} & 74.4 {\scriptsize $\pm$ 0.0} & \textbf{90.6} {\scriptsize $\pm$ 0.1} \\ 
    & \textsc{R+w+s}& 75.4 {\scriptsize $\pm$ 0.2} & \textbf{74.6} {\scriptsize $\pm$ 0.1} & 89.5 {\scriptsize $\pm$ 0.1} \\ 
    \midrule
    
    \multirow{7}{*}{\begin{turn}{90}GPT-4o\end{turn}} 
    & \textsc{0s}  & 78.4 {\scriptsize $\pm$ 0.1} & 77.2 {\scriptsize $\pm$ 0.0} & 89.7 {\scriptsize $\pm$ 0.1} \\ 
    & \textsc{fs}  & 81.0 {\scriptsize $\pm$ 0.7} & 79.7 {\scriptsize $\pm$ 0.1} & 92.8 {\scriptsize $\pm$ 0.2} \\ 
    & \textsc{w} & \textbf{86.9} {\scriptsize $\pm$ 0.8} & \textbf{83.3} {\scriptsize $\pm$ 0.4} & 87.3 {\scriptsize $\pm$ 0.9} \\ 
    & \textsc{w+s} & 85.0 {\scriptsize $\pm$ 0.8} & 82.6 {\scriptsize $\pm$ 0.1} & 88.9 {\scriptsize $\pm$ 0.9} \\ 
    & \textsc{R+fs} & 82.9 {\scriptsize $\pm$ 0.0} & 80.5 {\scriptsize $\pm$ 0.3} & 92.3 {\scriptsize $\pm$ 0.1} \\ 
    & \textsc{R+w} & 84.0 {\scriptsize $\pm$ 1.1} & 81.1 {\scriptsize $\pm$ 0.2} & 92.6 {\scriptsize $\pm$ 0.4} \\ 
    & \textsc{R+w+s}& 82.9 {\scriptsize $\pm$ 0.8} & 81.3 {\scriptsize $\pm$ 0.3} &\textbf{93.0} {\scriptsize $\pm$ 0.3} \\ 
    \midrule
    
    \multirow{7}{*}{\begin{turn}{90}\makecell{Llama3.3 \\ (70B)}\end{turn}}
    & \textsc{0s}& 79.6 {\scriptsize $\pm$ 1.5} & 77.1 {\scriptsize $\pm$ 0.8} & 84.0 {\scriptsize $\pm$ 0.4} \\ 
    & \textsc{fs}& 81.0 {\scriptsize $\pm$ 0.7} & 77.6 {\scriptsize $\pm$ 0.1} & 84.4 {\scriptsize $\pm$ 0.4} \\ 
    & \textsc{w}& 83.4 {\scriptsize $\pm$ 0.9} & \textbf{79.2} {\scriptsize $\pm$ 0.2} & \textbf{87.2} {\scriptsize $\pm$ 0.2} \\ 
    & \textsc{w+s}& \textbf{83.6} {\scriptsize $\pm$ 0.9} & 77.9 {\scriptsize $\pm$ 0.6} & 86.4 {\scriptsize $\pm$ 0.3} \\ 
    & \textsc{R+fs}& 79.2 {\scriptsize $\pm$ 0.2} & 75.2 {\scriptsize $\pm$ 0.1} & 83.8 {\scriptsize $\pm$ 0.2} \\ 
    & \textsc{R+w} & 78.1 {\scriptsize $\pm$ 0.8} & 76.4 {\scriptsize $\pm$ 0.1} & 85.1 {\scriptsize $\pm$ 0.5} \\ 
    & \textsc{R+w+s}& 78.3 {\scriptsize $\pm$ 1.0} & 76.8 {\scriptsize $\pm$ 0.2} & 84.5 {\scriptsize $\pm$ 0.3} \\ 
    \midrule
    
    \multirow{7}{*}{\begin{turn}{90}\makecell{Mistral3 \\ (24B)}\end{turn}}
    & \textsc{0s}& \textbf{75.0} {\scriptsize $\pm$ 0.5} & \textbf{75.3} {\scriptsize $\pm$ 0.2} & 80.8 {\scriptsize $\pm$ 0.2} \\ 
    & \textsc{fs}& 69.5 {\scriptsize $\pm$ 2.2} & 66.2 {\scriptsize $\pm$ 1.2} & 74.6 {\scriptsize $\pm$ 1.9} \\ 
    & \textsc{w}& 69.0 {\scriptsize $\pm$ 1.7} & 69.3 {\scriptsize $\pm$ 0.4} & 78.4 {\scriptsize $\pm$ 0.5} \\ 
    & \textsc{w+s}& 68.5 {\scriptsize $\pm$ 0.9} & 68.5 {\scriptsize $\pm$ 0.4} & 74.9 {\scriptsize $\pm$ 0.9} \\ 
    & \textsc{R+fs}& 73.6 {\scriptsize $\pm$ 1.1} & 70.9 {\scriptsize $\pm$ 0.1} & 82.4 {\scriptsize $\pm$ 1.3} \\ 
    & \textsc{R+w}& 70.7 {\scriptsize $\pm$ 0.3} & 69.0 {\scriptsize $\pm$ 0.4} & 84.2 {\scriptsize $\pm$ 0.4} \\ 
    & \textsc{R+w+s}& 70.9 {\scriptsize $\pm$ 0.2} & 70.1 {\scriptsize $\pm$ 0.3} & \textbf{84.5} {\scriptsize $\pm$ 0.2} \\ 
    \midrule
    
    \multirow{7}{*}{\begin{turn}{90}\makecell{Qwen2.5 \\ (72B)}\end{turn}}
    & \textsc{0s}& 71.3 {\scriptsize $\pm$ 0.4} & 73.6 {\scriptsize $\pm$ 0.2} & 83.6 {\scriptsize $\pm$ 0.1} \\ 
    & \textsc{fs}& 78.6 {\scriptsize $\pm$ 0.4} & 76.4 {\scriptsize $\pm$ 0.7} & 87.2 {\scriptsize $\pm$ 0.2} \\ 
    & \textsc{w}& \textbf{81.1} {\scriptsize $\pm$ 0.0} & 77.1 {\scriptsize $\pm$ 0.4} & 86.8 {\scriptsize $\pm$ 0.4} \\ 
    & \textsc{w+s}& 80.7 {\scriptsize $\pm$ 0.6} & 76.8 {\scriptsize $\pm$ 0.8} & 87.3 {\scriptsize $\pm$ 0.1} \\ 
    & \textsc{R+fs}& 80.2 {\scriptsize $\pm$ 0.7} & \textbf{77.5} {\scriptsize $\pm$ 0.6} & {88.5} {\scriptsize $\pm$ 0.6} \\ 
    & \textsc{R+w}& 77.2 {\scriptsize $\pm$ 0.2} & 75.0 {\scriptsize $\pm$ 0.1} & \textbf{89.4} {\scriptsize $\pm$ 0.5} \\ 
    & \textsc{R+w+s}& 77.0 {\scriptsize $\pm$ 1.1} & 75.8 {\scriptsize $\pm$ 0.6} & {88.7} {\scriptsize $\pm$ 0.1} \\ 
    \midrule
    
    \multirow{4}{*}{\begin{turn}{90}\makecell{R1-D \\ (70B)}\end{turn}}
     & \textsc{0s}& 74.8 {\scriptsize $\pm$ 0.5} & 75.3 {\scriptsize $\pm$ 0.6} & 86.7 {\scriptsize $\pm$ 0.3} \\ 
    & \textsc{fs}& 76.3 {\scriptsize $\pm$ 2.7} & 76.2 {\scriptsize $\pm$ 0.5} & 87.0 {\scriptsize $\pm$ 0.3} \\ 
    & \textsc{w}& \textbf{79.8} {\scriptsize $\pm$ 1.4} & 78.1 {\scriptsize $\pm$ 0.9} & 86.8 {\scriptsize $\pm$ 0.4} \\ 
    & \textsc{w+s}& 78.7 {\scriptsize $\pm$ 1.2} & \textbf{78.7} {\scriptsize $\pm$ 0.6} & \textbf{88.3} {\scriptsize $\pm$ 0.6} \\ 
    \midrule
    
    \multirow{4}{*}{\begin{turn}{90}\makecell{R1 \\ (671B)}\end{turn}}
     & \textsc{0s} & 74.2 {\scriptsize $\pm$ 0.4}  & 75.5 {\scriptsize $\pm$ 0.5} & 88.5 {\scriptsize $\pm$ 0.2} \\
     & \textsc{fs} & 76.6 {\scriptsize $\pm$ 0.5} & 78.4 {\scriptsize $\pm$ 0.4} & 90.8 {\scriptsize $\pm$ 0.2} \\
     & \textsc{w}  & 77.7  {\scriptsize $\pm$ 0.4} & 77.7 {\scriptsize $\pm$ 0.1} & 90.9 {\scriptsize $\pm$ 0.1} \\
     & \textsc{w+s} & \textbf{79.1} {\scriptsize $\pm$ 0.8} & \textbf{79.2} {\scriptsize $\pm$ 0.0} & \textbf{91.3} {\scriptsize $\pm$ 0.4} \\
    \midrule
    
     
    \multicolumn{2}{l}{RoBERTa-L} 
    & 64.0 {\scriptsize $\pm$ 0.1} & 65.7 {\scriptsize $\pm$ 0.1} & 92.5 {\scriptsize $\pm$ 0.3} \\
    \bottomrule
  \end{tabularx}
  \caption{Pun detection F1-score $\pm$ STD over 3 runs. Top results per model are in bold.}
  \label{tab:rq1-results}
\end{table}

\subsection{Results}
The average F1-score for these datasets across 3 runs is shown in Table \ref{tab:rq1-results}. 
Additional results can be found in the Appendix (see Table \ref{tab:hom_het_recall} and Fig. \ref{fig:boxplot-base-joker}).
All LLMs achieved F1-scores around 0.8 on the \NAPData\ dataset with the best prompting strategy. GPT-4o performed the best, while 
Mistral had the lowest scores. Notably, Mistral only performed well in the zero-shot setting and struggled to learn from additional context in the prompt.

Results on the JOKER dataset show a similar trend but are slightly more challenging for all models. This is likely due to its method of creating non-puns by replacing a single word, resulting in more ``adversarial'' examples. In contrast, in the \NAPData\ dataset, we rephrased puns into non-puns, potentially replacing or removing multiple words.


Aside from Mistral, which consistently performs poorly on our new datasets, the \textsc{w}/\textsc{w+s} prompts --- which require LLMs to justify each decision with a rationale --- achieve the highest performance, with an average improvement of +3\% in F1-score compared to the few-shot configuration (more details in Appendix §\ref{supp:results}). This finding aligns with previous research \cite{sun-etal-2022-expunations} that shows explanations can enhance pun detection performance. 
However, the reasoning prompt did not yield significant improvements over the rationale-augmented prompts (\textsc{w} and \textsc{w+s}) on the \NAPData\ and JOKER datasets. 
Overall, the strong performance suggests that LLMs can understand puns to some extent and benefit from jointly detecting and explaining.

\begin{figure}[t!]
  \includegraphics[width=0.32\linewidth]{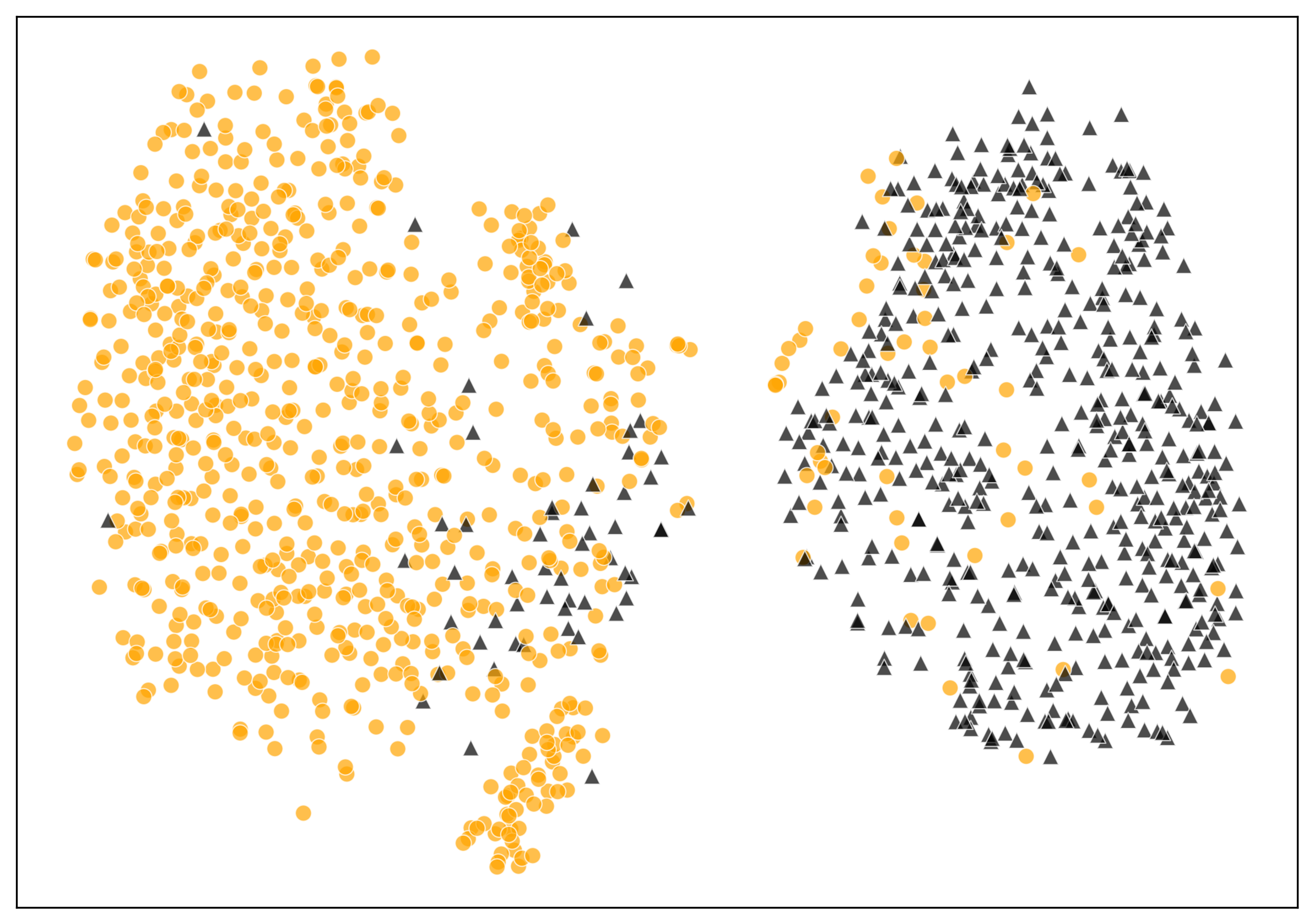}\hfill
  \includegraphics[width=0.32\linewidth]{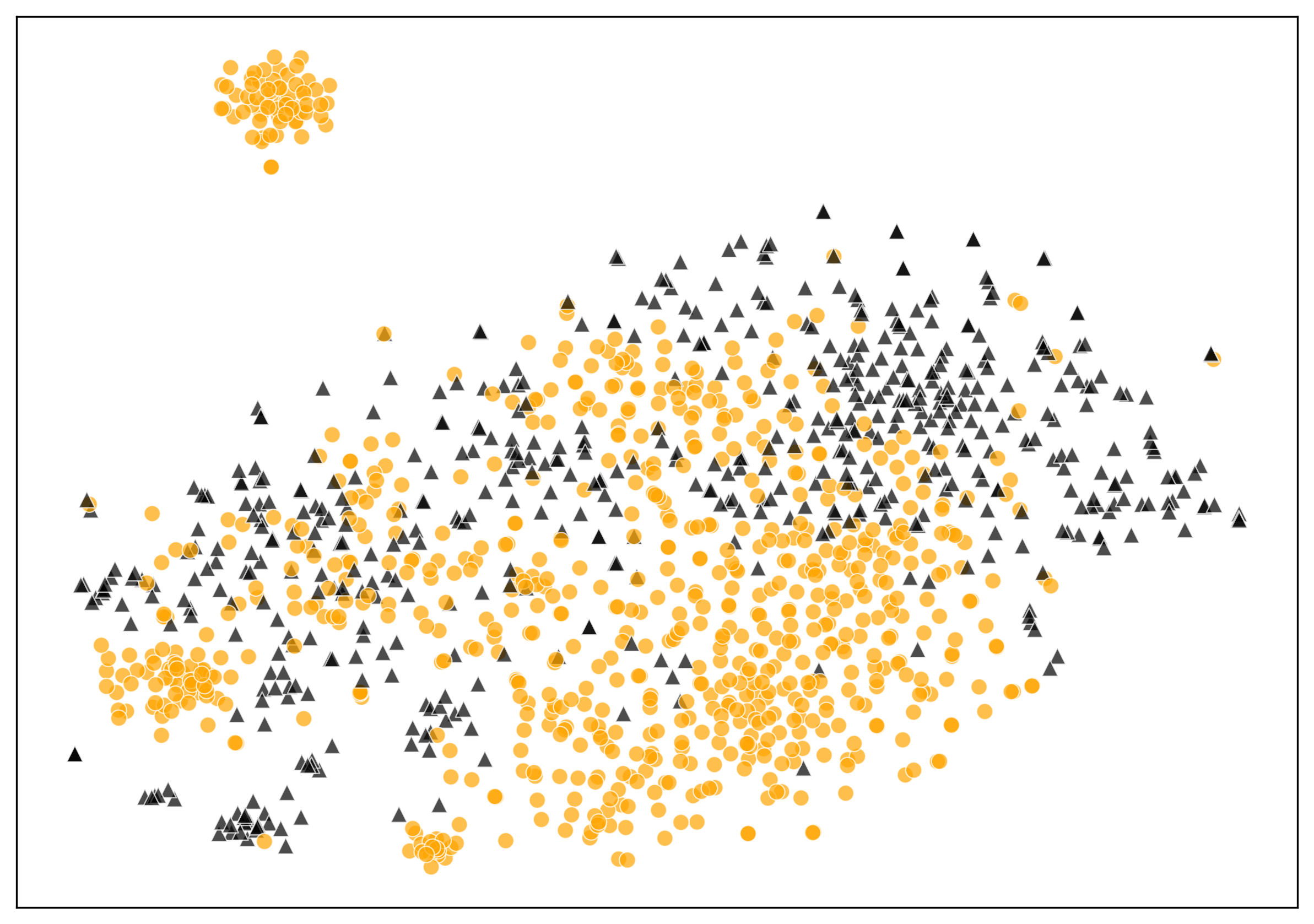}\hfill
  \includegraphics[width=0.32\linewidth]{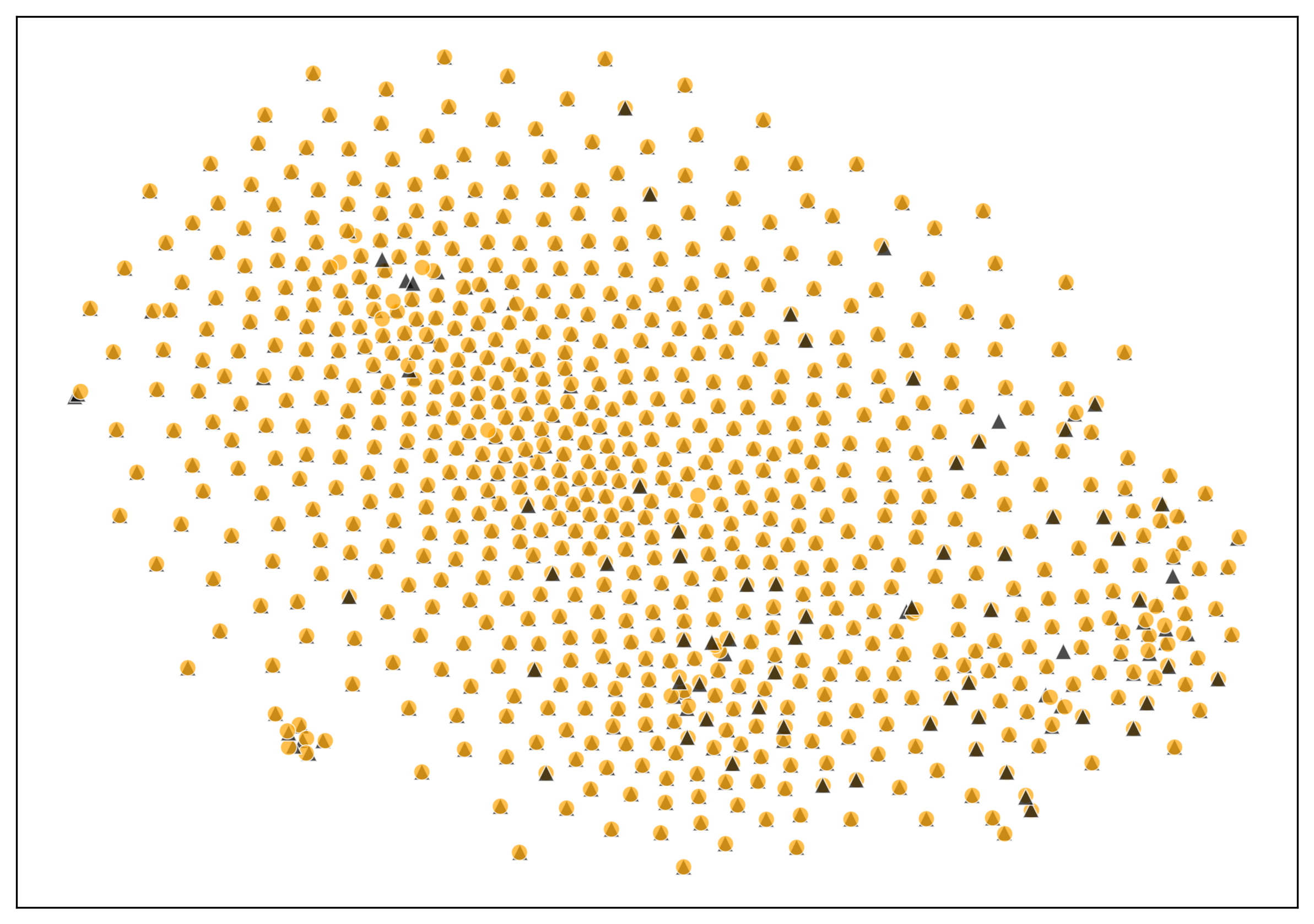}  
  \caption{Separation of PunEval embeddings for the RoBERTa model: after fine-tuning (left), before fine-tuning (middle), and \NAPData + JOKER separation before fine-tuning (right). Orange dots represent puns, while black triangles represent non-puns. Notably, PunEval puns exhibit greater separability in the embedding space compared to \NAPData/JOKER, even before fine-tuning.}
  \label{fig:emb-separation}
\end{figure}

\paragraph{Embedding Space Analysis.}
In addition to the comparison LLMs, we also include a RoBERTa-large encoder model \cite{liu2019robertarobustlyoptimizedbert} that has been fine-tuned on the PunEval training split. The significant performance difference observed between PunEval (92.5) and the other datasets (64.0 in \NAPData\ and 65.7 in JOKER) prompted an investigation to understand the reasons behind this disparity. We analyzed the embedding distribution of the RoBERTa model using t-SNE \cite{TSNEvandermaaten} to visualize the embeddings in a 2D space (Fig. \ref{fig:emb-separation}, see details in Appendix §\ref{supp:exp_details}).
The visualizations reveal poor separation between subtly altered positive and negative examples in the \NAPData\ and JOKER datasets (Fig. \ref{fig:emb-separation}, right). In contrast, PunEval embeddings show better separation even before fine-tuning (Fig. \ref{fig:emb-separation}, center), with clusters often corresponding to recurring templates. For instance, the top-left cluster from the middle image represents puns structured as ``Old [...] never die, they just'', a common pattern for creating puns (e.g., ``\textit{Old bankers never die, they just lose interest}''). 
This suggests that dataset artifacts or structural similarities in PunEval may facilitate shortcut learning. 

\section{RQ2: How Robust Are LLMs at Detecting Puns?}\label{sec:rq2}

Given the analysis in the previous section, one may wonder whether the performance gap between the PunEval dataset and the other two datasets is partly due to LLMs recognizing superficial cues or overfitting to specific pun structures without fully understanding their meanings. This observation prompts us to investigate the LLMs' robustness to common pun language patterns and simple pun alterations.

\subsection{Pun Language Patterns}\label{sec:rq2.1}
The PunEval dataset contains language patterns frequently used to create puns. Examples include ``\textit{Old [...] never die, they just [...]}'' and ``\textit{Tom}'' (a common name in English jokes), which may serve as shortcuts for recognizing puns.


To analyze the impact of these recurrent patterns on model performance, we extracted bag-of-n-gram features and trained a logistic regressor for pun detection on the PunEval training set. By examining the top-20 expressions based on learned coefficients and frequency, we manually identified six patterns that significantly correlate with the presence of a pun. These patterns, along with examples and their frequencies, are listed in Tables \ref{tab:semeval_issues} and \ref{tab:pattern_occurrences} in the Appendix. While their presence has been noted previously \cite{joker_corpus}, no detailed analysis of their effects has been provided.
In the PunEval test split, 171 out of 173 samples that contain these patterns are puns. Excluding them results in a 2-3\% drop in F1-score and a 2-6\% drop in precision across all models compared to the full test split, confirming that the presence of these recognizable patterns can inflate performance. To further investigate this issue, we created a new benchmark to test robustness against this bias.

\subsubsection{Data}
For each of the six extracted patterns, we sampled 100 puns from the PunEval test split that contain the same expression, supplementing with new examples collected from the Internet or generated using GPT-4o (verified manually) as needed to complete the set. More details on this collection are reported in Appendix §\ref{supp:unbiased}.
We also generated an equal number of verified non-pun sentences using GPT-4o (with human oversight), ensuring they contain the same expression. The resulting dataset, referred to as the \emph{\UnbiasedData} dataset, contains 1,200 samples, two of which are presented in Example \ref{example:datarq2.1}.
\begin{qaexample}[t!]
\small 
\begin{mergedcolorboxtop}[egreen]{Original pun}
I used to be a comedian, but my life became \underline{a joke} (\textit{joke}).
\end{mergedcolorboxtop}
\begin{mergedcolorboxbottom}[ered]{Rephrased pun}
I used to be a comedian, but my life became \underline{chaotic}.
\end{mergedcolorboxbottom}
\caption{A pun with a common pun language pattern and a derived non-pun.}
\label{example:datarq2.1}
\end{qaexample}

\subsubsection{Results}

Table \ref{tab:rq2.1-results-reduced} compares the average performance metrics over 3 runs between the entire \UnbiasedData\ dataset and the PunEval dataset. 
More detailed results for each language pattern can be found in the Appendix (see Table \ref{tab:rq2.1-results-complete} and Fig. \ref{fig:rq2.1-bias-plots}).
\begin{table}[t!]
  \small
  \centering
  \begin{tabular}{lHlll}
    \toprule
    \textbf{Model} & Best prompt & \textbf{F1} & \textbf{Precision} & \textbf{Recall} \\
    \midrule
    Gemini2.0 & \textsc{w+s}    & 76.9 (\red{-12.2})  & 66.7 (\red{-20.9}) & 91.6 (\green{~~+0.9}) \\ 
    GPT-4o & \textsc{w}         & \textbf{83.1} (\red{~~-4.2})   &\textbf{79.7} (\red{-18.2}) & 88.1 (\green{~~+9.4}) \\
    Llama3.3 & \textsc{w}       & 81.0 (~~\red{-6.2})    &71.5  (\red{-16.4}) & 94.2 (\green{~~+7.7}) \\
    Mistral3 & \textsc{0s}      & 77.4 (~~\red{-3.4})   &72.1 (\red{-14.8}) &87.5 (\green{+11.9}) \\
    Qwen2.5 & \textsc{w}        & 79.7 (~~\red{-7.1})   &74.1 (\red{-19.3}) &87.5 (\green{~~+6.5}) \\
    R1-D & \textsc{w+s}         & 78.3 (\red{-10.0})    &66.9 (\red{-18.7}) &94.8 (\green{~~+3.7}) \\
    R1 & \textsc{w+s}           & 81.1 (\red{-10.2})  &69.5 (\red{-15.9}) &\textbf{98.3} (\green{~~+0.3}) \\
    \bottomrule
  \end{tabular}
  \caption{F1-score, precision, and recall (\%) on the best prompt on \UnbiasedData\ and the absolute performance difference with the PunEval dataset.}
  \label{tab:rq2.1-results-reduced}
\end{table}
Overall, there is an average drop of 16-23\% in precision and 4-13\% in F1 on the \UnbiasedData\ dataset across all models. 
Additionally, we observe a sharp imbalance of low precision and high recall for some of the patterns (see Fig. \ref{fig:rq2.1-bias-plots}), which indicates LLMs are prone to identifying puns whenever they observe a typical pun-like pattern.
These results suggest that most LLMs tend to process certain patterns somewhat superficially, lacking understanding of the underlying principles of well-crafted puns. 
\subsection{Pun Alterations}\label{sec:rq2.2}

Motivated by our previous results, 
we investigate whether LLMs exhibit robustness to simple alterations of puns designed to ruin them. For humans, the pun effect relies on the perception of two well-supported senses that create ambiguity and contrast. Therefore, replacing the pun word ($w_p$) with a lexical unit that fails to convey this duplicity would result in the loss of such effect, and the failure of the intended pun. 

\subsubsection{Data}
To systematically investigate LLMs' understanding of this phenomenon, we replace the pun word $w_p$ with multiple alternative expressions and measure their ability to classify ``ruined'' puns. To this end, we randomly select 100 heterographic and 100 homographic puns from the \NAPData\ and PunEval datasets and modify each pun with four different substitutions:
\begin{itemize}
\itemsep-0.1em 
    \item \texttt{Pun syn}: a synonym or hypernym of the pun word $w_p$, different in phonetics and orthography.
    \item \texttt{Alt syn}: as above, but for the alternative word $w_a$.
    \item \texttt{Homophone}: a nonsensical expression phonetically similar to $w_p$.
    \item \texttt{Random}: a nonsensical random word.
\end{itemize}
Additionally, we include 100 randomly generated sentences (non-puns) as a control group, resulting in a total of 1,100 examples, which we refer to as the \textit{\SubstitutionData}\ dataset. A complete set of substitutions is shown in Example \ref{example:datarq2.2}, and more details on the generation procedure are in Appendix §\ref{supp:substitution_data}. Note that while some non-sensical replacements may be humorous due to the absurdity of the resulting sentences, they do not represent valid puns.
\begin{qaexample}[t!]
\small 
\begin{mergedcolorboxtop}[egreen]{Original pun}
Long fairy tales have a tendency to \underline{dragon} (\emph{drag on}).
\end{mergedcolorboxtop}
\begin{mergedcolorboxmid}[ered]{\texttt{Pun syn} substitution}
Long fairy tales have a tendency to \underline{wyvern}.
\end{mergedcolorboxmid}
\begin{mergedcolorboxmid}[ered]{\texttt{Alt syn} substitution}
Long fairy tales have a tendency to \underline{prolong}.
\end{mergedcolorboxmid}
\begin{mergedcolorboxmid}[ered]{\texttt{Homophone} substitution}
Long fairy tales have a tendency to \underline{brag gone}.
\end{mergedcolorboxmid}
\begin{mergedcolorboxbottom}[ered]{\texttt{Random} substitution}
Long fairy tales have a tendency to \underline{tick}.
\end{mergedcolorboxbottom}
\caption{Example substitutions from the \SubstitutionData.}
\label{example:datarq2.2}
\end{qaexample}

\subsubsection{Results}

\begin{figure}[t!]
  \centering
  \includegraphics[width=1.0\linewidth]{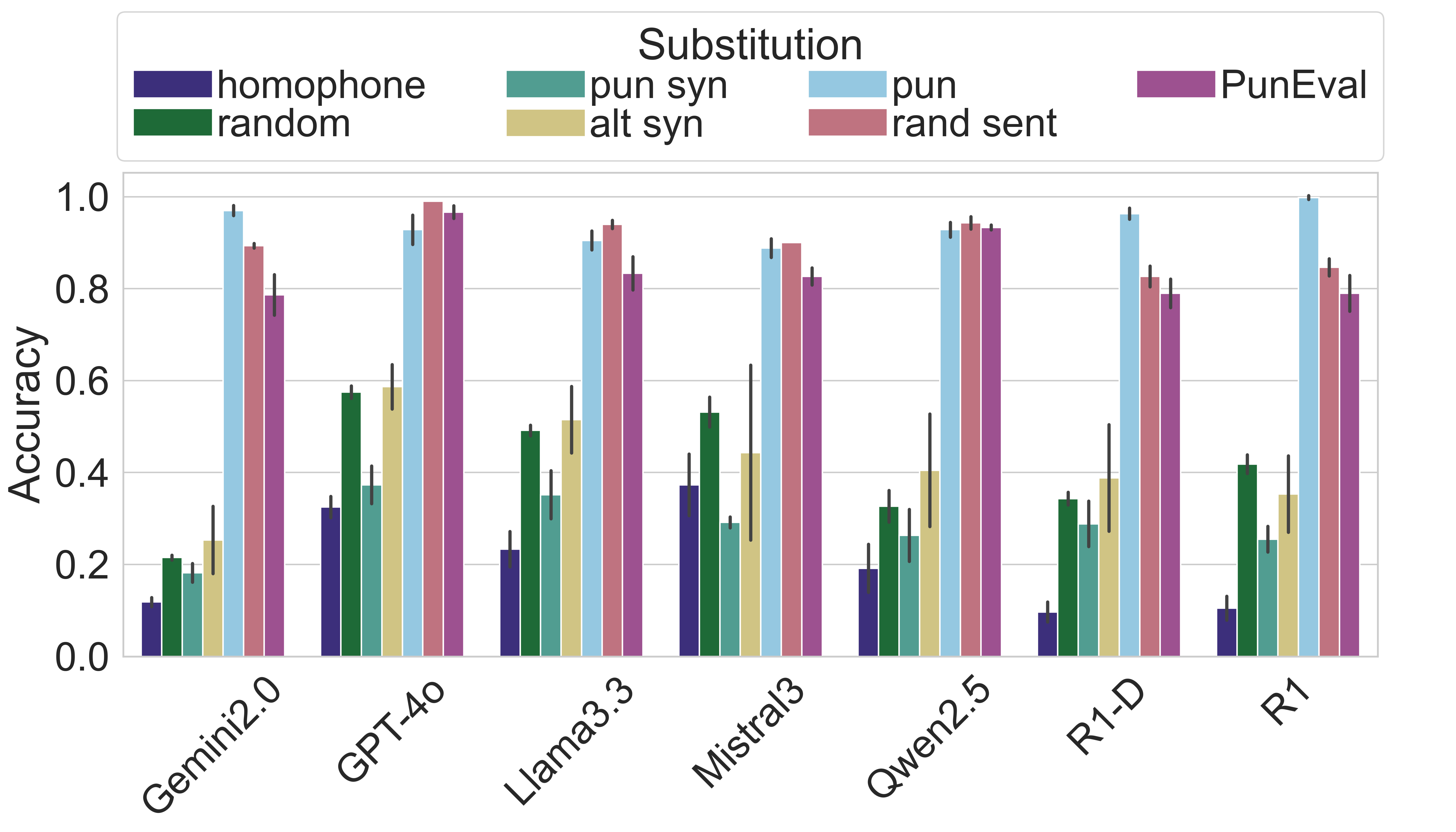}
  \caption {Binary accuracy of comparison LLMs with their best prompt on the \textit{\SubstitutionData}\ dataset, with error bars representing the standard deviation over 3 runs.}
  \label{fig:alterations_performance}
\end{figure}

The binary accuracy of each LLM for every substitution subset, using each model's best prompt from the RQ1 experiments, is shown in Fig. \ref{fig:alterations_performance}. The ``pun'' subset contains exactly 200 puns (100 het-puns and 100 hom-puns); every other subset contains exactly 200 non‑puns representing the various substitutions. Results from single prompts are reported in Appendix §\ref{supp:rq2.2}. The ``rand sent'' results are obtained from a control of 100 randomly generated non-puns to detect any unseen bias toward the pun class; scores exceed 0.8, indicating no such bias in the random sentences.

All models exhibit a bias toward labeling altered examples as puns, reflected in low accuracy across all substitution categories, which contain only negative examples (non-puns). In particular, GPT‑4o, the best performer overall, reaches a low of 0.33 on the \texttt{homophone} subset and a high of 0.59 on \texttt{alt syn}. Conversely, all models achieve accuracy of around 0.8 or higher on the negative examples from PunEval and the control group (\texttt{rand sent}) (see Table \ref{tab:rq2.2-results-recall} for details). The \texttt{homophone} subset is the most challenging overall, indicating that phonetic similarity between $w_p$ and its replacement degrades the LLMs' ability to distinguish them in context. This may also be influenced by orthographic similarity, as similar phonetics often imply similar orthography (e.g., \emph{stock}/\emph{stork} or \emph{dragon}/\emph{brag gone} in Example \ref{example:datarq2.2}).
Pun-word synonym substitutions (\texttt{pun syn}) also generally challenge the models, suggesting they may overlook the effect of the replaced word when it is semantically similar to a more fitting alternative. This suggests that LLMs might be actively ``fixing'' the pun in their reasoning process, thereby rectifying the alteration.
The \texttt{random} and \texttt{alt syn} sets show slightly higher accuracy overall, further indicating that semantic or phonetic similarity to valid puns misleads predictions more than loosely related substitutions.
These results suggest that LLMs fail to recognize contexts that should not trigger the \emph{pun effect} in cases where the target sentence has a typical pun structure.

\section{RQ3: To What Extent can LLMs Explain Puns?}\label{sec:rq4}
Our rationale-augmented prompts achieve the best performance in our previous experiments, notably reducing bias on the \UnbiasedData\ dataset with a 9.8\% increase in precision (see §\ref{supp:rationale-effect} in the Appendix for details on rationale impact). In this section, we systematically examine the quality of the generated rationales to gain insights into LLMs' understanding of puns.

\subsection{Keyword Evaluation}\label{sec:rq4.1}
Our goal is to measure the overlap between human-provided semi-structured rationales and LLM-generated rationales, while also accounting for prediction accuracy.
To this end, we quantify the alignment between the \emph{LLM-predicted pun pair} ($w_p'$, $w_a'$) and the true pun/alternative word pair ($w_p$, $w_a$) using the \emph{pun pair agreement} (PPA) metric. Each prediction is scored: 2 if both $w_p$ and $w_a$ are correctly generated, 1 if only one is correct, and 0 if neither is correct. Predictions with incorrect labels receive a score of 0, since no rationales are generated for non-pun predictions. For further details on the evaluation process, see §\ref{supp:ppa_details}.

\paragraph{Results.}
Table \ref{tab:rq3-agreement} reports the average PPA over three runs, representing the mean number of words correctly identified in generated rationales for pun pairs on a 0--2 scale. 
GPT‑4o is the top performer, with $\sim$1.5 out of 2 correctly predicted words per pun across all three datasets. It is followed by Llama3.3 and DeepSeek‑R1, whose performance is noticeably lower. Mistral has the lowest scores, trailing the next-lowest model, Gemini2, by $\sim$33\%. Among the 70B models, Llama3.3 performs best on average. We also observed no significant difference between the two prompts tested.
Consistent with our earlier results, scores on the PunEval datasets are generally higher than on JOKER and \NAPData, where performance drops by approximately 20\% and peaks at 1.5 out of 2. 
Furthermore, a PPA evaluation restricted to true positives (see Appendix Table \ref{tab:rq3-agreement-tp}) shows that most models, except Qwen and Mistral, can effectively justify their correct answers. In this specific evaluation, GPT-4o and DeepSeek-R1 emerge as the top performers by a wide margin. 
These findings motivate our investigation of the main error types in model reasoning, which is addressed in the next subsection.

\begin{table}[t]
  \small
  \centering
  \begin{tabular}{lcccccc}
    \toprule
    \multirow{2}{*}{\textbf{Model}} & 
    \multicolumn{2}{c}{\textbf{\NAPData}} & 
    \multicolumn{2}{c}{\textbf{JOKER}} & 
    \multicolumn{2}{c}{\textbf{PunEval}} \\
    \cmidrule(lr){2-3} \cmidrule(lr){4-5} \cmidrule(lr){6-7} 
    & \textsc{w} & \textsc{w+s} 
    & \textsc{w} & \textsc{w+s} 
    & \textsc{w} & \textsc{w+s} \\
    \midrule
    Gemini2.0
    & 1.2 & 1.2 
    & 1.1 & 1.1
    & 1.5 & 1.5 \\
    GPT-4o 
    & \textbf{1.5} & \textbf{1.5}
    & \textbf{1.5} & \textbf{1.4}
    & \textbf{1.6} & 1.6 \\
    Llama3.3 
    & 1.4 & 1.3 
    & 1.3 & 1.2
    & 1.5 & 1.5 \\  
    Mistral3 
    & 0.8 & 0.7
    & 0.8 & 0.7
    & 1.2 & 0.9\\
    Qwen2.5 
    & 1.2 & 1.2
    & 1.2 & 1.2
    & 1.5 & 1.5 \\
    R1-D 
    & 1.3 & 1.2
    & 1.2 & 1.2
    & 1.5 & 1.6 \\  
    R1 
    & 1.3 & 1.2
    & 1.2 & 1.3
    & \textbf{1.6} & \textbf{1.7}\\
    \bottomrule
  \end{tabular}
  \caption{Average Pun Pair Agreement (PPA), in $[0\text{--}2]$, for each prompt setting (\textsc{w}, \textsc{w+s}) across datasets. Standard deviation is 0.0 for all measurements, and the best results are marked in bold.}
  \label{tab:rq3-agreement}
\end{table}




\subsection{Error Analysis: Manual Evaluation}
The PPA metric measures the ability to match the original pun pair but does not indicate \emph{why} predictions were wrong, nor does it evaluate the complete rationale obtained with the more expressive \textsc{w+s} prompt. Upon inspection of the produced rationales, we identify four main categories of mistakes frequently made by our LLMs regarding the predicted pun pair ($w_p'$, $w_a'$) and the associated pair of senses ($s_p'$, $s_a'$):
\begin{itemize}
\itemsep-0.1em 
    \item \textbf{Word-sense association} (\texttt{word-sense} error): incorrect association between word and sense (i.e., $w_p'$--$s_p'$ or $w_a'$--$s_a'$).
    \item \textbf{Missing context} (\texttt{context}): the interpretation of $w_p'$ as $s_p'$ or $w_a'$ as $s_a'$ is not supported by the context.
    \item \textbf{Pun pair} (\texttt{pun pair}): $w_p'$ and $w_a'$ are not sufficiently similar in phonetic or orthographic terms to create wordplay (this only applies to het-puns).
    \item \textbf{Senses similarity} (\texttt{sense sim}): senses $s_p'$ and $s_a'$ are too similar. Puns work when polysemous words carry well-separated senses that create contrast.
\end{itemize}

These types of mistakes also highlight different weaknesses. A word-sense mistake can be considered a hallucination (e.g., ``crustacean'' interpreted as ``a species of bird'') while a pun pair mistake (as in Fig. \ref{example:intro}) indicates difficulty in grasping the phonological properties of words. 




To precisely characterize the model's mistakes in the categories above, we conducted an error analysis with five native English speakers specifically hired for this task. We randomly sampled 80 incorrect answers from each of the three top-performing LLMs, according to the PPA evaluation: GPT-4o, Llama3.3, and DeepSeek-R1. For each of the sampled answers, we designed a set of questions to determine the presence of each category of mistake. For example, the first two questions assess whether $s_p'$ (the LLM-predicted pun sense) and $s_a'$ (the predicted alternative sense) are legitimate interpretations of $w_p'$ (predicted pun word) and $w_a'$ (predicted alternative word), respectively. Annotators could answer with ``Yes'', ``No'', or ``Maybe'' if uncertain. If any of these questions is answered with ``No'' for a given sample, that sample is counted as a \texttt{word-sense} association error. Note that if both senses are incorrect, it still counts as a single error, ensuring a more comparable error count across all mistake categories. The full evaluation procedure, guidelines, and questionnaire are detailed in §\ref{supp:ea_details} of the Appendix. Table \ref{tab:examples-error-analysis} shows the error-analysis results for six example samples.

\paragraph{Results.}
\begin{table}[b]
  \small
  \centering
  \setlength{\tabcolsep}{9pt}
  \begin{tabular}{lccc}
    \toprule
    \multicolumn{4}{c}{\textbf{Error Breakdown by Category}} \\
    \midrule
    \textbf{Error Category} & \textbf{GPT-4o} & \textbf{Llama3.3} & \textbf{R1} \\
    \midrule
    \texttt{context}      & {66}   & \textbf{49}   & 50 \\
    \texttt{pun pair}     & {50}   & 39   & \textbf{27} \\
    \texttt{word-sense}   & 9             & {17}  & \textbf{8} \\
    \texttt{sense sim}    & \textbf{2}             & {3}   & \textbf{2} \\
    \midrule
    \multicolumn{4}{c}{\textbf{Aggregated statistics}} \\
    \midrule
    Total errors  & {128}  & 111  & \textbf{87} \\
    Avg. errors     & {1.6}  & 1.4  &\textbf{ 1.2} \\
    \bottomrule
  \end{tabular}
  \caption{Error statistics by category and overall on 80 misclassified samples from the \SubstitutionData\ dataset. Note that there may be more than one error per sample.}
  \label{tab:error_analysis_count}
\end{table}


    


Table \ref{tab:error_analysis_count} reports the results of the manual evaluation for the three analyzed LLMs.
Our analysis reveals that, in general, the most frequent errors involve: (i) missing the required context to support both $s_p$ and $s_a$ (\texttt{context}) and (ii) selecting unsuitable word pairs that do not fit into a pun (\texttt{pun pair}). The latter category only applies when $w_a' != w_p'$, and typically results from forcing words that are not phonetically compatible to create wordplay. A third fairly common mistake arises from incorrectly pairing words with their meanings (\texttt{word-sense}). Llama3.3, the smallest among the three LLMs, tends to struggle the most in this area, producing more sense hallucinations. In contrast, GPT-4o and R1 tend to assess the senses correctly, even when some words in the pun pair are incorrectly identified. 
These error patterns suggest a lack of understanding of the mechanisms of puns, particularly the inability to distinguish inappropriate contexts and difficulties in handling wordplay based on phonetic or orthographic similarity. Among the three evaluated models, DeepSeek-R1 proved to be the most precise, exhibiting the lowest number of errors in its responses. 


\subsection{Discussion}

Our findings align with prior work showing that LLMs struggle with humor, particularly when it depends on nuanced context or roleplay \cite{canAIjoke,robotwalks}. These studies note that safety constraints (e.g., the ``Harmless, Helpful, Honest'' criteria) can cause LLMs to misinterpret relational context in potentially offensive situations. These safety mechanisms bias models toward a safer, ``washed-out'' form of humor, stripping away the surprise and edge on which many jokes depend. 
Puns, in contrast, are rich in surprise, subtlety, and timing; they are also deeply rooted in human emotions and experiences, qualities that current LLMs cannot fully tap into.

LLMs also exhibit a form of ``regressive sycophancy'' --- a tendency to agree with a user prompt even when incorrect --- particularly in ambiguous or sensitive contexts \cite{syncophancyllmscauses,cheng2025socialsycophancy}. We suspect that this tendency causes the models to produce false positives by forcing inputs into a ``pun'' template. Such behavior is a known artifact of human-preference alignment (post-training), which can bias models toward agreeable responses at the expense of contextual and factual accuracy \cite{sharma2025understandingsycophancylanguagemodels}.

Finally, the frequent \texttt{pun pair} errors point to weaknesses in assessing phonetic and orthographic similarity. These issues are likely tied to tokenization, which can mask morphological elements critical to wordplay, and to limited phonological modeling \cite{liao_how_2025}. The composition of pre-training corpora and data cleaning practices may also deprive models of the informal, creative language where puns are common. Because developers rarely disclose detailed preprocessing and corpus information, it is difficult to determine how these practices affect fluency with such language.

Improving an AI's grasp of puns and humor in general will likely require deeper human-machine collaboration and greater social awareness. Prior work suggests that human alignment of LLMs should shift from a single global standard to a more granular, audience-aware approach \cite{canAIjoke,robotwalks}. This would help models to adapt to specific audiences and contexts, adjusting appropriateness and fairness criteria accordingly.

\section{Conclusion}

Results in pun detection benchmarks often show a high performance by current LLMs. However, it remains unclear whether this high performance comes from true understanding or other data-specific factors. 
To assess robustness, we challenged state-of-the-art LLMs with two new annotated datasets that we publicly released for future research: \UnbiasedData, which collects examples with language patterns common in English puns, and \SubstitutionData, containing sentences subtly modified to mimic puns.
Although LLMs can detect puns on existing benchmarks ($>$ 0.83 average accuracy), accuracy drops substantially on our more challenging datasets: $-$15\% on \UnbiasedData, and $-$50\% on \SubstitutionData, indicating limited understanding of pun mechanisms and over-reliance on similarity with known puns. 
We also used LLMs to explain puns by generating semi-structured rationales and evaluated those explanations with automatic and manual assessments. Only three models correctly identified the pun-related words in more than 70\% of cases. Moreover, they all struggled to recognize when contexts and word choice legitimately supported wordplay.  
This research underscores the need for more rigorous evaluations of LLMs' performance on ambiguous-language tasks, particularly those requiring subtle language understanding, such as pun detection and generation.

\section*{Limitations}
A limitation of our work is that, despite our efforts to create high-quality data, the newly contributed datasets have not been comprehensively validated by language experts. Our manual error analysis indicates the possibility of incorrect annotations or ``accidental'' puns, stemming from the inherently subjective nature of pun perception. Discussions with annotators also revealed that the definition of a pun should be more formally characterized; we therefore provided many examples before annotation and corrected some samples based on their feedback.
Despite these efforts, our manual error analysis remains limited by the subjectivity of several questions.

Additionally, there is the possibility that most LLMs have previously encountered many examples from our datasets, as puns often exist in various versions and can be easily found online. Consequently, we may not have fully disentangled this effect from our bias measurements on the \SubstitutionData\ and, in particular, the \UnbiasedData\ dataset, the latter of which includes many examples sourced from PunEval. We believe that further tests --- such as injecting the identified language patterns into arbitrary sentences --- could help clarify the impact of model bias on performance. 

Future work should investigate the severity and correctability of these biases. As reviewers suggested, it remains to be tested whether in-context learning with targeted negative examples, such as those from our new datasets, can mitigate the models' tendencies to classify most examples as puns. Moreover, the fact that our evaluation was conducted solely on English puns further limits the scope of the conclusions that can be drawn from these results.

Finally, our prompt design was tailored for Llama 3.1, and we utilized the same prompt for all LLMs. We acknowledge that custom-tailored prompts for each model are likely to impact performance. During this work, several new LLMs were published, and while we tested some of them, we did not have the opportunity to extensively optimize prompts for each one.

\section*{Ethics Statement}
All puns used in this study were either sourced from previous work publicly released for research purposes, devised by the authors, generated using LLMs, or obtained from other open sources. In all instances, proper citations are provided to acknowledge the original authors. All collected puns are used solely for research purposes, and we do not claim authorship over any of them. As per the authors' request, the JOKER dataset was not released and was used exclusively for this study.


The pun datasets do not contain any personal information and do not refer to specific individuals; however, some may be considered offensive by certain audiences. While we have limited the number of offensive puns in our new datasets, such content may still exist in other datasets utilized in this work.

Human judges were hired for their participation in the evaluation process. Each annotator responded to 7-9 questions, which included multiple-choice answers and a Likert scale for two questions. Annotators had the option to select a specific answer or skip questions if they did not possess sufficient knowledge to respond. They were compensated 15-20£ per hour, which is well above the minimum wage in the country where the annotations took place. The annotators, who were students from diverse backgrounds --- including Computer Science, Theology, and Literature --- were interviewed by the authors in a virtual meeting following the initial pilot study.


\section*{Acknowledgments}
We thank Felix Chadwick-Smith, Fayeja Farhana, Dhruvil Raval, Kayleigh Shier, and Hannah Trotman for their work as annotators on the manual error analysis of rationales. We also thank the entire Cardiff NLP group for their guidance on the project and for feedback on an early draft of the paper. Finally, we are grateful to Dr. Tristan Miller and co‑authors for kindly allowing us to use the JOKER dataset in our evaluation.

This study was funded by the European Union - NextGenerationEU, in the framework of the iNEST - Interconnected Nord-Est Innovation Ecosystem (iNEST ECS\_00000043 – CUP H43C22000540006). The views and opinions expressed are solely those of the authors and do not necessarily reflect those of the European Union, nor can the European Union be held responsible for them. 

Jose Camacho-Collados was supported by a UKRI Future Leaders Fellowship.


\bibliography{custom}

\appendix
\clearpage

\section{Experimental Details}
\label{supp:exp_details}

Throughout all experiments, the temperature was set to 0 to ensure more deterministic outputs, while the beam search decoding strategy was used for models that support it (beam size = 2). RoBERTa-large\footnote{\url{https://huggingface.co/FacebookAI/roberta-large}} was trained on the PunEval training split for max 4 epochs with early stopping on the validation set, based on F1-score. We use a batch size of 32 and a learning rate of $1.5e^{-5}$. 

For testing, we utilized hosted instances of Llama 3.3-70B, Mistral, DeepSeek-R1-DistiLlama, and Gemini 2.0-Flash through OpenRouter\footnote{\url{https://openrouter.ai}}. The RoBERTa model was fine-tuned on a local server using the latest dumps available on Hugging Face.

For the embedding space analysis, we averaged token embeddings to obtain a sentence-level representation for each sample, preferring this strategy over \texttt{CLS} token embeddings, as the \texttt{CLS} token is not used during RoBERTa pre-training. Clusters representing patterns were identified using DBSCAN \cite{dbscan} with UMAP \cite{umap} for dimensionality reduction. We employed k-nearest neighbor (kNN) to determine the optimal \textit{eps} hyperparameter.

\subsection{Prompts}\label{supp:prompts-valid}
The final prompts utilized are reported below. Table \ref{tab:prompt-eng} presents a comparison of performance with other prompting strategies based on the validation split of the PunEval dataset using GPT-4o, the best-performing LLM.
The \textsc{+zc} prompts refer to those used with the Zero-shot CoT strategy. We edited the original prompt by removing all examples and instructing the models to "think step by step." We first asked them to produce reasoning and then to output the final answer. Since the performance was very similar to the reasoning (\textsc{+R}) prompt, we preferred the latter, as it allowed for a single prompt call instead of two.
The \textsc{rationale} and \textsc{base} prompts refer to the initial prompts that utilized JSON formatting for output. This strategy did not work well for smaller models, leading to the adoption of our custom prompting format, showcased in Figs. \ref{prompt:0s},\ref{prompt:fs},\ref{prompt:w},\ref{prompt:w+s}.

\begin{table}[ht!]
  \small
  \centering
  \begin{tabular}{lc}
    \toprule
    \textbf{Prompt} & \textbf{F1-score} \\
    \midrule
    \multicolumn{2}{c}{Final prompts}\\
    \midrule
    \textsc{0s}&90.2\\
    \textsc{fs}&\textbf{95.6}\\
    \textsc{w}&93.9\\
    \textsc{w+s}&89.7\\
    \midrule
    \multicolumn{2}{c}{JSON prompts}\\
    \midrule
    \textsc{json-w+s}& \textbf{87.1} \\
    \textsc{json-0s}& 85.0 \\
    \midrule
    \multicolumn{2}{c}{Reasoning and Zero-shot CoT}\\
    \midrule
    \textsc{fs+R}&\textbf{95.1}\\
    \textsc{w+R}&93.3\\
    \textsc{w+s+R}&93.3\\
    \textsc{fs+zc}&91.4\\
    \textsc{w+zc}&88.3\\
    \textsc{w+s+zc}&87.7\\
    \bottomrule
  \end{tabular}
  \caption{F1-score of GPT-4o over different versions of prompts tried.}\label{tab:prompt-eng}
\end{table}

\subsection{Keywords Evaluation}\label{supp:ppa_details}
The output of the automatic evaluation indicates how many correct words the model produces in the rationale. During the evaluation process, both pun pairs are tokenized, lemmatized, and stripped of punctuation, with the overlap between each pair serving as the scoring metric. A score of 2 is assigned only if both $w_p'$ and $w_a'$ correctly match the true pun pair. A score of 1 is given if only one is correct, while a score of 0 is assigned if both are incorrect.
This evaluation is conducted both with and without lemmatization, retaining the maximum score to mitigate issues arising from incorrect lemmatization. 
Pun words and alternative words are lemmatized using SpaCy, specifically the \texttt{en\_core\_web\_lg} model\footnote{\url{https://spacy.io/models/en}}.

\subsection{Error Analysis}\label{supp:ea_details}



For our error analysis, we sample 80 examples from the set of false positives identified in the results on the \SubstitutionData\ dataset (which has the worst overall performance), sampling from results using the \textsc{w+s} prompt, which provides the most expressive rationale. Given that we have 200 examples for each of the four substitution categories, this represents 10\% of the dataset size.

We hired five native English speakers from various backgrounds based on their expertise in corpus annotation. Specifically, the group consists of three men and two women who are students at our university, ranging from undergraduate to postgraduate levels.

The annotators were asked to answer 7-9 multiple-choice questions assessing the reasonableness of the models' responses for each selected sample. We conducted an initial round of annotation on a small subset to validate agreement among the annotators and gather feedback. The pilot study resulted in moderate inter-annotator agreement, with Fleiss's $\kappa=0.41$).

We use Google Forms for collecting answers, and we show a screenshot of all questions in Fig. \ref{fig:ea_abc}, \ref{fig:ea_def}, and \ref{fig:ea_ghi}. Each annotator reviewed a batch of 40 examples and was asked to annotate according to a set of guidelines that clarified the nature of the task and provided examples. The full guidelines, along with the questions, are provided at the end of this Appendix. Note that after the pilot study, we added two additional questions to assess the presence of ``accidental'' puns. However, we ultimately excluded these from the final study because the results were inconsistent among reviewers, which led us to conclude that the questions were not properly formulated and produced overly subjective responses that did not provide reliable evidence.

\subsubsection{Results processing}\label{app:user-study-processing}

We identify each question with a letter from A to I (see Fig. \ref{fig:ea_abc}). Let $a(\ell)$ be the answer to the question identified by $\ell \in \{A, B, ..., I\}$. All questions have four possible answers: ``Yes'', ``No'', ``Maybe'' and ``I don't understand''. The exceptions are questions C and H, where respondents must express their rating on a Likert scale from 0 to 5. Although we received only one answer to question I, several annotators rated a small number of examples as possible puns, despite the low agreement on this question observed during the pilot test.

In our analysis, we adopt a conservative approach to detecting errors based on the annotators' responses, ensuring that errors are recorded only when the annotators express clear certainty. Hence, for answers on a qualitative scale, we assume $\text{``No''} = 0$ and all other answers are coded with $1$. In this case, each error category is assigned to samples when specific conditions are met:
\begin{itemize}
    \item \textbf{Word-sense association} (\texttt{word-sense}): $a(A)$ AND $a(B)$ (i.e., at least one of the senses must be clearly wrong);
    \item \textbf{Missing context} (\texttt{context}): $a(D)$ AND $a(E)$;
    \item \textbf{Pun pair} (\texttt{pun pair}): $a(F)$ OR $a(G)$ (i.e., the pun words must be both phonetically and orthographically diverse);
    \item \textbf{Senses similarity} (\texttt{sense sim}): $a(C) \le 1$.
\end{itemize}

Note that additional error categories may be applicable. In some cases, annotators, despite finding no errors according to our guidelines, marked the example as a non-pun. This observation highlights that our guidelines may not be exhaustive, in part due to the inherent subjectivity of pun interpretation.

\begin{figure*}
    \footnotesize

    \begin{promptop}[title={\textsc{w+s} (system)}]
    \emph{/* Definition */ }
    
    Puns are a type of wordplay that use words with multiple meanings or similar-sounding words to create humor by juxtaposing these different meanings.
    Non-puns are jokes or statements that do not rely on this kind of wordplay.
    A pun is created by a pair of words or short expressions, referred to as "w\_p" (the pun word) and "w\_a" (the alternative word), which together create a humorous effect.
    Note that "w\_p" and "w\_a" must be the minimal text spans that create the pun.
    Depending on the type of pun, either "w\_p" equals "w\_a", or only "w\_p" appears in the text, with "w\_a" being evoked by the context.
    Each of these expressions, "w\_p" and "w\_a", carries its own meanings, denoted as "s\_p" and "s\_a" respectively, and these meanings are supported by a set of contextual words.
    
    \emph{/* Instruction */}
    
    You are a helpful assistant tasked with analyzing texts to determine if they contain a pun or not.
    You must first answer with 'yes' if the given text is a pun and 'no' if it is a non-pun.
    If you think it is a pun, you must also justify your answer by providing "w\_p" and "w\_a", along with their meanings "s\_p" and "s\_a".
    \end{promptop}

    \begin{promptbot}[title={\textsc{w+s} (user)}]
    You must answer with 'yes' if the given text is a pun and 'no' if it is a non-pun.
    
    If you think it is a pun, you must also justify your answer by providing the words, or short expressions, "w\_p" and "w\_a", along with their meanings "s\_p" and "s\_a".
    Note that "w\_p" and "w\_a" must be the minimal text spans that create the pun, or empty strings if the text is a non-pun.
    "s\_p" and "s\_a" must contain short definitions of "w\_p" and "w\_a" that match their meanings in the context of the sentence, or empty strings if the text is a non-pun.
    Please provide your answer in one line using the following formats: 'yes <w\_p> <w\_a> <s\_p> <s\_a>' for puns and 'no <> <> <> <>' for non-puns.
    Do not add any additional text or characters.
    
    \emph{/* Examples */}
    
    Text: A carpenter sat on his drill and was bored to tears. 
    
    Output: yes <bored> <bored> <make a hole, especially with a pointed power or hand tool> <cause to be bored>

    Text: Don't kill the goose that lays the golden eggs? 
    
    Output: no <> <> <> <>

    Text: I scream, you scream, we all scream for ice cream! 
    
    Output: yes <ice cream> <I scream> <a dessert made from frozen sweetened cream, usually flavored> <to utter a long loud piercing cry, as from pain or joy>

    Text: He's dead Jim. Grab his tricorder. I'll get his wallet! 
    
    Output: no <> <> <> <>

    Text: The whistling fisherman was always out of tuna. 
    
    Output: yes <tuna> <tune> <any very large marine food and game fish of the genus Thunnus> <the property of producing accurately a note of a given pitch>

    Text: Better go about than fall into the ditch. 
    
    Output: no <> <> <> <>

    Text: \textbf{INPUT TEXT} Output:
    \end{promptbot}
    \caption{\textsc{w+s} system and user prompts.}\label{prompt:w+s}
\end{figure*}

\begin{figure*}
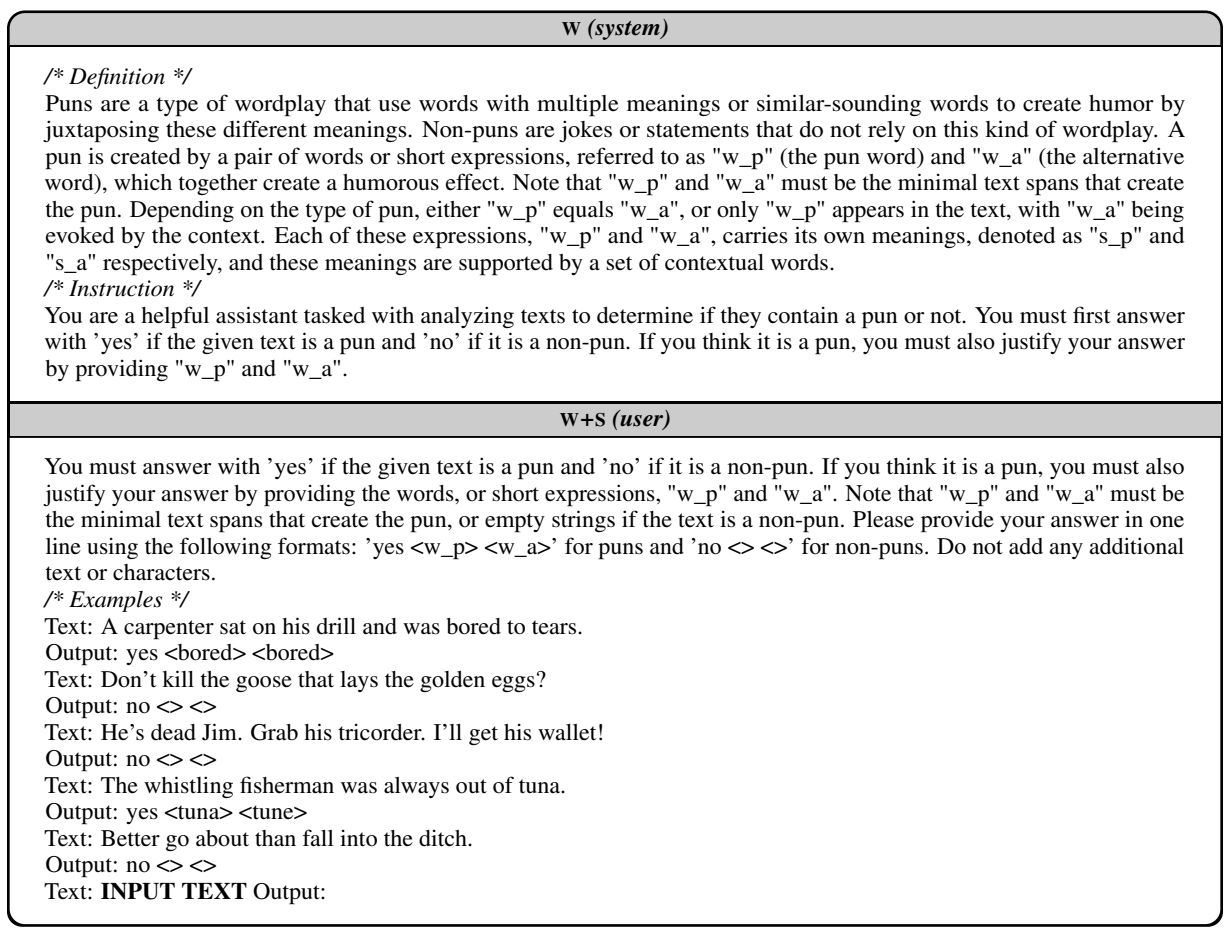

    \footnotesize

    \begin{promptop}[title={\textsc{w} (system)}]
    \emph{/* Definition */ }
    
    Puns are a type of wordplay that use words with multiple meanings or similar-sounding words to create humor by juxtaposing these different meanings.
    Non-puns are jokes or statements that do not rely on this kind of wordplay.
    A pun is created by a pair of words or short expressions, referred to as "w\_p" (the pun word) and "w\_a" (the alternative word), which together create a humorous effect.
    Note that "w\_p" and "w\_a" must be the minimal text spans that create the pun.
    Depending on the type of pun, either "w\_p" equals "w\_a", or only "w\_p" appears in the text, with "w\_a" being evoked by the context.
    Each of these expressions, "w\_p" and "w\_a", carries its own meanings, denoted as "s\_p" and "s\_a" respectively, and these meanings are supported by a set of contextual words.

    \emph{/* Instruction */}
    
    You are a helpful assistant tasked with analyzing texts to determine if they contain a pun or not.
    You must first answer with 'yes' if the given text is a pun and 'no' if it is a non-pun.
    If you think it is a pun, you must also justify your answer by providing "w\_p" and "w\_a".

    \end{promptop}

    \begin{promptbot}[title={\textsc{w+s} (user)}]
    You must answer with 'yes' if the given text is a pun and 'no' if it is a non-pun.
    If you think it is a pun, you must also justify your answer by providing the words, or short expressions, "w\_p" and "w\_a".
    Note that "w\_p" and "w\_a" must be the minimal text spans that create the pun, or empty strings if the text is a non-pun.
    Please provide your answer in one line using the following formats: 'yes <w\_p> <w\_a>' for puns and 'no <> <>' for non-puns.
    Do not add any additional text or characters.

    \emph{/* Examples */}
    
    Text: A carpenter sat on his drill and was bored to tears. 
    
    Output: yes <bored> <bored>

    Text: Don't kill the goose that lays the golden eggs? 

    Output: no <> <>

    Text: He's dead Jim. Grab his tricorder. I'll get his wallet! 
    
    Output: no <> <>

    Text: The whistling fisherman was always out of tuna. 
    
    Output: yes <tuna> <tune>

    Text: Better go about than fall into the ditch. 
    
    Output: no <> <>

    Text: \textbf{INPUT TEXT} Output:

    \end{promptbot}
    \caption{\textsc{w} system and user prompts.}\label{prompt:w}
\end{figure*}

\begin{figure*}
    \footnotesize

    \begin{promptop}[title={\textsc{fs} (system)}]
    \emph{/* Definition */ }

    Puns are a type of wordplay that use words with multiple meanings or similar-sounding words to create humor by juxtaposing these different meanings.
    Non-puns are jokes or statements that do not rely on this kind of wordplay.
    A pun is created by a pair of words or short expressions, referred to as "w\_p" (the pun word) and "w\_a" (the alternative word), which together create a humorous effect.
    Note that "w\_p" and "w\_a" must be the minimal text spans that create the pun.
    Depending on the type of pun, either "w\_p" equals "w\_a", or only "w\_p" appears in the text, with "w\_a" being evoked by the context.
    Each of these expressions, "w\_p" and "w\_a", carries its own meanings, denoted as "s\_p" and "s\_a" respectively, and these meanings are supported by a set of contextual words.

    \emph{/* Instruction */}
    
    You are a helpful assistant tasked with analyzing texts to determine if they contain a pun or not.
    You must answer only with 'yes' if the given text is a pun and 'no' if it is a non-pun.
    \end{promptop}

    \begin{promptbot}[title={\textsc{fs} (user)}]
    You must answer only with 'yes' if the given text is a pun and 'no' if it is a non-pun. Do not add any additional text or characters.

    \emph{/* Examples */}
    
    Text: A carpenter sat on his drill and was bored to tears. 
    
    Output: yes

    Text: Don't kill the goose that lays the golden eggs? 

    Output: no

    Text: I scream, you scream, we all scream for ice cream! 
    
    Output: yes

    Text: He's dead Jim. Grab his tricorder. I'll get his wallet! 
    
    Output: no

    Text: The whistling fisherman was always out of tuna. 
    
    Output: yes

    Text: Better go about than fall into the ditch. 
    
    Output: no

    Text: \textbf{INPUT TEXT} Output:
    \end{promptbot}
    \caption{\textsc{fs} system and user prompts.}\label{prompt:fs}
\end{figure*}

\begin{figure*}
    \footnotesize

    \begin{promptop}[title={\textsc{0s} (system)}]
    You are a helpful assistant tasked with analyzing texts to determine if they contain a pun or not.
    You must answer only with 'yes' if the given text is a pun and 'no' if it is a non-pun.
    \end{promptop}

    \begin{promptbot}[title={\textsc{0s} (user)}]
    You must answer only with 'yes' if the given text is a pun and 'no' if it is a non-pun. Do not add any additional text or characters.

    Text: \textbf{INPUT TEXT} Output:
    \end{promptbot}
    \caption{\textsc{0s} system and user prompts.}\label{prompt:0s}
\end{figure*}

\section{Dataset information}\label{supp:data}


Aggregated statistics for all datasets are presented in Table \ref{tab:data_statistics}. In the PunEval dataset, we initially identified annotation issues by locating puns where the annotated $w_p$ did not appear in the text. We detected 13 such issues, along with several typos, and addressed them by re-annotating the examples. The PunEval dataset also includes a list of contextual words that support both senses and a human explanation. Additionally, we removed contextual words that did not appear in the text (consistency errors). While the contextual words and human explanations are not utilized in this study, we have incorporated all these refinements in the published data splits.

\begin{table*}[htp]
  \small
  \centering
  \begin{tabular}{lcc|cccc}
    \toprule
    \multirow{2}{*}{\textbf{Dataset}} & \multicolumn{2}{c}{\textbf{Avg. Statistics}} & \multicolumn{4}{c}{\textbf{Labels Count}} \\
    \cmidrule{2-3}\cmidrule{4-7}
    & \textbf{N. chars} & \textbf{N. words} & \textbf{Heterographic puns} & \textbf{Homographic puns} & \textbf{Non-puns} & \textbf{Total samples} \\
    \midrule

    PunEval (train) & 58 & 13 & 279 & 304 & 488 & 1071 \\
    PunEval (test) & 58 & 13 & 312 & 461 & 568 & 1341 \\
    PunEval (val) & 59 & 13 & 56 & 46 & 75 & 177 \\
    \textbf{\NAPData} & 67 & 16 & 64 & 64 & 128 & 256 \\
    JOKER & 64 & 14 & 381 & 251 & 632 & 1264 \\
    \textbf{\UnbiasedData} & 64 & 15 & 212 & 388 & 600 & 1200 \\
    \textbf{\SubstitutionData} & 66 & 15 & 100 & 100 & 900 & 1100 \\
    
    \bottomrule
\end{tabular}
\caption{Dataset statistics. Datasets in bold represent the new ones introduced in this work. \textit{N. chars} and \textit{N. words} denote the average number of characters and words per sample, respectively.}\label{tab:data_statistics}
\end{table*}

\subsection{\UnbiasedData}\label{supp:unbiased}

To select common pun patterns for the \UnbiasedData, we use scikit-learn\footnote{\url{https://scikit-learn.org/stable/index.html}} to train a logistic regressor (LR) on the PunEval training split and then infer the target labels on the testing split, using default parameters. The LR coefficients with higher absolute values are deemed more significant for the outcome. We train four separate LRs with n-gram features ($n \in \{1-4\}$), imposing a minimum frequency threshold for each feature (10 for unigrams and 3 for the others). In addition, we compute the percentage difference between the occurrences of each feature in puns and non-puns, as we seek features that effectively differentiate the two sets. The final feature score is calculated by multiplying this percentage difference by the absolute value of the corresponding coefficient, and the top 20 features are then manually reviewed.

We observed that n-grams belonging to the same pattern often appeared as separate features (e.g., both ``\textit{she was only}'' and ``\textit{daughter}'' are part of the pattern ``\textit{she was only a [...] daughter but}''). Consequently, we retained the 6 most representative patterns for each feature. Some features were discarded because they did not represent a genuine pattern (for example, the expression ``\textit{lot of}'' appeared only 8 times and likely reflects a spurious correlation rather than a common pun construction).

After selecting the patterns, we sampled all puns containing these language patterns from the PunEval dataset. Missing examples were collected online and generated via GPT-4o using the prompt shown in Fig. \ref{prompt:generation_prompts}, with the temperature set to 1.0. All generated examples were manually reviewed, and those that were poorly crafted were either revised or discarded. We queried the model multiple times until we obtained 100 puns for each pattern.

\begin{figure*}
    \footnotesize

    \begin{prompt}[title={Non-puns generation}]
    \textit{/* Definition */}

    Puns are a form of wordplay that exploits multiple meanings of a word or similar-sounding words to produce a humorous effect.
    Puns typically involve a clever juxtaposition of words or phrases, where the humor arises from the interplay of their meanings or sounds. In contrast, non-puns are jokes or statements that do not rely on such linguistic ambiguities.

    \textit{/* Instructions */}

    You are a creative linguist who thinks out-of-the-box.
    You receive a JSON-formatted string in the format {{"expression": "EXPR1 [X] EXPR2", "count": C}} where [X] represents a missing expression, and you must generate C pairs of distinct short paragraphs where:
    - the first text in the pair must be a short paragraph containing a pun and respect the template given in "expression";
    - the second text in the pair must be a rephrased version of the first one, that removes the pun (i.e., a non-pun) but keeps a similar meaning.
    You can add words before or after the template, but you must ensure that each generated paragraph contains the given template with proper substitutions.
    You must format each pair in JSON format: {{"pun": "[generated pun]", "non-pun": "[generated non-pun]"}}.
    You must make sure that field "pun" contains a good pun, and "non-pun" contains no pun at all.
    You must write each JSON-formatted pair in a new line, respecting the JSONL format. Do not add any superfluous text and report a JSON-formatted string in each line.

    \textit{/* Examples */}

    INPUT: {"expression": "Old [X] never die, they just", "count": 3}

    OUTPUT:
    {"pun": "Old wizards never die, they just improve their spelling.", "non-pun": "Old wizards never die, they just improve their pronunciation."}
    {"pun": "Old squirrels never die, they just go nuts!", "non-pun": "Old squirrels never die, they just go crazy!"}
    {"pun": "Old crabs never die, they just became a little more shellfish!", "non-pun": "Old crabs never die, they just became a little more egocentric!"}

    INPUT: {"expression": "is", "count": 2}
    
    OUTPUT:
    {"pun": "Denial is a river in Egypt.", "non-pun": "The river Nile is a river in Egypt."}
    {"pun": "The cyclist is two tired to win", "non-pun": "The cyclist is too drained to win"}

    INPUT: {"expression": "This [X] always", "count": 2}

    OUTPUT:
    {"pun": "This vacuum always sucks!", "non-pun": "This vacuum always disappoints!"}
    {"pun": "This candy cane always remains in mint condition.", "non-pun": "This candy cane always remains in good condition."}
    
    INPUT: \textbf{INPUT EXPRESSION}
    
    OUTPUT:    
    \end{prompt}
    \caption{Prompt used to generate puns and non-puns containing language patterns.}\label{prompt:generation_prompts}
\end{figure*}

\subsection{\SubstitutionData}\label{supp:substitution_data}

To create the \SubstitutionData, we selected 100 heterographic puns and 100 homographic puns from the \NAPData\ and PunEval test split. We generated alternatives using GPT-4o-mini for \texttt{homophone} replacements and employed Google Translate to ensure that the substituted words sounded similar to the originals.
For \texttt{random} replacements, we compiled a list of random words and replaced them in the dataset, ensuring that these words did not fit contextually within the sentences. To generate synonyms (\texttt{alt syn} and \texttt{pun syn}), we utilized online thesauruses and dictionaries to find synonyms for each word; when synonyms were unavailable, we opted for a hypernym instead. We do our best to ensure that none of these substitutions result in accidental puns. 
For example, consider the pun ``\textit{\guillemotleft Consult an investment broker\guillemotright was Tom's \underline{stock} (stock) answer}'', which plays on \textit{stock} as both ``\textit{commonly used or brought forward}'' ($s_p$) and ``\textit{capital raised by a corporation representing ownership interest}'' ($s_a$). For the synonym substitutions, we replaced ``\textit{stock}'' with ``\textit{routine}'', which is roughly equivalent in context, and \textit{equity}, respectively. We purposely avoided using ``\textit{default}'' in the second substitution, as it carries an economic connotation.
As a nonsensical but phonetically similar expression, we used ``\textit{stork}'' (a bird) and ``\textit{yawn}'' as a random replacement.

\section{Additional Results}
\label{supp:results}

Table \ref{tab:hom_het_recall} reports recall over all prompts and the dataset, divided by heterographic and homographic puns.

\begin{table*}[tbp]
    \small
    \centering
    \begin{tabular}{llcccccccc}
    \toprule
    \multirow{2}{*}{\textbf{Model}} & \multirow{2}{*}{\textbf{Prompt}} & \multicolumn{2}{c}{\textbf{\NAPData}} & \multicolumn{2}{c}{\textbf{JOKER}} & \multicolumn{2}{c}{\textbf{PunEval}} & \multicolumn{2}{c}{\textbf{\UnbiasedData}} \\
    \cmidrule(lr){3-4} \cmidrule(lr){5-6} \cmidrule(lr){7-8} \cmidrule(lr){9-10}
    & & \textbf{hom} & \textbf{het} & \textbf{hom} & \textbf{het} & \textbf{hom} & \textbf{het} & \textbf{hom} & \textbf{het} \\
    \midrule
    \multirow{4}{*}{Gemini2.0} 
    & \textsc{0s}  & 100.0 {\scriptsize$\pm$0.0} & 100.0 {\scriptsize$\pm$0.0} & 96.0 {\scriptsize$\pm$0.0} & 95.9 {\scriptsize$\pm$0.2} & 97.3 {\scriptsize$\pm$0.1} & 92.3 {\scriptsize$\pm$0.0} & 97.2 {\scriptsize$\pm$3.6} & 95.4 {\scriptsize$\pm$6.0} \\
    & \textsc{fs}  & 96.9 {\scriptsize$\pm$0.0} & 98.4 {\scriptsize$\pm$0.0} & 92.4 {\scriptsize$\pm$0.0} & 93.6 {\scriptsize$\pm$0.2} & 95.3 {\scriptsize$\pm$0.1} & 88.3 {\scriptsize$\pm$0.2} & 97.1 {\scriptsize$\pm$3.1} & 96.2 {\scriptsize$\pm$5.7} \\
    & \textsc{w}   & 94.3 {\scriptsize$\pm$0.9} & 97.9 {\scriptsize$\pm$0.9} & 90.4 {\scriptsize$\pm$0.0} & 94.0 {\scriptsize$\pm$0.0} & 94.3 {\scriptsize$\pm$0.1} & 86.9 {\scriptsize$\pm$0.0} & 92.0 {\scriptsize$\pm$8.7} & 95.3 {\scriptsize$\pm$7.6} \\
    & \textsc{w+s} & 98.4 {\scriptsize$\pm$0.0} & 98.4 {\scriptsize$\pm$0.0} & 87.7 {\scriptsize$\pm$0.0} & 91.9 {\scriptsize$\pm$0.0} & 94.0 {\scriptsize$\pm$0.1} & 85.7 {\scriptsize$\pm$0.4} & 91.5 {\scriptsize$\pm$8.0} & 91.4 {\scriptsize$\pm$10.3} \\
    \midrule
    \multirow{4}{*}{GPT-4o} 
    & \textsc{0s}  & 99.0 {\scriptsize$\pm$0.9} & 100.0 {\scriptsize$\pm$0.0} & 95.2 {\scriptsize$\pm$0.0} & 96.3 {\scriptsize$\pm$0.0} & 96.4 {\scriptsize$\pm$0.3} & 93.0 {\scriptsize$\pm$0.6} & 98.2 {\scriptsize$\pm$2.3} & 98.3 {\scriptsize$\pm$2.3} \\
    & \textsc{fs}  & 97.4 {\scriptsize$\pm$0.9} & 98.4 {\scriptsize$\pm$0.0} & 93.6 {\scriptsize$\pm$0.0} & 94.6 {\scriptsize$\pm$0.2} & 94.2 {\scriptsize$\pm$0.3} & 88.9 {\scriptsize$\pm$0.4} & 96.9 {\scriptsize$\pm$3.5} & 96.9 {\scriptsize$\pm$2.9} \\
    & \textsc{w}   & 84.9 {\scriptsize$\pm$2.4} & 98.4 {\scriptsize$\pm$0.0} & 79.9 {\scriptsize$\pm$0.3} & 92.7 {\scriptsize$\pm$0.4} & 77.7 {\scriptsize$\pm$3.0} & 80.3 {\scriptsize$\pm$0.7} & 86.2 {\scriptsize$\pm$8.3} & 91.3 {\scriptsize$\pm$9.8} \\
    & \textsc{w+s} & 87.5 {\scriptsize$\pm$1.6} & 97.9 {\scriptsize$\pm$0.9} & 85.9 {\scriptsize$\pm$0.3} & 93.0 {\scriptsize$\pm$0.2} & 81.1 {\scriptsize$\pm$1.9} & 82.8 {\scriptsize$\pm$1.4} & 88.2 {\scriptsize$\pm$7.2} & 94.7 {\scriptsize$\pm$6.1} \\
    \midrule
    \multirow{4}{*}{Llama3.3} 
    & \textsc{0s}  & 96.9 {\scriptsize$\pm$0.0} & 99.5 {\scriptsize$\pm$0.9} & 92.8 {\scriptsize$\pm$0.0} & 94.2 {\scriptsize$\pm$0.4} & 94.6 {\scriptsize$\pm$1.3} & 86.7 {\scriptsize$\pm$2.5} & 97.3 {\scriptsize$\pm$3.8} & 94.4 {\scriptsize$\pm$6.7} \\
    & \textsc{fs}  & 98.4 {\scriptsize$\pm$0.0} & 97.4 {\scriptsize$\pm$2.4} & 92.6 {\scriptsize$\pm$0.3} & 93.3 {\scriptsize$\pm$0.6} & 94.9 {\scriptsize$\pm$0.3} & 88.1 {\scriptsize$\pm$1.1} & 97.0 {\scriptsize$\pm$4.2} & 96.2 {\scriptsize$\pm$4.7} \\
    & \textsc{w}   & 90.1 {\scriptsize$\pm$0.9} & 93.2 {\scriptsize$\pm$1.8} & 87.9 {\scriptsize$\pm$0.3} & 90.0 {\scriptsize$\pm$0.0} & 89.4 {\scriptsize$\pm$0.6} & 82.2 {\scriptsize$\pm$0.7} & 95.1 {\scriptsize$\pm$4.3} & 92.7 {\scriptsize$\pm$5.1} \\
    & \textsc{w+s} & 96.9 {\scriptsize$\pm$0.0} & 97.4 {\scriptsize$\pm$0.9} & 91.8 {\scriptsize$\pm$0.3} & 92.0 {\scriptsize$\pm$0.6} & 92.6 {\scriptsize$\pm$0.7} & 83.0 {\scriptsize$\pm$0.6} & 93.7 {\scriptsize$\pm$7.5} & 93.9 {\scriptsize$\pm$7.7} \\
    \midrule
    \multirow{4}{*}{Mistral3} 
    & \textsc{0s}  & 93.2 {\scriptsize$\pm$0.9} & 80.7 {\scriptsize$\pm$0.9} & 79.5 {\scriptsize$\pm$0.3} & 86.0 {\scriptsize$\pm$0.6} & 80.3 {\scriptsize$\pm$1.2} & 68.7 {\scriptsize$\pm$0.8} & 87.8 {\scriptsize$\pm$12.4} & 87.9 {\scriptsize$\pm$16.3} \\
    & \textsc{fs}  & 88.0 {\scriptsize$\pm$3.3} & 87.0 {\scriptsize$\pm$9.2} & 84.7 {\scriptsize$\pm$1.4} & 86.6 {\scriptsize$\pm$4.5} & 81.1 {\scriptsize$\pm$1.1} & 81.8 {\scriptsize$\pm$4.0} & 88.3 {\scriptsize$\pm$3.9} & 86.2 {\scriptsize$\pm$5.3} \\
    & \textsc{w}   & 88.0 {\scriptsize$\pm$3.9} & 96.9 {\scriptsize$\pm$3.1} & 91.4 {\scriptsize$\pm$0.8} & 94.6 {\scriptsize$\pm$0.6} & 84.0 {\scriptsize$\pm$1.2} & 82.5 {\scriptsize$\pm$0.5} & 92.2 {\scriptsize$\pm$6.0} & 90.8 {\scriptsize$\pm$8.2} \\
    & \textsc{w+s} & 95.3 {\scriptsize$\pm$1.6} & 96.4 {\scriptsize$\pm$2.4} & 95.6 {\scriptsize$\pm$0.6} & 97.6 {\scriptsize$\pm$0.4} & 93.2 {\scriptsize$\pm$1.2} & 92.1 {\scriptsize$\pm$0.8} & 97.6 {\scriptsize$\pm$1.6} & 97.5 {\scriptsize$\pm$4.1} \\
    \midrule
    \multirow{4}{*}{Qwen2.5} 
    & \textsc{0s}  & 100.0 {\scriptsize$\pm$0.0} & 100.0 {\scriptsize$\pm$0.0} & 92.8 {\scriptsize$\pm$0.6} & 96.5 {\scriptsize$\pm$0.2} & 96.1 {\scriptsize$\pm$0.4} & 88.6 {\scriptsize$\pm$0.4} & 97.1 {\scriptsize$\pm$4.2} & 95.2 {\scriptsize$\pm$11.2} \\
    & \textsc{fs}  & 95.8 {\scriptsize$\pm$0.9} & 96.4 {\scriptsize$\pm$0.9} & 84.5 {\scriptsize$\pm$1.7} & 91.3 {\scriptsize$\pm$0.0} & 87.6 {\scriptsize$\pm$0.1} & 78.5 {\scriptsize$\pm$0.3} & 92.2 {\scriptsize$\pm$7.5} & 91.6 {\scriptsize$\pm$13.1} \\
    & \textsc{w}   & 94.8 {\scriptsize$\pm$0.9} & 93.8 {\scriptsize$\pm$1.6} & 80.7 {\scriptsize$\pm$0.8} & 86.5 {\scriptsize$\pm$0.2} & 85.0 {\scriptsize$\pm$0.6} & 75.1 {\scriptsize$\pm$0.7} & 88.1 {\scriptsize$\pm$11.8} & 87.6 {\scriptsize$\pm$17.0} \\
    & \textsc{w+s} & 94.3 {\scriptsize$\pm$0.9} & 94.3 {\scriptsize$\pm$0.9} & 82.1 {\scriptsize$\pm$1.7} & 87.8 {\scriptsize$\pm$0.2} & 87.8 {\scriptsize$\pm$0.8} & 75.8 {\scriptsize$\pm$1.0} & 88.5 {\scriptsize$\pm$12.3} & 88.4 {\scriptsize$\pm$16.3} \\
    \midrule
    \multirow{4}{*}{R1-D} 
    & \textsc{0s}  & 100.0 {\scriptsize$\pm$0.0} & 99.0 {\scriptsize$\pm$0.9} & 97.8 {\scriptsize$\pm$0.3} & 98.3 {\scriptsize$\pm$0.2} & 96.8 {\scriptsize$\pm$0.3} & 96.6 {\scriptsize$\pm$0.4} & 98.4 {\scriptsize$\pm$1.9} & 99.0 {\scriptsize$\pm$2.1} \\
    & \textsc{fs}  & 98.4 {\scriptsize$\pm$1.6} & 96.9 {\scriptsize$\pm$2.7} & 96.4 {\scriptsize$\pm$1.7} & 96.8 {\scriptsize$\pm$0.4} & 96.8 {\scriptsize$\pm$0.2} & 93.3 {\scriptsize$\pm$1.2} & 97.0 {\scriptsize$\pm$3.4} & 98.0 {\scriptsize$\pm$3.3} \\
    & \textsc{w}   & 93.7 {\scriptsize$\pm$2.7} & 99.0 {\scriptsize$\pm$0.9} & 91.8 {\scriptsize$\pm$0.3} & 96.6 {\scriptsize$\pm$1.1} & 91.9 {\scriptsize$\pm$1.1} & 91.8 {\scriptsize$\pm$1.0} & 92.6 {\scriptsize$\pm$3.5} & 97.0 {\scriptsize$\pm$3.6} \\
    & \textsc{w+s} & 94.8 {\scriptsize$\pm$0.9} & 99.0 {\scriptsize$\pm$1.8} & 92.6 {\scriptsize$\pm$2.0} & 94.4 {\scriptsize$\pm$0.2} & 93.2 {\scriptsize$\pm$1.0} & 87.9 {\scriptsize$\pm$1.0} & 93.9 {\scriptsize$\pm$5.2} & 95.9 {\scriptsize$\pm$4.8} \\
    \midrule
    \multirow{4}{*}{R1} 
    & \textsc{0s}  & 100.0 {\scriptsize$\pm$0.0} & 100.0 {\scriptsize$\pm$0.0} & 99.0 {\scriptsize$\pm$0.3} & 99.0 {\scriptsize$\pm$0.4} & 99.3 {\scriptsize$\pm$0.1} & 99.2 {\scriptsize$\pm$0.2} & 99.7 {\scriptsize$\pm$0.5} & 99.0 {\scriptsize$\pm$1.6} \\
    & \textsc{fs}  & 99.5 {\scriptsize$\pm$0.9} & 99.0 {\scriptsize$\pm$0.9} & 97.8 {\scriptsize$\pm$0.3} & 97.9 {\scriptsize$\pm$0.4} & 98.9 {\scriptsize$\pm$0.1} & 98.0 {\scriptsize$\pm$0.5} & 98.8 {\scriptsize$\pm$0.9} & 98.3 {\scriptsize$\pm$1.6} \\
    & \textsc{w}   & 98.4 {\scriptsize$\pm$0.0} & 100.0 {\scriptsize$\pm$0.0} & 98.4 {\scriptsize$\pm$0.0} & 98.6 {\scriptsize$\pm$0.2} & 98.8 {\scriptsize$\pm$0.3} & 98.5 {\scriptsize$\pm$0.2} & 99.2 {\scriptsize$\pm$1.1} & 99.1 {\scriptsize$\pm$1.1} \\
    & \textsc{w+s} & 99.0 {\scriptsize$\pm$0.9} & 100.0 {\scriptsize$\pm$0.0} & 97.0 {\scriptsize$\pm$0.9} & 97.9 {\scriptsize$\pm$0.4} & 98.4 {\scriptsize$\pm$0.1} & 97.4 {\scriptsize$\pm$0.9} & 98.3 {\scriptsize$\pm$1.2} & 98.7 {\scriptsize$\pm$1.7} \\
    \bottomrule
\end{tabular}
    \caption{Recall (puns) on the subsets of heterographic (het) and homographic (hom) puns for each dataset.}
    \label{tab:hom_het_recall}
\end{table*}

\subsection{Effect of rationale}\label{supp:rationale-effect}

Fig. \ref{fig:boxplot-base-joker} highlights the impact of rationale-augmented prompts for all models on the \NAPData\ and JOKER datasets.
Fig. \ref{fig:rq2.1-bias-rationale} compares the impact of the best rationale-augmented prompt (\textsc{w} or \textsc{w+s}) on pun detection performance over the \UnbiasedData\ dataset, excluding the \texttt{tom} and \texttt{when} patterns that do not consistently bias the tested LLMs. Fig. \ref{fig:alterations_rationale} shows the same results on the \SubstitutionData, when considering only the four substitution sets (800 negative examples). The best prompt used is the one highlighted in Table \ref{tab:rq2.1-results-complete}, except for Mistral3, where we select the \textsc{w} prompt as it appears to work better overall. In the violin plots, the right side (light blue) represents the results distributions over the three runs with the rationale-augmented prompt, while the left side (red) represents performance using the \textsc{fs} prompt --- typically the best prompt without rationales. Note that in all cases except Mistral, incorporating rationalization reduces the number of false positives, thereby lowering recall and improving precision, countering the bias effect.

\subsection{RQ1}

\begin{figure*}[hbpt]
  \includegraphics[width=\textwidth]{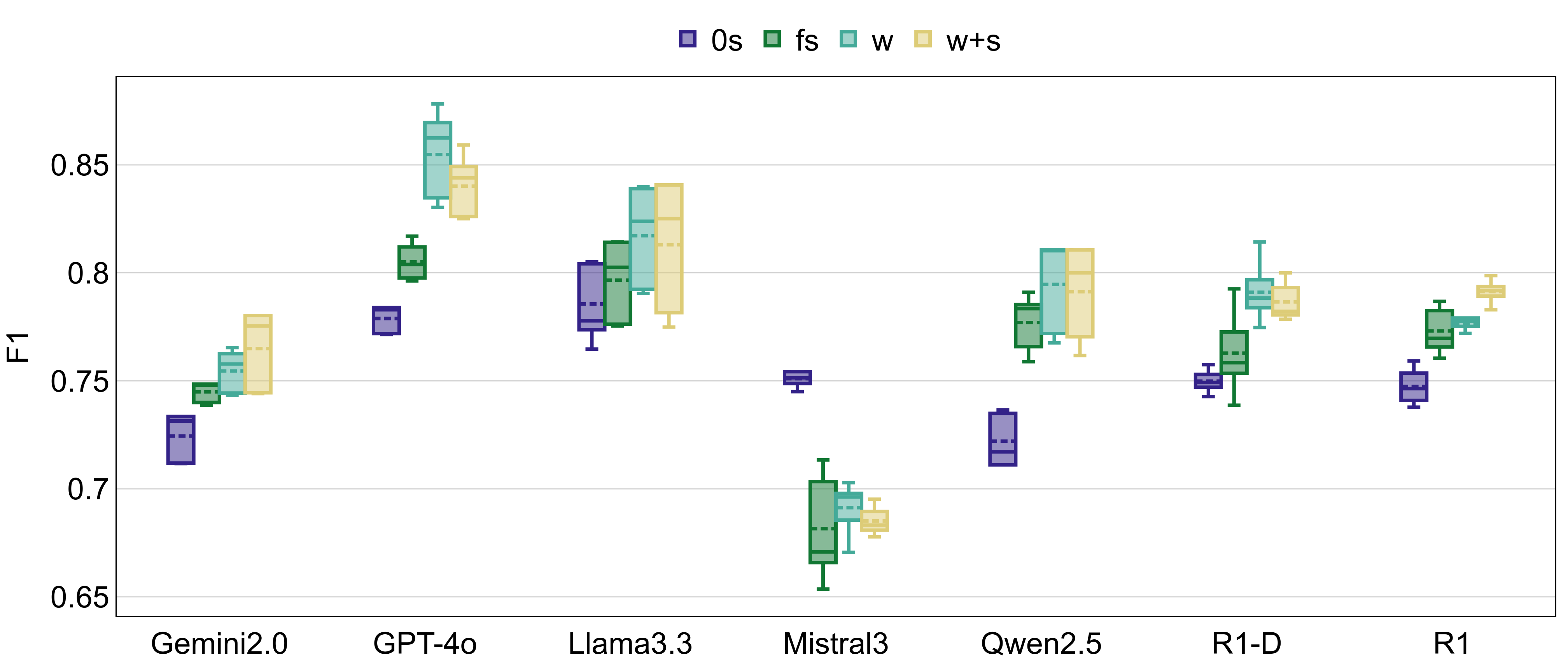}
  \caption{Distribution of F1-score over 3 runs on the \NAPData\ and JOKER datasets. The dotted line represents the mean, while the solid line is the median.}
  \label{fig:boxplot-base-joker}
\end{figure*}

\begin{figure*}[hbpt]
  \includegraphics[width=\textwidth]{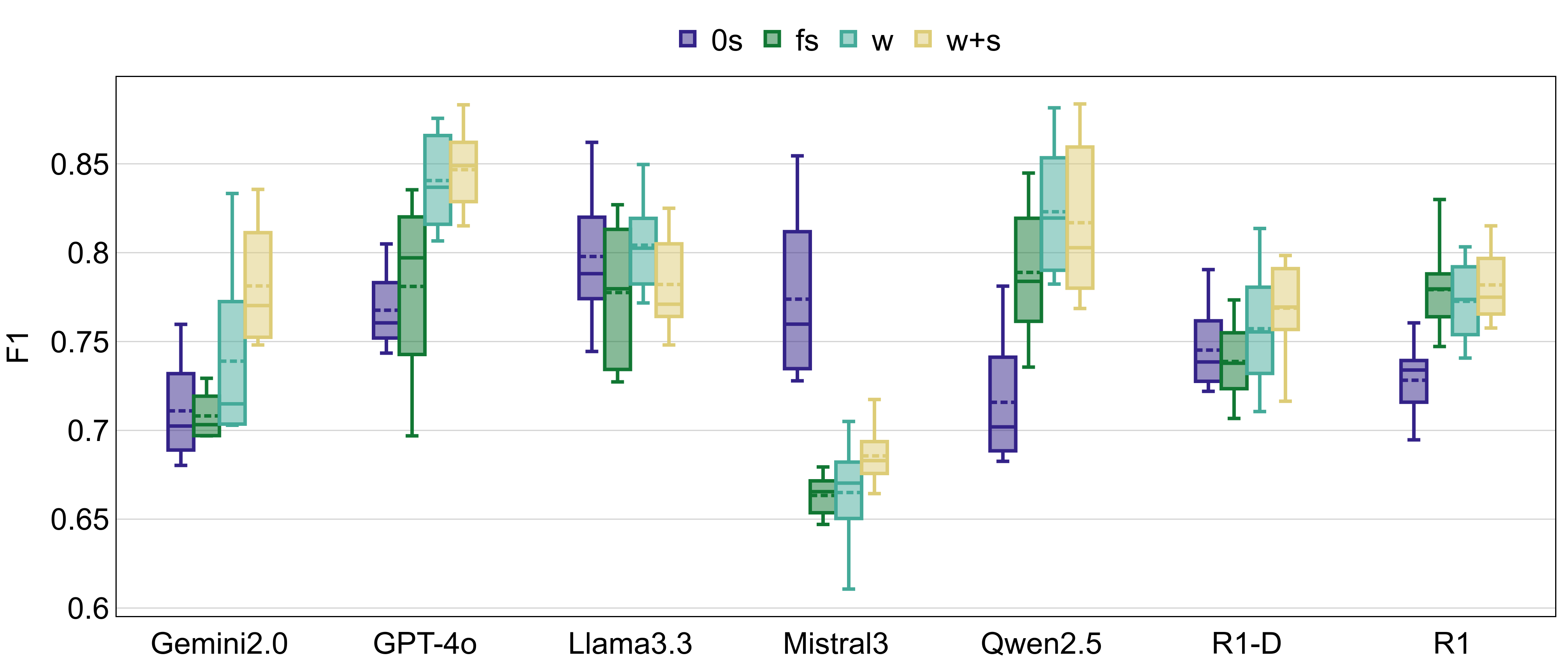}
  \caption{Distribution of F1-score over 3 runs on the \UnbiasedData. The dotted line represents the mean, while the solid line is the median.}
  \label{fig:boxplot-unbiased}
\end{figure*}

Fig. \ref{fig:boxplot-base-joker} shows the difference in performance for each prompt used. Fig. \ref{fig:boxplot-unbiased} shows the same effect over the \UnbiasedData\ dataset.
This shows that, overall, the \textsc{w} or \textsc{w+s} prompt improves performance over the simpler prompts.

\subsection{RQ2: Pun Language Patterns}
\label{supp:rq2.1}

\paragraph{Pattern identification.} Table \ref{tab:semeval_issues} lists the six language patterns identified as commonly used to create puns. Table \ref{tab:pattern_occurrences} summarizes the statistics for these patterns in our datasets. We omit data for the \UnbiasedData\ dataset, as all examples in it are designed to contain a language pattern.

\begin{table*}[ht!]
  \small
  \begin{tabularx}{\textwidth}{llXHH}
    \toprule
    \textbf{Structure} & \textbf{Short name} &\textbf{Example} & \textbf{\#train} & \textbf{\#test} \\
    \midrule
    \arrayrulecolor{white}
    Old [...] never die [...] they & \texttt{never\_die} & OLD BANKERS never die they just lose interest & 67 & 62 \\\hline
    Tom & \texttt{tom} & ``I've been to a film festival in Southern France'' said Tom cannily & 61 & 61  \\\hline
    When the & \texttt{when} & When the TV repairman got married the reception was excellent & 13 & 20 \\\hline
    She was only [...] daughter but & \texttt{daughter} & She was only a horseman's daughter, but she didn't know how to say neigh & 13 & 19 \\\hline
    Doctor, Doctor & \texttt{doctor} & Doctor, Doctor, I keep thinking I'm a spoon. - Sit there and don't stir & 3 & 7 \\ \hline
    used to [...] but & \texttt{used} & I used to be addicted to soap, but I'm clean now & 7 & 4 \\\arrayrulecolor{black}\bottomrule
  \end{tabularx}
  \caption{Pun language patterns and their occurrences in the PunEval dataset.}
  \label{tab:semeval_issues}
\end{table*}
\begin{table*}[ht!]
  \small
  \centering
  \begin{tabular}{lcccccc}
    \toprule
    \multirow{2}{*}{\textbf{Pattern}} & \multicolumn{5}{c}{\textbf{Occurrences}} \\
    \cmidrule{2-7}
    &\makecell{\textbf{PunEval} \\ \textbf{(train)}}  & \makecell{\textbf{PunEval} \\ \textbf{(test)}} & \makecell{\textbf{PunEval} \\ \textbf{(val)}} & \textbf{\NAPData} & \textbf{JOKER} & \textbf{\SubstitutionData} \\
    \midrule
    \texttt{never\_die}  & 65 & 62 & 1 & 2 & 12 & 50 \\
    \texttt{tom}         & 61 & 61 & 1 & 0 & 0 & 15 \\
    \texttt{when}        & 13&20&0&0&22&5 \\
    \texttt{daughter}    &11&19&1&0&0&15\\
    \texttt{doctor}      &3&7&0&0&0&0\\
    \texttt{used}        &2&4&1&2&10&10\\ 
    \bottomrule
  \end{tabular}
  \caption{Occurrences of pun language patterns in all the datasets.}\label{tab:pattern_occurrences}
\end{table*}

\begin{figure}[htbp]
  \includegraphics[width=\columnwidth]{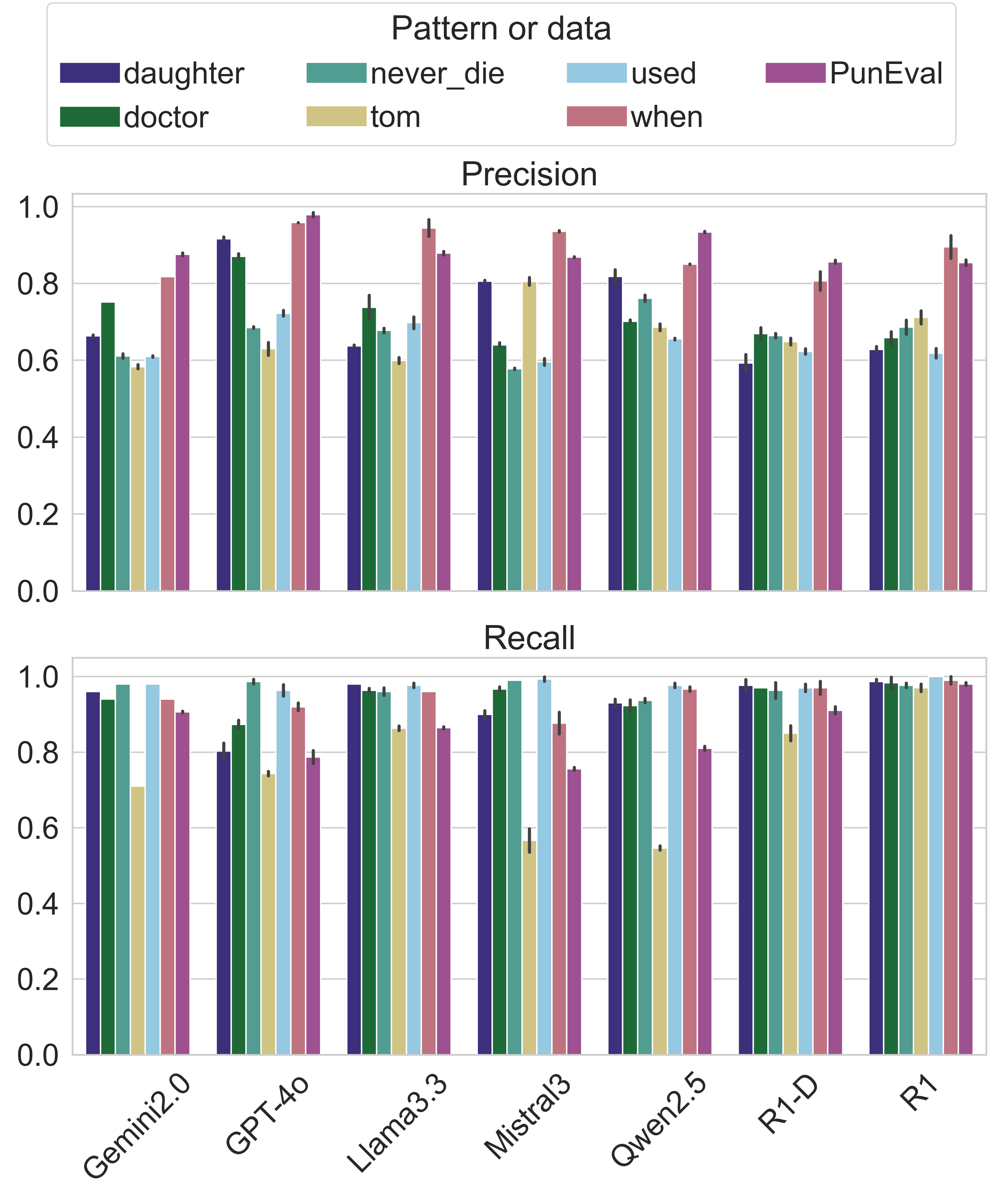}
  \caption{Precision and recall on the \UnbiasedData\ dataset using the best prompt (all patterns).}
  \label{fig:rq2.1-bias-plots}
\end{figure}

\begin{figure}[htbp]
  \includegraphics[width=\columnwidth]{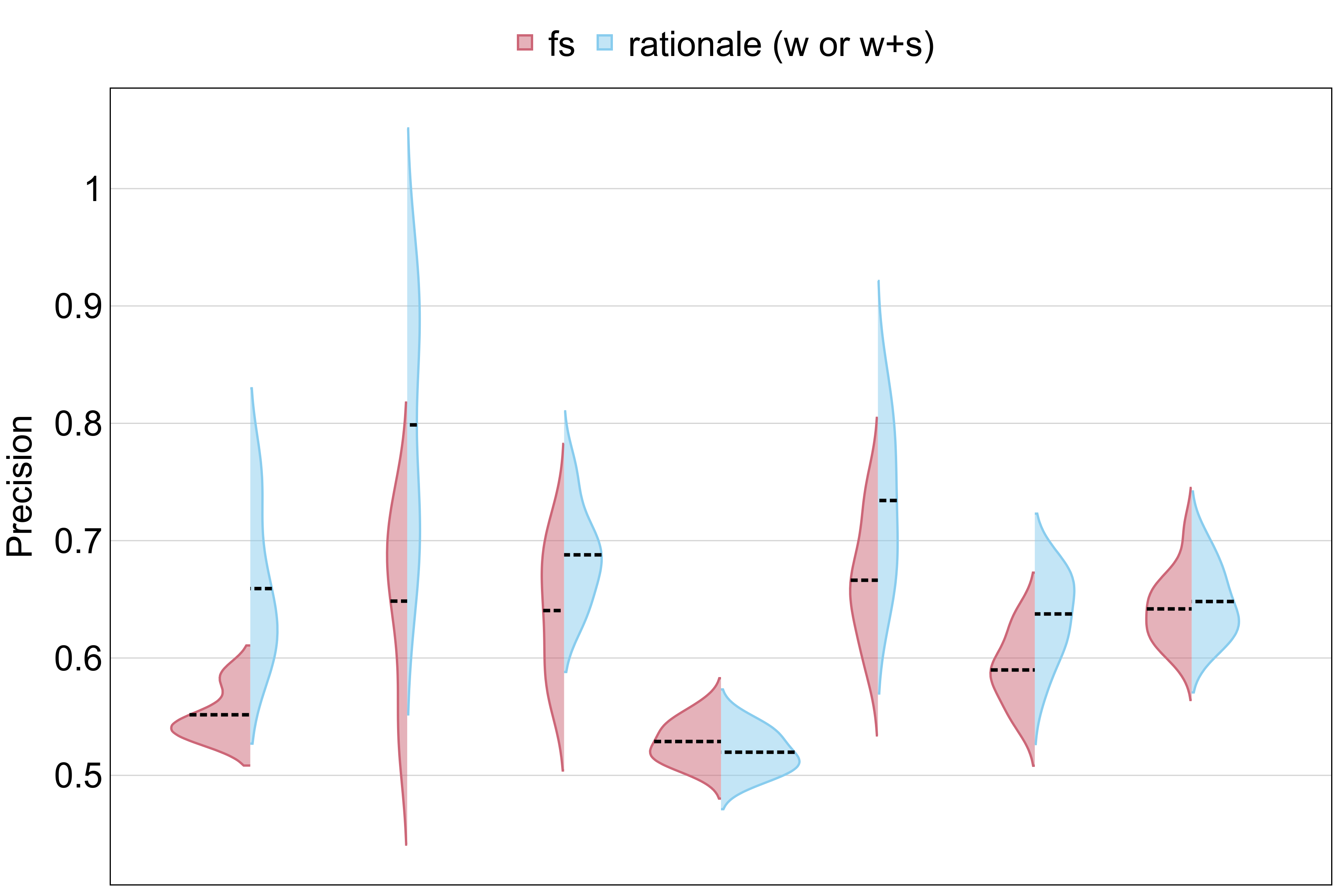}\hfill
  \includegraphics[width=\columnwidth]{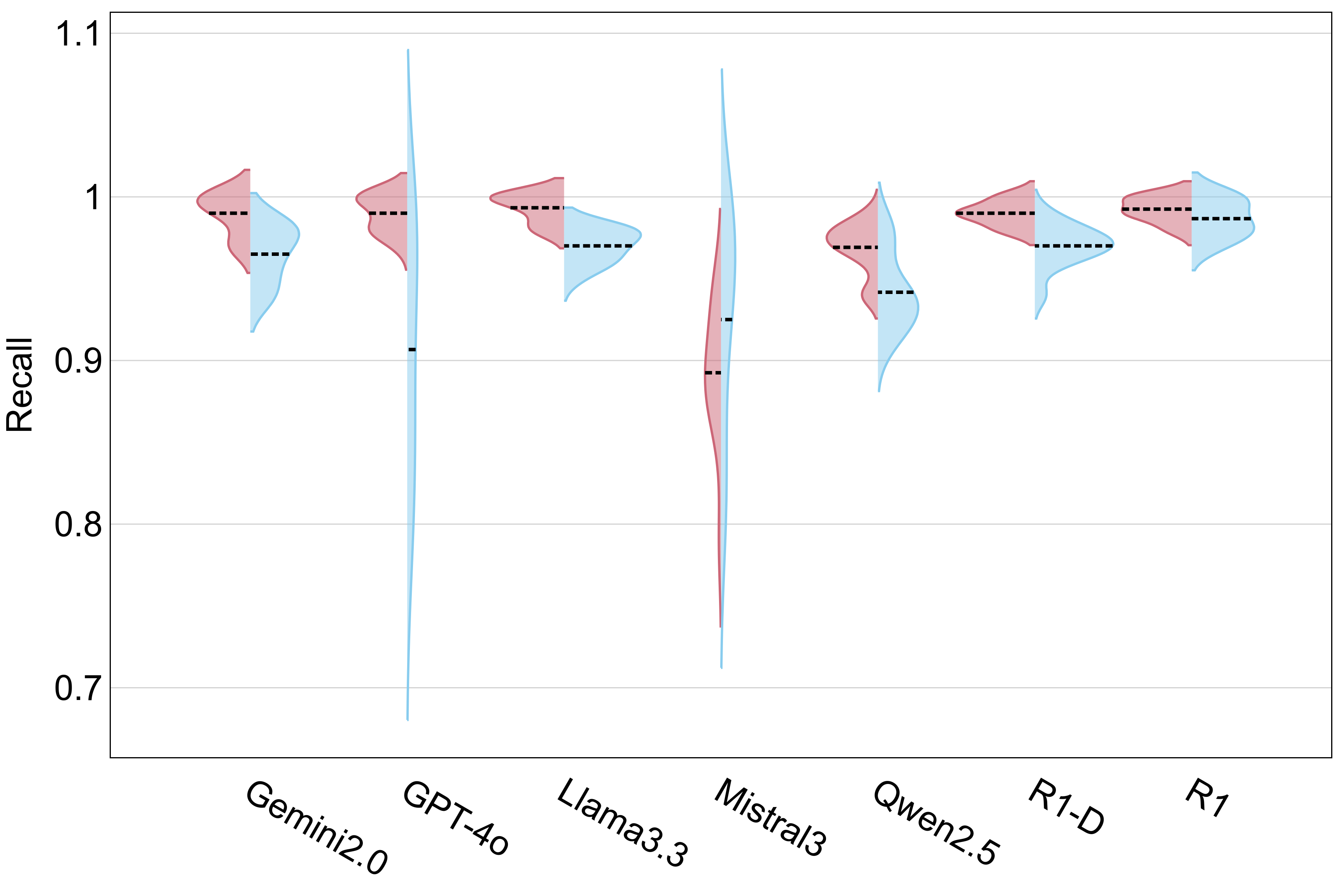}
  \caption{Precision and recall over the \UnbiasedData\ dataset using the few-shot (\texttt{fs}) and the best prompt between \texttt{w} and \texttt{w+s}.}
  \label{fig:rq2.1-bias-rationale}
\end{figure}

\begin{figure}[htb]
  \includegraphics[width=\columnwidth]{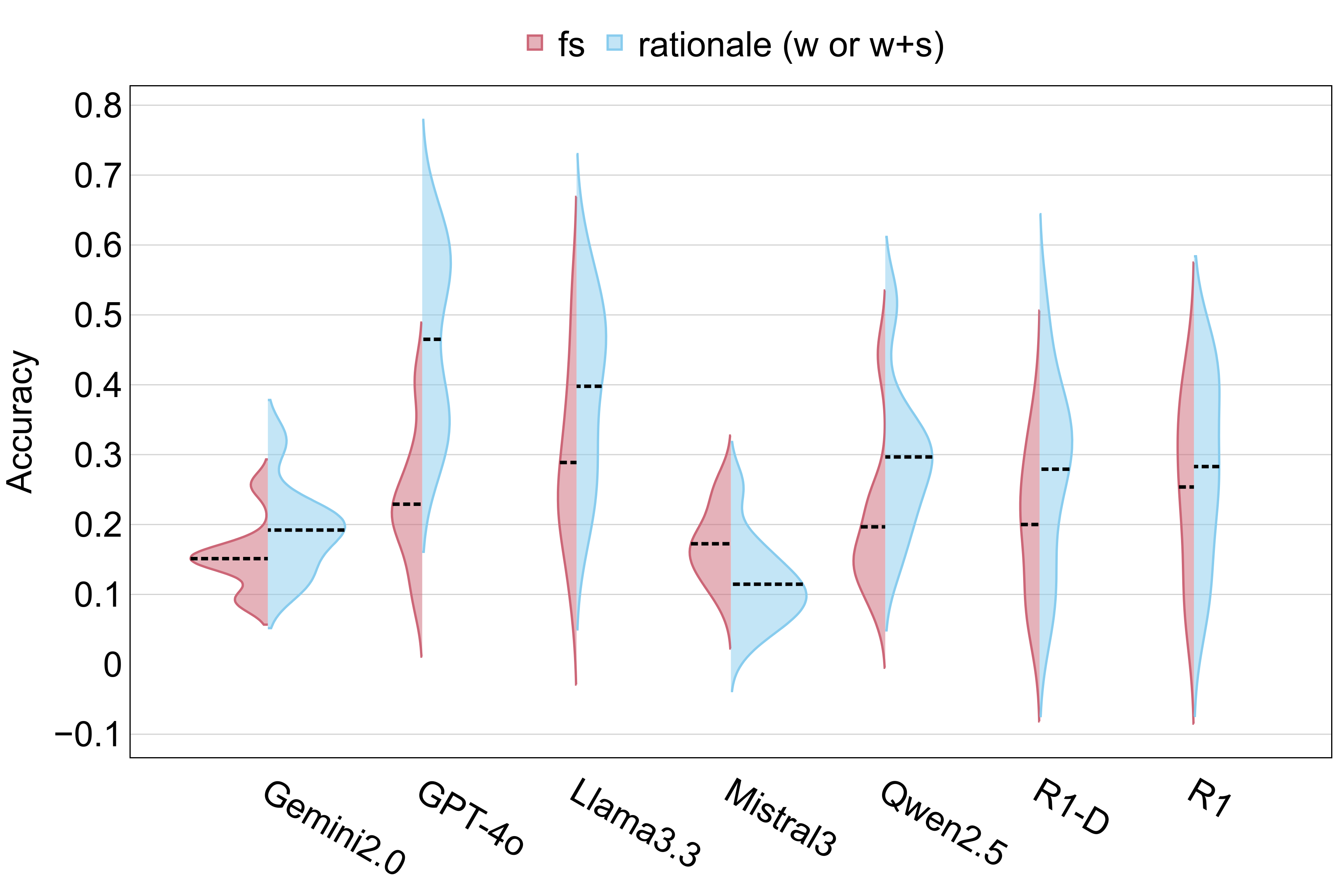}
  \caption{Accuracy over the \SubstitutionData\ dataset (only negative examples) using the few-shot (\texttt{fs}) and the best prompt between \texttt{w} and \texttt{w+s}.}
  \label{fig:alterations_rationale}
\end{figure}

\paragraph{Performance.} Table \ref{tab:rq2.1-results-complete} compares the results obtained using each prompt for PunEval and \UnbiasedData. Note that the best prompts (in bold) are the same as those that performed best in our RQ1 (Table \ref{tab:rq1-results}). Fig. \ref{fig:rq2.1-bias-plots} visualizes the performance differences for each set of expressions containing language patterns. We interpret bias as being present in models with very high recall but low precision, indicating they tend to classify all instances containing the pattern as puns.

We observe a sharp imbalance of low precision and high recall for the \texttt{never\_die}, \texttt{used}, and \texttt{doctor} patterns (Fig. \ref{fig:rq2.1-bias-plots}), indicating that these patterns tend to mislead LLMs into identifying puns. In contrast, the \texttt{tom} and \texttt{when} patterns do not appear to consistently affect them.
While all LLMs tend to produce many false positives with the \texttt{used}, \texttt{never\_die}, and \texttt{doctor} patterns, results for the other patterns are more varied. For instance, GPT-4o, Qwen2.5, and Mistral3 are more resilient to the \texttt{daughter} pattern compared to the other models. Interestingly, examples containing the word ``Tom'' produce both false positives and negatives, suggesting that this pattern is unlikely to serve as a reliable shortcut for classifying puns. 
In contrast, sentences beginning with ``When the'' are classified more reliably by all models.

\begin{table*}[htbp]
  \small
  \begin{tabularx}{\textwidth}{llcccccccc}
    \toprule
    \multirow{2}{*}{\textbf{Model}} 
    & \multirow{2}{*}{\textbf{\begin{turn}{90}Prom.\end{turn}}} 
    & \multicolumn{2}{c}{\textbf{F1}} 
    & \multicolumn{2}{c}{\textbf{Precision}} 
    & \multicolumn{2}{c}{\textbf{Recall}}
    & \multicolumn{2}{c}{\textbf{Accuracy}} \\
    \cmidrule(lr){3-10}
    
      & 
      & \textbf{PP} 
      & \textbf{PE} 
      & \textbf{PP} 
      & \textbf{PE} 
      & \textbf{PP} 
      & \textbf{PE}
      & \textbf{PP} 
      & \textbf{PE}  \\
    \midrule
    \multirow{4}{*}{Gemini2} & \textsc{0s}
     & 71.9 {\scriptsize $\pm$ 4.6} 
     & 85.7 {\scriptsize $\pm$ 0.2}
     & 57.6 {\scriptsize $\pm$ 5.4}
     & 78.0 {\scriptsize $\pm$ 0.2} 
     & 96.2 {\scriptsize $\pm$ 5.3} 
     & \textbf{95.3} {\scriptsize $\pm$ 0.1}
     & 62.2 {\scriptsize $\pm$ 7.6} 
     & 81.7 {\scriptsize $\pm$ 0.2} \\
         & \textsc{fs}
     & 73.0 {\scriptsize $\pm$ 6.4} 
     & \textbf{89.3} {\scriptsize $\pm$ 0.1}
     & 59.1 {\scriptsize $\pm$ 8.5} 
     & 86.3 {\scriptsize $\pm$ 0.1} 
     & \textbf{96.4} {\scriptsize $\pm$ 4.8} 
     & 92.5 {\scriptsize $\pm$ 0.2} 
     & 63.7 {\scriptsize $\pm$ 10.1} 
     & \textbf{87.2} {\scriptsize $\pm$ 0.1} \\
         & \textsc{w}
     & 74.7 {\scriptsize $\pm$ 6.6} 
     & 88.3 {\scriptsize $\pm$ 0.1}
     & 62.7 {\scriptsize $\pm$ 8.8} 
     & 85.5 {\scriptsize $\pm$ 0.2} 
     & 93.6 {\scriptsize $\pm$ 7.5} 
     & 91.3 {\scriptsize $\pm$ 0.1} 
     & 67.7 {\scriptsize $\pm$ 10.0} 
     & 86.0 {\scriptsize $\pm$ 0.1} \\
         & \textbf{\textsc{w+s}}
     & \textbf{76.9} {\scriptsize $\pm$ 7.5} 
     & 89.1 {\scriptsize $\pm$ 0.2} 
     & \textbf{66.7} {\scriptsize $\pm$ 8.6} 
     & \textbf{87.6} {\scriptsize $\pm$ 0.4} 
     & 91.6 {\scriptsize $\pm$ 9.5} 
     & 90.7 {\scriptsize $\pm$ 0.2}
     & \textbf{72.3} {\scriptsize $\pm$ 9.0} 
     & \textbf{87.2} {\scriptsize $\pm$ 0.2} \\
     \midrule
     
     \multirow{4}{*}{GPT-4o} & \textsc{0s}
     & 78.3 {\scriptsize $\pm$ 8.1} 
     & 89.7 {\scriptsize $\pm$ 0.1}
     & 65.8 {\scriptsize $\pm$ 12.2} 
     & 84.9 {\scriptsize $\pm$ 0.2} 
     & \textbf{98.3} {\scriptsize $\pm$ 2.2} 
     & \textbf{95.0} {\scriptsize $\pm$ 0.4} 
     & 71.8 {\scriptsize $\pm$ 11.6} 
     & 87.4 {\scriptsize $\pm$ 0.1} \\
       & \textsc{fs}
     & 78.8 {\scriptsize $\pm$ 9.0} 
     & \textbf{92.8} {\scriptsize $\pm$ 0.2}
     & 67.0 {\scriptsize $\pm$ 12.7} 
     & 93.5 {\scriptsize $\pm$ 0.3} 
     & 97.1 {\scriptsize $\pm$ 3.1} 
     & 92.1 {\scriptsize $\pm$ 0.2} 
     & 72.6 {\scriptsize $\pm$ 13.9} 
     & \textbf{91.8} {\scriptsize $\pm$ 0.2} \\
       & \textbf{\textsc{w}}
     & \textbf{83.1} {\scriptsize $\pm$ 7.8} 
     & 87.3 {\scriptsize $\pm$ 0.9} 
     & \textbf{79.7} {\scriptsize $\pm$ 12.3}
     & \textbf{97.9} {\scriptsize $\pm$ 0.5}
     & 88.1 {\scriptsize $\pm$ 8.6} 
     & 78.7 {\scriptsize $\pm$ 1.7}
     & \textbf{81.7} {\scriptsize $\pm$ 9.2} 
     & 86.8 {\scriptsize $\pm$ 0.7} \\
       & \textsc{w+s}
     & 83.0 {\scriptsize $\pm$ 8.7} 
     & 88.9 {\scriptsize $\pm$ 0.9}
     & 77.4 {\scriptsize $\pm$ 13.5} 
     & 97.3 {\scriptsize $\pm$ 0.1} 
     & 90.8 {\scriptsize $\pm$ 5.9} 
     & 81.8 {\scriptsize $\pm$ 1.6} 
     & 80.6 {\scriptsize $\pm$ 11.6} 
     & 88.2 {\scriptsize $\pm$ 0.8} \\
     \midrule
     
     \multirow{4}{*}{Llama3.3} & \textsc{0s}
     & 79.4 {\scriptsize $\pm$ 7.3} 
     & 84.0 {\scriptsize $\pm$ 0.4}
     & 68.1 {\scriptsize $\pm$ 9.0} 
     & 77.8 {\scriptsize $\pm$ 1.8} 
     & 95.9 {\scriptsize $\pm$ 6.1} 
     & 91.4 {\scriptsize $\pm$ 1.8} 
     & 74.8 {\scriptsize $\pm$ 9.4} 
     & 79.9 {\scriptsize $\pm$ 0.9} \\
       & \textsc{fs}
     & 78.9 {\scriptsize $\pm$ 7.5} 
     & 84.4 {\scriptsize $\pm$ 0.4}
     & 67.2 {\scriptsize $\pm$ 10.5} 
     & 77.8 {\scriptsize $\pm$ 0.7} 
     & \textbf{96.8} {\scriptsize $\pm$ 4.6} 
     & \textbf{92.2} {\scriptsize $\pm$ 0.6} 
     & 73.5 {\scriptsize $\pm$ 10.4} 
     & 80.3 {\scriptsize $\pm$ 0.6} \\
       & \textbf{\textsc{w}}
     & \textbf{81.0} {\scriptsize $\pm$ 7.5} 
     & \textbf{87.2} {\scriptsize $\pm$ 0.2} 
     & \textbf{71.5} {\scriptsize $\pm$ 11.1}
     & \textbf{87.9} {\scriptsize $\pm$ 0.5} 
     & 94.2 {\scriptsize $\pm$ 4.6} 
     & 86.5 {\scriptsize $\pm$ 0.3}
     & \textbf{77.3} {\scriptsize $\pm$ 9.5} 
     & \textbf{85.3} {\scriptsize $\pm$ 0.3} \\
       & \textsc{w+s}
     & 78.7 {\scriptsize $\pm$ 9.1} 
     & 86.4 {\scriptsize $\pm$ 0.3}
     & 68.4 {\scriptsize $\pm$ 12.6} 
     & 84.2 {\scriptsize $\pm$ 0.9} 
     & 94.0 {\scriptsize $\pm$ 7.4} 
     & 88.7 {\scriptsize $\pm$ 0.3} 
     & 74.0 {\scriptsize $\pm$ 11.5} 
     & 83.9 {\scriptsize $\pm$ 0.5} \\
     \midrule
     
     \multirow{4}{*}{Mistral3} 
        & \textbf{\textsc{0s}}
     & \textbf{77.4} {\scriptsize $\pm$ 8.0} 
     & \textbf{80.8} {\scriptsize $\pm$ 0.2} 
     & \textbf{72.1} {\scriptsize $\pm$ 13.2}
     & \textbf{86.9} {\scriptsize $\pm$ 0.1} 
     & 87.5 {\scriptsize $\pm$ 14.9} 
     & 75.6 {\scriptsize $\pm$ 0.4} 
     & \textbf{74.2} {\scriptsize $\pm$ 9.8} 
     & \textbf{79.3} {\scriptsize $\pm$ 0.2} \\
       & \textsc{fs}
     & 66.3 {\scriptsize $\pm$ 2.2} 
     & 74.6 {\scriptsize $\pm$ 1.9}
     & 53.7 {\scriptsize $\pm$ 2.1} 
     & 68.8 {\scriptsize $\pm$ 1.9} 
     & 87.2 {\scriptsize $\pm$ 5.1} 
     & 81.4 {\scriptsize $\pm$ 2.3} 
     & 55.8 {\scriptsize $\pm$ 3.1} 
     & 68.0 {\scriptsize $\pm$ 2.4} \\
       & \textsc{w}
     & 68.2 {\scriptsize $\pm$ 3.5} 
     & 78.4 {\scriptsize $\pm$ 0.5}
     & 54.6 {\scriptsize $\pm$ 3.8} 
     & 74.0 {\scriptsize $\pm$ 0.6} 
     & 91.4 {\scriptsize $\pm$ 7.3} 
     & 83.4 {\scriptsize $\pm$ 0.9} 
     & 57.3 {\scriptsize $\pm$ 5.6} 
     & 73.6 {\scriptsize $\pm$ 0.6} \\
       & \textsc{w+s}
     & 68.7 {\scriptsize $\pm$ 1.6} 
     & 74.9 {\scriptsize $\pm$ 0.9}
     & 53.1 {\scriptsize $\pm$ 1.7} 
     & 62.8 {\scriptsize $\pm$ 1.3} 
     & \textbf{97.2} {\scriptsize $\pm$ 2.9} 
     & \textbf{92.8} {\scriptsize $\pm$ 1.0} 
     & 55.6 {\scriptsize $\pm$ 3.0} 
     & 64.2 {\scriptsize $\pm$ 1.7} \\
     \midrule
     
     \multirow{4}{*}{Qwen2.5} & \textsc{0s}
     & 74.0 {\scriptsize $\pm$ 8.3} 
     & 83.6 {\scriptsize $\pm$ 0.1}
     & 61.1 {\scriptsize $\pm$ 10.9} 
     & 75.9 {\scriptsize $\pm$ 0.0} 
     & \textbf{95.6} {\scriptsize $\pm$ 9.0} 
     & \textbf{93.1} {\scriptsize $\pm$ 0.3} 
     & 65.5 {\scriptsize $\pm$ 12.4} 
     & 79.0 {\scriptsize $\pm$ 0.1} \\
       & \textsc{fs}
     & 77.9 {\scriptsize $\pm$ 7.1} 
     & 87.2 {\scriptsize $\pm$ 0.2}
     & 68.4 {\scriptsize $\pm$ 6.7} 
     & 90.7 {\scriptsize $\pm$ 0.3} 
     & 91.4 {\scriptsize $\pm$ 11.1} 
     & 83.9 {\scriptsize $\pm$ 0.2} 
     & 74.2 {\scriptsize $\pm$ 7.6} 
     & 85.8 {\scriptsize $\pm$ 0.2} \\
       & \textbf{\textsc{w}}
     & \textbf{79.7} {\scriptsize $\pm$ 9.6} 
     & 86.8 {\scriptsize $\pm$ 0.4} 
     & \textbf{74.1} {\scriptsize $\pm$ 7.2}
     & \textbf{93.4} {\scriptsize $\pm$ 0.2} 
     & 87.5 {\scriptsize $\pm$ 15.1} 
     & 81.0 {\scriptsize $\pm$ 0.5}
     & \textbf{78.2} {\scriptsize $\pm$ 8.5} 
     & 85.8 {\scriptsize $\pm$ 0.3} \\
       & \textsc{w+s}
     & 79.1 {\scriptsize $\pm$ 9.2} 
     & \textbf{87.3} {\scriptsize $\pm$ 0.1}
     & 72.6 {\scriptsize $\pm$ 7.6} 
     & 92.1 {\scriptsize $\pm$ 0.6} 
     & 88.3 {\scriptsize $\pm$ 14.8} 
     & 82.9 {\scriptsize $\pm$ 0.3} 
     & 77.1 {\scriptsize $\pm$ 8.5} 
     & \textbf{86.0} {\scriptsize $\pm$ 0.2} \\
     \midrule
     
     \multirow{4}{*}{R1-D} & \textsc{0s}
     & 75.4 {\scriptsize $\pm$ 4.1} 
     & 86.7 {\scriptsize $\pm$ 0.3}
     & 61.2 {\scriptsize $\pm$ 5.2} 
     & 78.5 {\scriptsize $\pm$ 0.4} 
     & \textbf{98.5} {\scriptsize $\pm$ 2.0} 
     & \textbf{96.7} {\scriptsize $\pm$ 0.1} 
     & 67.5 {\scriptsize $\pm$ 6.5} 
     & 82.9 {\scriptsize $\pm$ 0.4} \\
       & \textsc{fs}
     & 75.1 {\scriptsize $\pm$ 4.4} 
     & 87.0 {\scriptsize $\pm$ 0.3}
     & 61.3 {\scriptsize $\pm$ 5.6} 
     & 80.0 {\scriptsize $\pm$ 0.5} 
     & 97.4 {\scriptsize $\pm$ 3.1} 
     & 95.3 {\scriptsize $\pm$ 0.6} 
     & 67.4 {\scriptsize $\pm$ 7.0} 
     & 83.5 {\scriptsize $\pm$ 0.4} \\
       & \textsc{w}
     & 77.2 {\scriptsize $\pm$ 4.6} 
     & 86.8 {\scriptsize $\pm$ 0.4}
     & 65.6 {\scriptsize $\pm$ 6.2} 
     & 82.2 {\scriptsize $\pm$ 0.7} 
     & 94.1 {\scriptsize $\pm$ 3.1} 
     & 91.9 {\scriptsize $\pm$ 1.0} 
     & 71.9 {\scriptsize $\pm$ 6.9} 
     & 83.8 {\scriptsize $\pm$ 0.5} \\
       & \textbf{\textsc{w+s}}
     & \textbf{78.3} {\scriptsize $\pm$ 4.9} 
     & \textbf{88.3} {\scriptsize $\pm$ 0.6} 
     & \textbf{66.9} {\scriptsize $\pm$ 6.7} 
     & \textbf{85.6} {\scriptsize $\pm$ 0.4} 
     & 94.8 {\scriptsize $\pm$ 4.5} 
     & 91.1 {\scriptsize $\pm$ 1.0}
     & \textbf{73.4} {\scriptsize $\pm$ 6.9} 
     & \textbf{86.1} {\scriptsize $\pm$ 0.7} \\
     \midrule
     
     \multirow{4}{*}{R1} & \textsc{0s}
     & 76.4 {\scriptsize $\pm$ 6.6} 
     & 88.5 {\scriptsize $\pm$ 0.2}
     & 62.5 {\scriptsize $\pm$ 9.4} 
     & 79.8 {\scriptsize $\pm$ 0.3} 
     & \textbf{99.3} {\scriptsize $\pm$ 1.2} 
     & \textbf{99.2} {\scriptsize $\pm$ 0.1}
     & 68.5 {\scriptsize $\pm$ 10.2} 
     & 85.1 {\scriptsize $\pm$ 0.3} \\
       & \textsc{fs}
     & 80.9 {\scriptsize $\pm$ 6.5} 
     & 90.8 {\scriptsize $\pm$ 0.2}
     & 69.1 {\scriptsize $\pm$ 10.3} 
     & 84.3 {\scriptsize $\pm$ 0.4} 
     & 98.6 {\scriptsize $\pm$ 1.1} 
     & 98.5 {\scriptsize $\pm$ 0.2} 
     & 76.1 {\scriptsize $\pm$ 8.9} 
     & 88.5 {\scriptsize $\pm$ 0.3} \\
       & \textsc{w}
     & 79.9 {\scriptsize $\pm$ 5.8} 
     & 90.9 {\scriptsize $\pm$ 0.1}
     & 67.3 {\scriptsize $\pm$ 8.8} 
     & 84.2 {\scriptsize $\pm$ 0.2} 
     & 99.0 {\scriptsize $\pm$ 0.9} 
     & 98.7 {\scriptsize $\pm$ 0.1} 
     & 74.6 {\scriptsize $\pm$ 8.3} 
     & 88.6 {\scriptsize $\pm$ 0.1} \\
       & \textbf{\textsc{w+s}}
     & \textbf{81.1} {\scriptsize $\pm$ 6.0} 
     & \textbf{91.3} {\scriptsize $\pm$ 0.4} 
     & \textbf{69.5} {\scriptsize $\pm$ 9.4}
     & \textbf{85.4} {\scriptsize $\pm$ 0.7} 
     & 98.3 {\scriptsize $\pm$ 1.0} 
     & 98.0 {\scriptsize $\pm$ 0.4}
     & \textbf{76.6} {\scriptsize $\pm$ 8.2} 
     & \textbf{89.2} {\scriptsize $\pm$ 0.6} \\
    \bottomrule
  \end{tabularx}
  \caption{Percentage F1-score, precision and recall ($\pm$ standard deviation) on \UnbiasedData\ (PP) and PunEval (PE) datasets. Best results and best overall prompt per model are in bold.}
  \label{tab:rq2.1-results-complete}
\end{table*}

\subsection{RQ2: Pun Alterations}\label{supp:rq2.2}

\begin{table*}
  \small
  \centering
  \begin{tabular}{llccccc|cc}
    \toprule
    \multirow{2}{*}{\textbf{Model}} & \multirow{2}{*}{\textbf{Prompt}} & \multicolumn{5}{c}{\textbf{Substitution Category}} & \multicolumn{2}{c}{\textbf{Control groups}}\\
    \cmidrule{3-9}
     & 
     & \textbf{\texttt{Homophone}} 
     & \textbf{\texttt{Random}} 
     & \textbf{\texttt{Pun syn}} 
     & \textbf{\texttt{Alt syn}}
     & \textbf{{Pun}} 
     & \textbf{{Rand sent}} 
     & \textbf{JOKER} \\
    \midrule
    
    \multirow{4}{*}{Gemini2.0} 
    & \textsc{0s}
    & \textbf{23.7} {\scriptsize $\pm$ 5.5} 
    & \textbf{37.0} {\scriptsize $\pm$ 1.5} 
    & 11.5 {\scriptsize $\pm$ 0.5} 
    & 18.0 {\scriptsize $\pm$ 5.5} 
    & \textbf{99.0} {\scriptsize $\pm$ 1.1} 
    & 70.3 {\scriptsize $\pm$ 0.5} 
    & 26.3 {\scriptsize $\pm$ 0.1} \\
    
    & \textsc{fs}
    & 9.3 {\scriptsize $\pm$ 0.8} 
    & 16.2 {\scriptsize $\pm$ 1.0} 
    & 14.7 {\scriptsize $\pm$ 0.5} 
    & 20.3 {\scriptsize $\pm$ 5.9} 
    & 98.0 {\scriptsize $\pm$ 1.1} 
    & 85.7 {\scriptsize $\pm$ 0.5} 
    & 41.3 {\scriptsize $\pm$ 0.2} \\ 
    
    & \textsc{w}
    & 10.0 {\scriptsize $\pm$ 1.1} 
    & 14.5 {\scriptsize $\pm$ 4.9} 
    & 13.2 {\scriptsize $\pm$ 1.0} 
    & 20.3{\scriptsize $\pm$ 7.7} 
    & 97.5 {\scriptsize $\pm$ 1.4} 
    & 87.0 {\scriptsize $\pm$ 0.0} 
    & 43.8 {\scriptsize $\pm$ 0.3} \\ 
    
    & \textbf{\textsc{w+s}}
    & 11.8 {\scriptsize $\pm$ 1.0} 
    & 21.5 {\scriptsize $\pm$ 0.5} 
    & \textbf{18.2} {\scriptsize $\pm$ 2.0} 
    & \textbf{25.3}{\scriptsize $\pm$ 7.3} 
    & 97.0 {\scriptsize $\pm$ 1.1} 
    & \textbf{89.3} {\scriptsize $\pm$ 0.5} 
    & \textbf{47.9} {\scriptsize $\pm$ 0.1} \\ 
    \midrule
    
    \multirow{4}{*}{GPT-4o}
    & \textsc{0s} 
    & 9.2 {\scriptsize $\pm$ 1.3} 
    & 23.8 {\scriptsize $\pm$ 1.6} 
    & 10.7 {\scriptsize $\pm$ 0.5}
    & 23.2 {\scriptsize $\pm$ 4.9} 
    & \textbf{99.2} {\scriptsize $\pm$ 0.4} 
    & 87.7 {\scriptsize $\pm$ 1.4} 
    & 47.4 {\scriptsize $\pm$ 0.1} \\  
    
    & \textsc{fs}
    & 12.5 {\scriptsize $\pm$ 3.1} 
    & 26.2 {\scriptsize $\pm$ 3.5} 
    & 20.5 {\scriptsize $\pm$ 0.5} 
    & 32.5 {\scriptsize $\pm$ 9.1} 
    & 97.2 {\scriptsize $\pm$ 1.2} 
    & 91.3 {\scriptsize $\pm$ 0.5} 
    & 57.8 {\scriptsize $\pm$ 0.6} \\  
    
    & \textbf{\textsc{w}}
    &\textbf{32.5} {\scriptsize $\pm$ 2.3} 
    & \textbf{57.5} {\scriptsize $\pm$ 1.4}
    & \textbf{37.3} {\scriptsize $\pm$ 4.1} 
    & \textbf{58.7}{\scriptsize $\pm$ 4.8} 
    & 92.8 {\scriptsize $\pm$ 3.2} 
    & \textbf{99.0} {\scriptsize $\pm$ 0.0} 
    & \textbf{77.4} {\scriptsize $\pm$ 0.9} \\ 
    
    & \textsc{w+s}
    & 24.0 {\scriptsize $\pm$ 0.9} 
    & 39.5 {\scriptsize $\pm$ 5.5} 
    & 30.0 {\scriptsize $\pm$ 1.5} 
    & 49.0 {\scriptsize $\pm$ 5.9} 
    & 93.5 {\scriptsize $\pm$ 3.6} 
    & 97.7 {\scriptsize $\pm$ 0.5} 
    & 71.8 {\scriptsize $\pm$ 0.6} \\  
    \midrule
    
    \multirow{4}{*}{Llama3.3}
    & \textsc{0s} 
    & 22.0 {\scriptsize $\pm$ 1.4} 
    & 45.5 {\scriptsize $\pm$ 3.4} 
    & 16.5 {\scriptsize $\pm$ 2.7} 
    & 35.2 {\scriptsize $\pm$ 11.1} 
    & \textbf{96.8} {\scriptsize $\pm$ 1.5}
    & 79.7 {\scriptsize $\pm$ 2.1} 
    & 50.6 {\scriptsize $\pm$ 3.0} \\ 
    
    & \textsc{fs}
    & 16.3 {\scriptsize $\pm$ 6.0} 
    & 37.0 {\scriptsize $\pm$ 4.3} 
    & 21.8 {\scriptsize $\pm$ 1.5} 
    & 40.3 {\scriptsize $\pm$ 13.0} 
    & 96.7 {\scriptsize $\pm$ 1.6} 
    & 88.7 {\scriptsize $\pm$ 1.0} 
    & 53.2 {\scriptsize $\pm$ 0.1} \\  
    
    & \textbf{\textsc{w}}
    & \textbf{23.3} {\scriptsize $\pm$ 3.8} 
    & \textbf{49.2} {\scriptsize $\pm$ 1.2} 
    & \textbf{35.2} {\scriptsize $\pm$ 5.2}
    & \textbf{51.5}{\scriptsize $\pm$ 7.2} 
    & 90.5 {\scriptsize $\pm$ 2.1} 
    & \textbf{94.0} {\scriptsize $\pm$ 0.9} 
    & \textbf{63.9} {\scriptsize $\pm$ 0.7} \\  
    
    & \textsc{w+s}
    & 13.2 {\scriptsize $\pm$ 3.8} 
    & 32.8 {\scriptsize $\pm$ 5.9}
    & 26.5 {\scriptsize $\pm$ 2.7} 
    & 42.5 {\scriptsize $\pm$ 7.6} 
    & 92.3 {\scriptsize $\pm$ 1.2} 
    & 93.0 {\scriptsize $\pm$ 0.0} 
    & 56.0 {\scriptsize $\pm$ 1.6} \\ 
    \midrule
    
    \multirow{4}{*}{Mistral3}
    & \textbf{\textsc{0s}} 
    & \textbf{37.3} {\scriptsize $\pm$ 6.7} 
    & \textbf{53.2} {\scriptsize $\pm$ 3.3}
    & \textbf{29.2} {\scriptsize $\pm$ 1.2} 
    & \textbf{44.3}{\scriptsize $\pm$ 19.0}
    & 88.8 {\scriptsize $\pm$ 2.0} 
    & \textbf{90.0} {\scriptsize $\pm$ 0.0} 
    & \textbf{61.9} {\scriptsize $\pm$ 0.2} \\
    
    & \textsc{fs}
    & 14.5  {\scriptsize $\pm$ 4.6} 
    & 17.0 {\scriptsize $\pm$ 5.8} 
    & 15.8 {\scriptsize $\pm$ 2.0} 
    & 21.7 {\scriptsize $\pm$ 5.0} 
    & 86.7 {\scriptsize $\pm$ 2.4}
    & 37.7 {\scriptsize $\pm$ 1.4}
    & 26.4 {\scriptsize $\pm$ 0.2} \\    
    
    & \textsc{w}
    & 5.2 {\scriptsize $\pm$ 1.9} 
    & 9.2 {\scriptsize $\pm$ 1.5} 
    & 12.3 {\scriptsize $\pm$ 2.0} 
    & 19.2 {\scriptsize $\pm$ 5.1} 
    & 93.3 {\scriptsize $\pm$ 3.4} 
    & 48.7 {\scriptsize $\pm$ 3.1} 
    & 24.1 {\scriptsize $\pm$ 1.6} \\ 
    
    & \textsc{w+s}
    & 2.7 {\scriptsize $\pm$ 1.4}
    & 3.5 {\scriptsize $\pm$ 2.5} 
    & 6.5 {\scriptsize $\pm$ 1.8} 
    & 9.5 {\scriptsize $\pm$ 3.4} 
    & \textbf{96.7} {\scriptsize $\pm$ 1.5} 
    & 25.7 {\scriptsize $\pm$ 4.2} 
    & 14.0 {\scriptsize $\pm$ 1.7} \\   
    \midrule
    
    \multirow{4}{*}{Qwen2.5}
    & \textsc{0s} 
    & 5.0 {\scriptsize $\pm$ 0.6} 
    & 11.2 {\scriptsize $\pm$ 3.7} 
    & 5.7 {\scriptsize $\pm$ 0.8} 
    & 11.3 {\scriptsize $\pm$ 3.8} 
    & \textbf{98.2} {\scriptsize $\pm$ 0.4} 
    & 59.3 {\scriptsize $\pm$ 2.1} 
    & 36.6 {\scriptsize $\pm$ 0.3} \\ 
    
    & \textsc{fs}
    & 11.3 {\scriptsize $\pm$ 3.4} 
    & 14.7 {\scriptsize $\pm$ 2.8} 
    & 19.0 {\scriptsize $\pm$ 4.6} 
    & 33.7 {\scriptsize $\pm$ 11.7} 
    & 94.7 {\scriptsize $\pm$ 1.0} 
    & 90.0 {\scriptsize $\pm$ 0.9} 
    & 56.5 {\scriptsize $\pm$ 0.9} \\ 
    
    & \textbf{\textsc{w}}
    & \textbf{19.2} {\scriptsize $\pm$ 5.2} 
    & \textbf{32.7} {\scriptsize $\pm$ 3.4} 
    & 26.3 {\scriptsize $\pm$ 5.6} 
    & \textbf{40.5}{\scriptsize $\pm$ 12.2}
    & 92.8 {\scriptsize $\pm$ 1.6} 
    & \textbf{94.3} {\scriptsize $\pm$ 1.4} 
    & \textbf{65.7} {\scriptsize $\pm$ 0.5} \\
    
    & \textsc{w+s} 
    & 15.5  {\scriptsize $\pm$ 4.4} 
    & 29.8 {\scriptsize $\pm$ 5.8} 
    & \textbf{27.0} {\scriptsize $\pm$ 3.6} 
    & 37.2 {\scriptsize $\pm$ 11.9} 
    & 95.0 {\scriptsize $\pm$ 1.4} 
    & 92.7 {\scriptsize $\pm$ 1.4} 
    & 62.7 {\scriptsize $\pm$ 1.1} \\   
    \midrule
    
    \multirow{4}{*}{R1-D} 
    & \textsc{0s} 
    & 6.2 {\scriptsize $\pm$ 1.2} 
    & 24.9 {\scriptsize $\pm$ 6.6}
    & 14.2 {\scriptsize $\pm$ 1.2} 
    & 21.2{\scriptsize $\pm$ 8.1} 
    & \textbf{99.3} {\scriptsize $\pm$ 0.5} 
    & 59.0 {\scriptsize $\pm$ 2.4} 
    & 37.5 {\scriptsize $\pm$ 2.2} \\   
    
    & \textsc{fs}
    & 7.1 {\scriptsize $\pm$ 2.3} 
    & 27.3 {\scriptsize $\pm$ 3.8} 
    & 18.8 {\scriptsize $\pm$ 4.5} 
    & 26.9 {\scriptsize $\pm$ 9.8} 
    & 98.5 {\scriptsize $\pm$ 0.5} 
    & 69.7 {\scriptsize $\pm$ 5.5} 
    & 42.9 {\scriptsize $\pm$ 1.3} \\    
    
    & \textsc{w}
    & 8.5 {\scriptsize $\pm$ 3.7}
    & \textbf{36.9} {\scriptsize $\pm$ 3.1} 
    & 26.7 {\scriptsize $\pm$ 2.7} 
    & 34.6{\scriptsize $\pm$ 8.3} 
    & 96.5 {\scriptsize $\pm$ 2.0} 
    & 79.3 {\scriptsize $\pm$ 5.2} 
    & 52.1 {\scriptsize $\pm$ 1.4} \\  
    
    & \textbf{\textsc{w+s}}
    & \textbf{9.7} {\scriptsize $\pm$ 2.2} 
    & 34.3 {\scriptsize $\pm$ 1.4} 
    & \textbf{28.8} {\scriptsize $\pm$ 5.0} 
    &  \textbf{38.8}{\scriptsize $\pm$ 11.6} 
    & 96.3 {\scriptsize $\pm$ 1.2} 
    & \textbf{82.7} {\scriptsize $\pm$ 2.3} 
    & \textbf{55.5} {\scriptsize $\pm$ 3.3} \\   
    \midrule
    
    \multirow{4}{*}{R1} 
    & \textsc{0s}
    & 6.5 {\scriptsize $\pm$ 2.7} 
    & 27.3 {\scriptsize $\pm$ 4.1} 
    & 12.2 {\scriptsize $\pm$ 2.0} 
    & 18.8 {\scriptsize $\pm$ 7.3} 
    & 99.8 {\scriptsize $\pm$ 0.4} 
    & 71.3 {\scriptsize $\pm$ 1.0} 
    & 37.0 {\scriptsize $\pm$ 1.7} \\ 
    
    & \textsc{fs}
    & 8.5 {\scriptsize $\pm$ 2.3} 
    & 38.8 {\scriptsize $\pm$ 3.7} 
    & 22.3 {\scriptsize $\pm$ 2.9} 
    & 31.8 {\scriptsize $\pm$ 4.1} 
    & 99.3 {\scriptsize $\pm$ 0.8} 
    & 83.0 {\scriptsize $\pm$ 0.9}
    & 48.2 {\scriptsize $\pm$ 1.5} \\
    
    & \textsc{w}
    & 9.5 {\scriptsize $\pm$ 1.8} 
    & 40.3 {\scriptsize $\pm$ 2.7} 
    & 22.3 {\scriptsize $\pm$ 3.1} 
    & 34.0 {\scriptsize $\pm$ 6.0} 
    & 99.8 {\scriptsize $\pm$ 0.4} 
    & 84.7 {\scriptsize $\pm$ 2.3} 
    & 44.9 {\scriptsize $\pm$ 0.5} \\
    
    & \textbf{\textsc{w+s}}
    & \textbf{10.5} {\scriptsize $\pm$ 2.6} 
    & \textbf{41.8}{\scriptsize $\pm$ 2.0} 
    & \textbf{25.5} {\scriptsize $\pm$ 2.8} 
    & \textbf{35.3}{\scriptsize $\pm$ 8.3} 
    & \textbf{99.8} {\scriptsize $\pm$ 0.4} 
    & \textbf{84.7} {\scriptsize $\pm$ 1.9} 
    & \textbf{51.2} {\scriptsize $\pm$ 0.1} \\  
    \bottomrule
  \end{tabular}
  \caption{Percentage recall (on the true class) $\pm$ standard deviation for each substitution category on the \SubstitutionData\ dataset. Note that for all columns except JOKER, \textit{recall} equals \textit{accuracy} because each evaluation set contains only puns or only non-puns. JOKER reports recall for the negative class on the full dataset. The best overall prompt per model is marked in bold.}
  \label{tab:rq2.2-results-recall}
\end{table*}

Table \ref{tab:rq2.2-results-recall} contains the performance (recall) on the substitution categories for each prompt. For the ``Pun'' column, the recall is computed for class 1 (pun), while for the other columns, recall is relative to class 0 (non-pun). The ``Rand sent'' column represents the set of 100 generated non-puns used to control for unseen bias toward class 1 (pun). The highest values for each model are highlighted in bold. 

\paragraph{Model confidence.}\label{sec:model-confidence-sub} 
We observe that, on most substitution categories in the \SubstitutionData\ dataset, the standard deviation between measurements is significantly higher compared to the examples of positive (i.e., \texttt{pun}) and random sentences (\texttt{rand sent}). This indicates that models are less confident when classifying altered puns. To further investigate this, we retrieve the confidence values for the responses across the entire dataset. Specifically, we collect the log probabilities for the ``yes'' or ``no'' tokens. 

Since not all LLMs provide the option to retrieve confidence scores at the time of writing (e.g., DeepSeek models and Gemini), we limit this step to GPT-4o, Qwen 2.5, and Llama3.3, which represent three top-performing models in the previous tasks.

\begin{figure}[ht!]
  \includegraphics[width=\columnwidth]{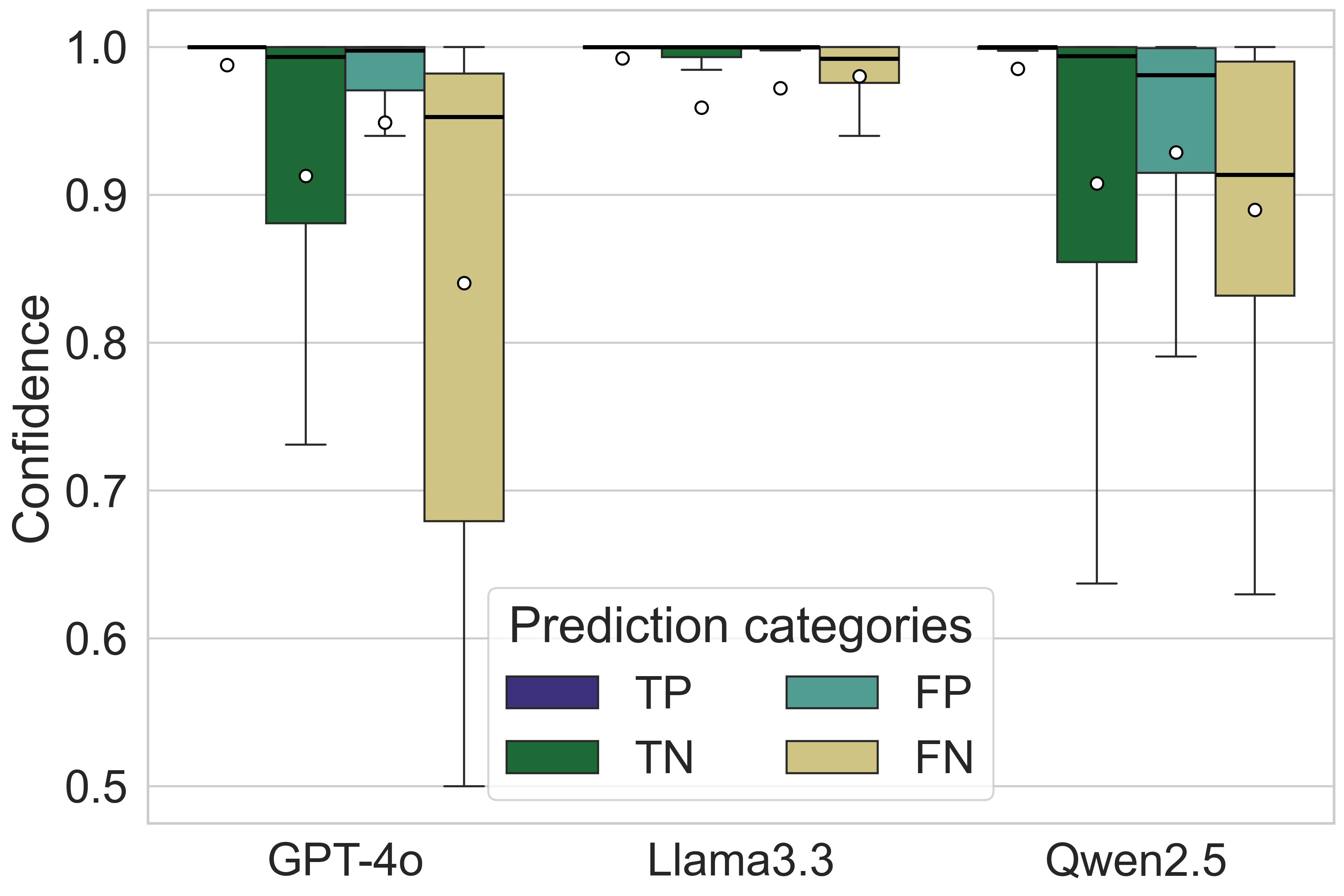}
  \caption {Confidence on each prediction class (\textsc{w+s} prompt).}
  \label{fig:confidence_per_class}
\end{figure}

Our analysis confirms that, on the \SubstitutionData\ dataset, these models exhibit very high confidence when predicting puns but are much less confident when predicting the non-pun class. Fig. \ref{fig:confidence_per_class} clearly illustrates this trend, with the True Negative (TN) and False Negative (FN) classes showing significantly more variance than the other classes. It is important to note that the number of FNs is quite low, with only 38 instances across the three models.
The lower confidence levels indicate that classifying altered examples as non-puns is significantly more challenging for the models compared to correctly identifying puns from the PunEval dataset and the negative sentences we generated as a control set. 


\subsection{RQ3}

To isolate the quality of the rationales from raw detection performance, we introduce an alternative evaluation criterion. For this analysis, we calculate the PPA score exclusively on true positives (i.e., puns that the LLMs correctly identified). The results for this specific subset are presented in Table \ref{tab:rq3-agreement-tp}.

This targeted analysis reveals that most models can generate reasonably accurate rationales for the puns they successfully identify. Mistral and Qwen achieve lower scores in this setting, producing up to 40\% incorrect rationales, while GPT‑4o, Gemini, and R1 produce correct predictions for the wrong reasons at least 20\% of the time.

However, this metric does not account for a model's overall detection rate; a model might excel at explaining the few puns it identifies, yet fail to detect many others. We therefore argue that the comprehensive PPA score in Table \ref{tab:rq3-agreement}, which considers all predictions, provides a more robust and useful comparison. Consequently, we base our selection of models for manual analysis on that initial metric.

\begin{table}[tbp]
  \small
  \centering
  \begin{tabular}{lcccccc}
    \toprule
    \multirow{2}{*}{\textbf{Model}} & 
    \multicolumn{2}{c}{\textbf{\NAPData}} & 
    \multicolumn{2}{c}{\textbf{JOKER}} & 
    \multicolumn{2}{c}{\textbf{PunEval}} \\
    \cmidrule(lr){2-3} \cmidrule(lr){4-5} \cmidrule(lr){6-7} 
    & \textsc{w} & \textsc{w+s} 
    & \textsc{w} & \textsc{w+s} 
    & \textsc{w} & \textsc{w+s} \\
    \midrule
    Gemini2.0
    & \textbf{1.6} & \textbf{1.5} 
    & 1.4 & 1.4
    & 1.6 & 1.6 \\
    GPT-4o 
    & 1.5 & \textbf{1.5}
    & \textbf{1.6} & 1.5
    & \textbf{1.8} & \textbf{1.8} \\
    Llama3.3 
    & 1.4 & 1.4 
    & 1.4 & 1.4
    & 1.6 & 1.6 \\ 
    Mistral3 
    & 1.2 & 1.2
    & 1.3 & 1.2
    & 1.4 & 1.3 \\
    Qwen2.5 
    & 1.3 & 1.3
    & 1.3 & 1.4
    & 1.5 & 1.6 \\
    R1-D 
    & 1.5 & \textbf{1.5}
    & 1.5 & 1.5
    & 1.7 & 1.7 \\
    R1 
    & 1.5 & \textbf{1.5}
    & \textbf{1.6} & \textbf{1.6}
    & \textbf{1.8} & \textbf{1.8} \\
    \bottomrule
  \end{tabular}
  \caption{Average Pun Pair Agreement (PPA) measured on the true positive examples, in $[0\text{--}2]$, for each prompt setting (\textsc{w}, \textsc{w+s}) across datasets. Standard deviation is 0.0 for all measurements, and the best results are marked in bold.}
  \label{tab:rq3-agreement-tp}
\end{table}

\begin{figure}[htb]
  \centering
  \includegraphics[width=\columnwidth]{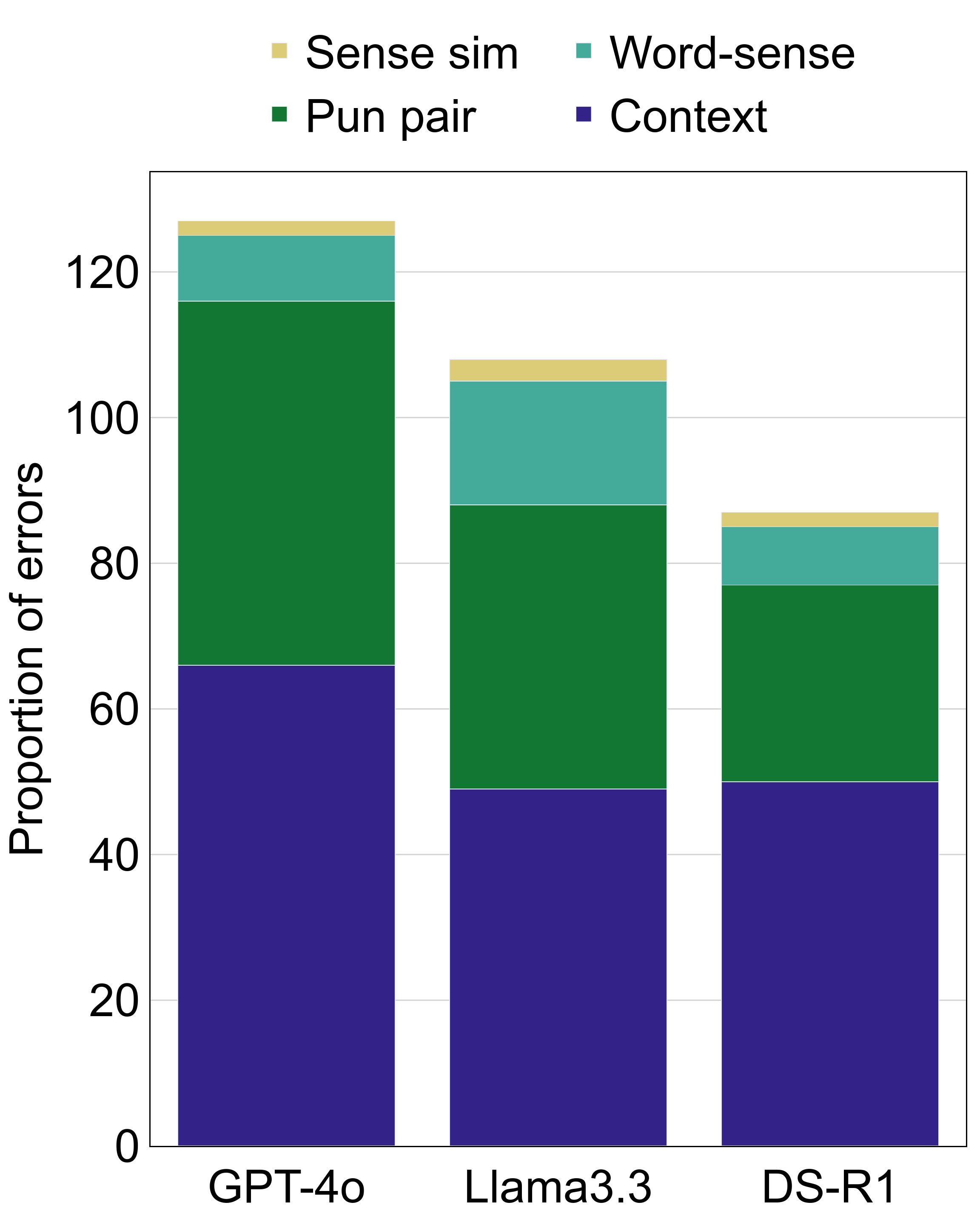}
  \caption {Frequency of LLMs' mistakes in the generated rationales.}
  \label{fig:error_analysis_categories}
\end{figure}

\begin{figure*}[p]
  \centering
  \includegraphics[width=\linewidth]{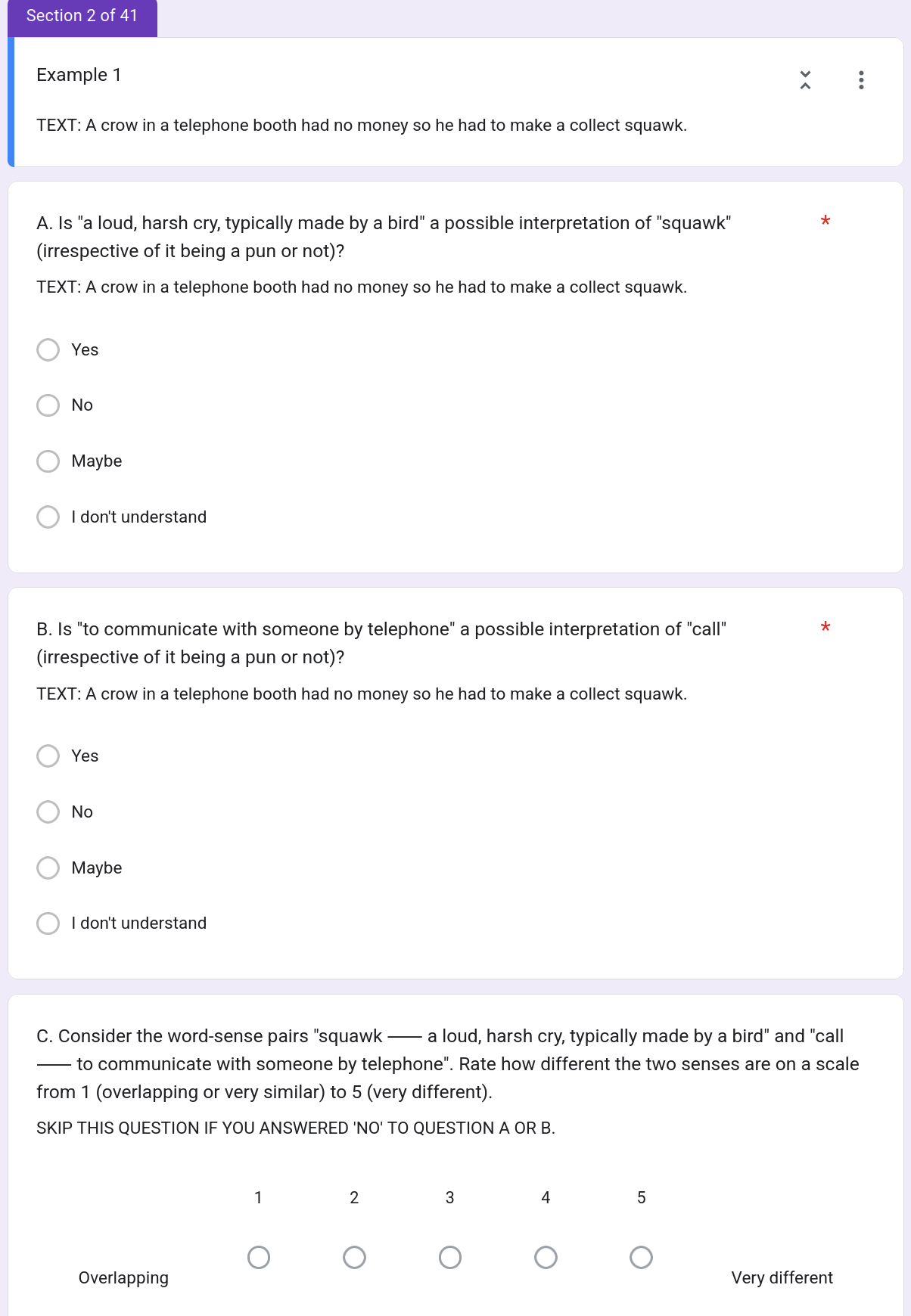}
  \caption {Questions A-B-C utilized in the Error Analysis.}
  \label{fig:ea_abc}
\end{figure*}

\begin{figure*}[p]
  \centering
  \includegraphics[width=\linewidth]{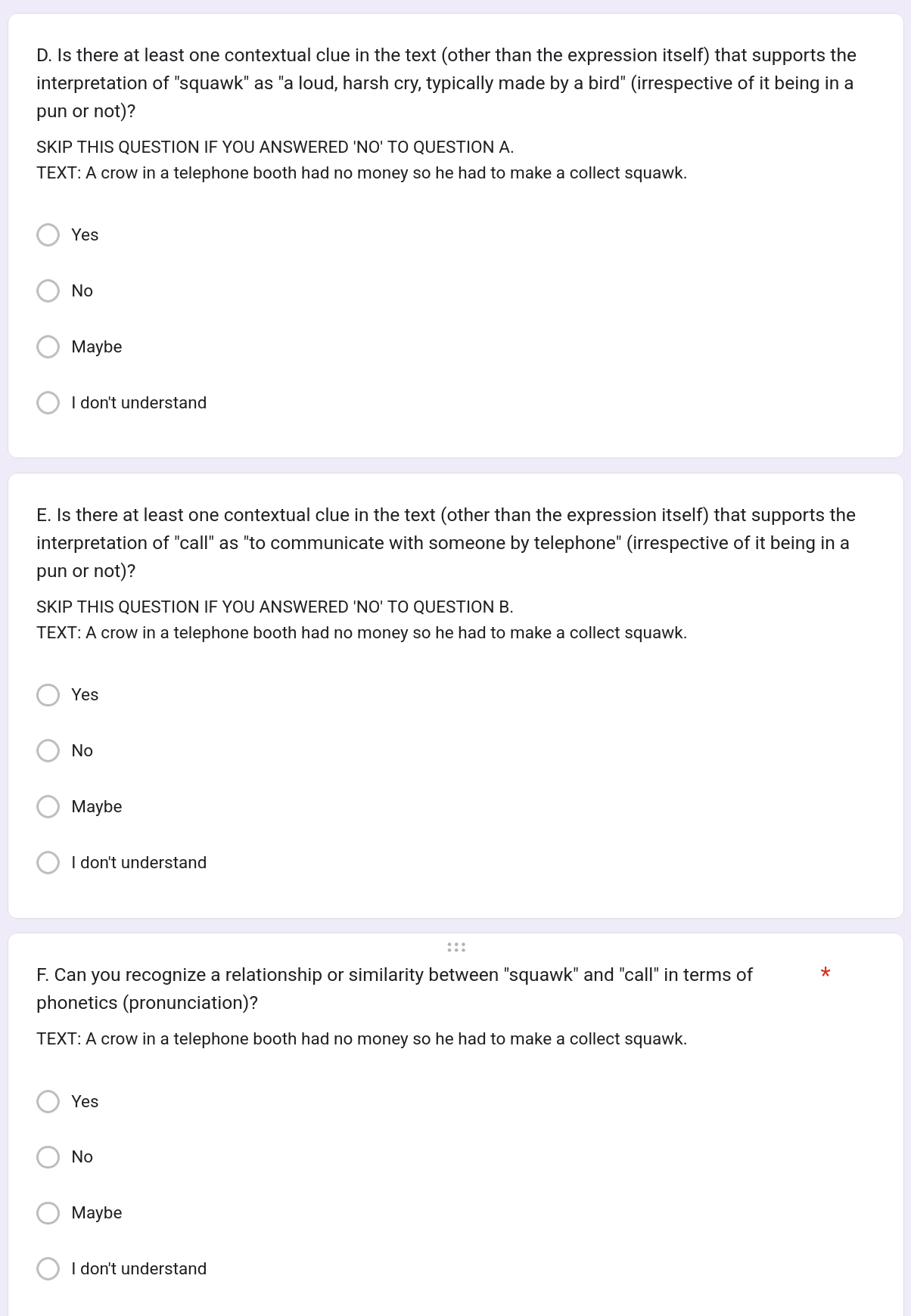}
  \caption {Questions D-E-F utilized in the Error Analysis.}
  \label{fig:ea_def}
\end{figure*}

\begin{figure*}[p]
  \centering
  \includegraphics[width=\linewidth]{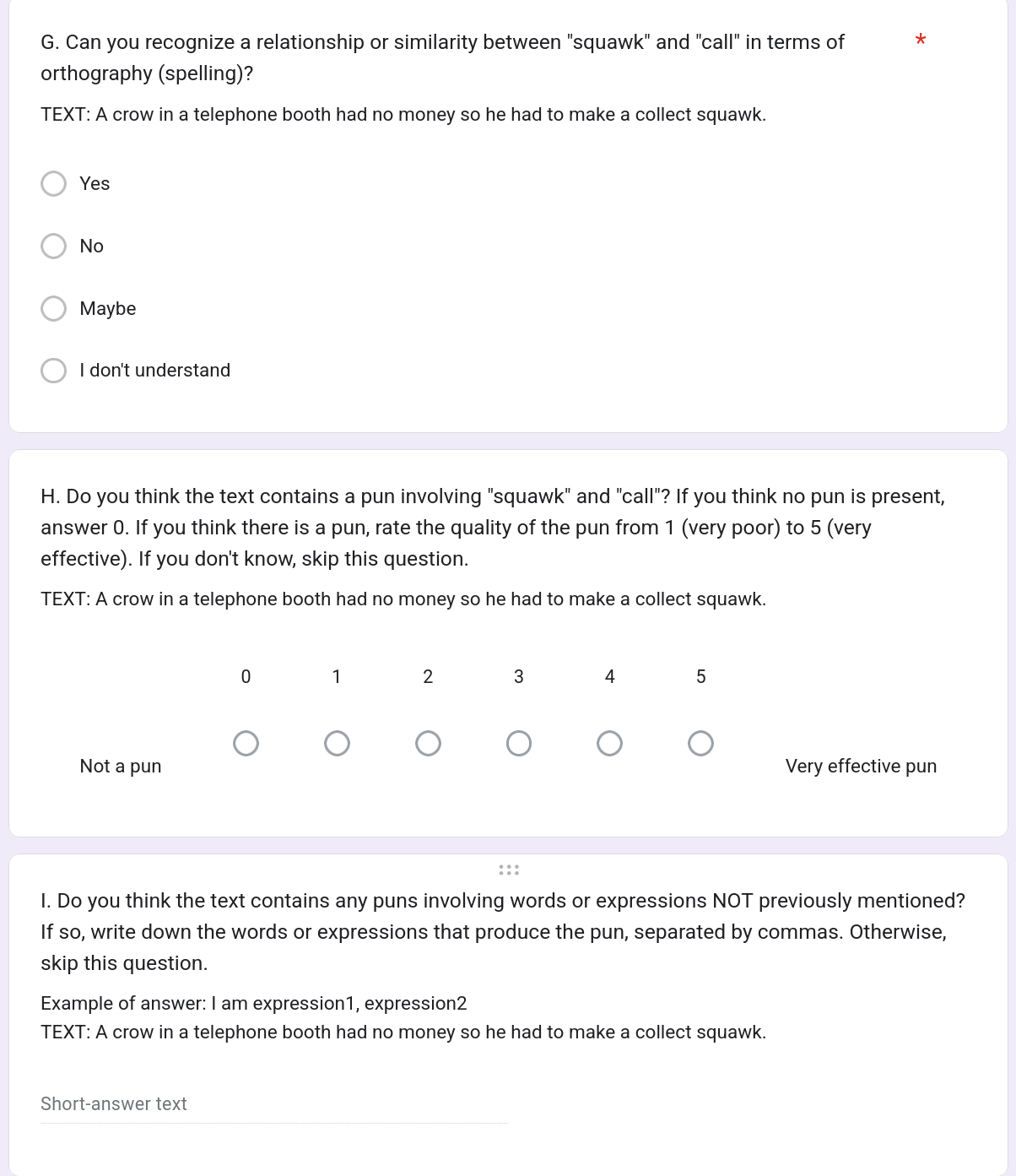}
  \caption {Questions G-H-I utilized in the Error Analysis.}
  \label{fig:ea_ghi}
\end{figure*}

The barplot in Fig. \ref{fig:error_analysis_categories} shows the distribution of errors made by the LLMs in each category.
Our analysis indicates that the most frequent error in the rationales is the identification of double meanings that lack support from the surrounding context (\texttt{Context}). For instance, the phrase ``\textit{Long fairy tales have a tendency to wyvern}'' was incorrectly interpreted as a play on words between \textit{wyvern} (a mythical dragon-like creature) and \textit{wizen} (to become dry or shriveled). Although the meanings of the words are correctly matched (so it is not a \texttt{Word-sense pair} error), there are no contextual clues to support the use of \textit{wizen}. Additionally, the annotators could not recognize any similarity in spelling or pronunciation between the two words, highlighting a mistake in the \texttt{Pun pair} category. 
A third typical mistake is the incorrect pairing of words to their meanings (\texttt{Word-sense pair}). For example, in the non-pun phrase ``\textit{Her decision to take up rock climbing was preset-pit-toes to say the least}'' Llama incorrectly interpreted the expression \textit{preset-pit-toes} as ''ready to jump or spring into action'' apparently referring back to the original pun word ``precipitous'' and ignoring the different spelling. 

Examples of all the error categories are shown in Table \ref{tab:examples-error-analysis}, along with human annotators' questionnaire responses. For brevity, the questionnaire items shown in the table have been slightly shortened; the full questionnaire, guidelines, and annotation examples are attached at the end of this document.

\begin{table*}[htbp]
    \centering
    \small
    \setlist[itemize]{
        nosep, 
        leftmargin=*, 
        label=\textbullet,
    }
    \begin{tabularx}{\textwidth}{p{4em}p{6em}Xp{4.5em}}
        \toprule
        \multicolumn{1}{c}{\textbf{Input}} & \multicolumn{1}{c}{\textbf{Output}} & \multicolumn{1}{c}{\textbf{Answers by annotators}} & \textbf{Errors} \\
        
        \midrule
        \multicolumn{4}{c}{\textbf{GPT-4o}}\\
        \midrule
        
        German sausage jokes are just the sausage
        & yes <the sausage> <the wurst> <the main part or essence of something> <a type of German sausage>
        & 
        \begin{minipage}[t]{\linewidth}
        \begin{itemize}
        \item A. Can ``the sausage'' be interpreted as ``the main part or essence of something''? \textbf{\red{No}}
        \item B. Can ``the wurst'' be interpreted as ``a type of German sausage''? Yes
        \item E. Does the context support the interpretation of ``the wurst'' as ``a type of German sausage''? Yes
        \item FG. Are ``the sausage'' and ``the wurst'' orthographically or phonetically close? \textbf{\red{No}}
        \item H. Rate pun quality in $0-5$: 1
        \end{itemize}
        \end{minipage}
        & \texttt{word-sense} ($A\&B$), \texttt{pun pair} ($F|G$) \\
        \midrule
        
        Those who make magnets can be quite charming
        & yes <charming> <charming> <attractive or pleasing> <using magic>
        & 
        \begin{minipage}[t]{\linewidth}
        \begin{itemize}
        \item A. Can ``charming'' be interpreted as ``attractive or pleasing''? Yes
        \item B. Can ``charming'' be interpreted as ``using magic''? Yes
        \item C. How diverse are the two previous senses in $0-5$? 2
        \item D. Does the context support the interpretation of ``charming'' as ``using magic''? \textbf{\red{No}}
        \item E. Does the context support the interpretation of ``charming'' as ``attractive or pleasing''? Yes
        \item H. Rate pun quality in $0-5$: 0
        \end{itemize}
        \end{minipage}
        & \texttt{context} ($D\&E$) \\
        
        \midrule
        \multicolumn{4}{c}{\textbf{Llama3.3}}\\
        \midrule
        
        The sailor called his girlfriend's belly ring an oven destroyer
        & yes <oven destroyer> <over destroyer> <a device that ruins an oven> <one who destroys from above>
        & 
        \begin{minipage}[t]{\linewidth}
        \begin{itemize}
        \item A. Can ``oven destroyer'' be interpreted as ``a device that ruins an oven''? Yes
        \item B. Can ``over destroyer'' be interpreted as ``one who destroys from above''? \textbf{\red{No}}
        \item D. Does the context support the interpretation of ``oven destroyer'' as ``a device that ruins an oven''? \textbf{\red{No}}
        \item FG. Are ``oven destroyer'' and ``over destroyer'' orthographically or phonetically close? Yes
        \item H. Rate pun quality in $0-5$: 0
        \end{itemize}
        \end{minipage}
        & \texttt{word-sense} ($A\&B$), \texttt{context} ($D\&E$) \\
        \midrule

        A circus lion won't eat clowns because they taste humorous
        & yes <humorous> <humorous> <having or showing a sense of humor> <having a funny or unusual taste>
        &
        \begin{minipage}[t]{\linewidth}
        \begin{itemize}
        \item A. Can ``humorous'' be interpreted as ``having or showing a sense of humor''? Yes
        \item B. Can ``humorous'' be interpreted as ``having a funny or unusual taste''? Maybe
        \item C. How diverse are the two previous senses in $0-5$? \textbf{\red{1}}
        \item D. Does the context support the interpretation of ``humorous'' as ``having or showing a sense of humor''? Yes
        \item E. Does the context support the interpretation of ``humorous'' as ``having a taste, especially a funny or unusual one''? Yes
        \item H. Rate pun quality in $0-5$: 1
        \end{itemize}
        \end{minipage}
        & \texttt{sense sim} (C) \\

        \midrule
        \multicolumn{4}{c}{\textbf{DeepSeek-R1}}\\
        \midrule

        If a dog was computer literate, would his bark be worse than his octet?
        & yes <octet> <bite> <a group of eight bits in computing> <the act of using teeth to cut or pierce>
        &
        \begin{minipage}[t]{\linewidth}
        \begin{itemize}
        \item A. Can ``octet'' be interpreted as ``a group of eight bits in computing''? Yes
        \item B. Can ``bite'' be interpreted as ``the act of using teeth to cut or pierce''? Yes
        \item C. How diverse are the two previous senses in $0-5$? 4
        \item D. Does the context support the interpretation of ``octet'' as ``a group of eight bits in computing''? \textbf{\red{No}}
        \item E. Does the context support the interpretation of ``bite'' as ``the act of using teeth to cut or pierce''? \textbf{\red{No}}
        \item FG. Are ``octet'' and ``bite'' orthographically or phonetically close? \textbf{\red{No}}
        \item H. Rate pun quality in $0-5$: 0
        \end{itemize}
        \end{minipage}
        & \texttt{context} ($D\&E$), \texttt{pun pair} ($F|G$) \\
        \midrule
        
        Why are frogs so happy? They eat whatever feather them
        & yes <feather> <fed them> <to cover with feathers> <to provide food for them>
        &
        \begin{minipage}[t]{\linewidth}
        \begin{itemize}
        \item A. Can ``feather'' be interpreted as ``to cover with feathers''? \textbf{\red{No}}
        \item B. Can ``fed them'' be interpreted as ``to provide food for them''? Yes
        \item E. Does the context support the interpretation of ``fed them'' as ``to provide food for them''? Yes
        \item FG. Are ``feather'' and ``fed them'' orthographically or phonetically close? Yes
        \item H. Rate pun quality in $0-5$: 2
        \end{itemize}
        \end{minipage}
        & \texttt{word-sense} ($A\&B$) \\
        
        \bottomrule
    \end{tabularx}
    \caption{Examples of error-analysis results for the three models analyzed in RQ3. The last column lists the errors derived from annotators' answers and used in the final statistics. The logical expression in brackets indicates how each error was computed; see §\ref{app:user-study-processing} for details. Questions that were not relevant to the sample or were unanswered are omitted. Questions \textit{F} and \textit{G} are combined for brevity. Answers marked in red indicate the model's errors.}
    \label{tab:examples-error-analysis}
\end{table*}

\clearpage

\clearpage


\section*{Supplementary material}

**Section 2 of 41**

**Example 1**

TEXT: A crow in a telephone booth had no money so he had to make a collect squawk.

A. Is "a loud, harsh cry, typically made by a bird" a possible interpretation of "squawk" (irrespective of it being a pun or not)? *

TEXT: A crow in a telephone booth had no money so he had to make a collect squawk.

○ Yes

○ No

○ Maybe

○ I don't understand

B. Is "to communicate with someone by telephone" a possible interpretation of "call" (irrespective of it being a pun or not)? *

TEXT: A crow in a telephone booth had no money so he had to make a collect squawk.

○ Yes

○ No

○ Maybe

○ I don't understand

C. Consider the word-sense pairs "squawk —— a loud, harsh cry, typically made by a bird" and "call —— to communicate with someone by telephone". Rate how different the two senses are on a scale from 1 (overlapping or very similar) to 5 (very different).

SKIP THIS QUESTION IF YOU ANSWERED 'NO' TO QUESTION A OR B.

| | 1 | 2 | 3 | 4 | 5 | |
|---|---|---|---|---|---|---|
| Overlapping | ○ | ○ | ○ | ○ | ○ | Very different |

Figure 16: Questions A-B-C utilized in the Error Analysis.

D. Is there at least one contextual clue in the text (other than the expression itself) that supports the interpretation of "squawk" as "a loud, harsh cry, typically made by a bird" (irrespective of it being in a pun or not)?

SKIP THIS QUESTION IF YOU ANSWERED 'NO' TO QUESTION A.
TEXT: A crow in a telephone booth had no money so he had to make a collect squawk.

○ Yes

○ No

○ Maybe

○ I don't understand

E. Is there at least one contextual clue in the text (other than the expression itself) that supports the interpretation of "call" as "to communicate with someone by telephone" (irrespective of it being in a pun or not)?

SKIP THIS QUESTION IF YOU ANSWERED 'NO' TO QUESTION B.
TEXT: A crow in a telephone booth had no money so he had to make a collect squawk.

○ Yes

○ No

○ Maybe

○ I don't understand

F. Can you recognize a relationship or similarity between "squawk" and "call" in terms of phonetics (pronunciation)? *

TEXT: A crow in a telephone booth had no money so he had to make a collect squawk.

○ Yes

○ No

○ Maybe

○ I don't understand

Figure 17: Questions D-E-F utilized in the Error Analysis.

G. Can you recognize a relationship or similarity between "squawk" and "call" in terms of orthography (spelling)? *

TEXT: A crow in a telephone booth had no money so he had to make a collect squawk.

○ Yes

○ No

○ Maybe

○ I don't understand

H. Do you think the text contains a pun involving "squawk" and "call"? If you think no pun is present, answer 0. If you think there is a pun, rate the quality of the pun from 1 (very poor) to 5 (very effective). If you don't know, skip this question.

TEXT: A crow in a telephone booth had no money so he had to make a collect squawk.

| | 0 | 1 | 2 | 3 | 4 | 5 | |
|---|---|---|---|---|---|---|---|
| Not a pun | ○ | ○ | ○ | ○ | ○ | ○ | Very effective pun |

I. Do you think the text contains any puns involving words or expressions NOT previously mentioned? If so, write down the words or expressions that produce the pun, separated by commas. Otherwise, skip this question.

Example of answer: I am expression1, expression2
TEXT: A crow in a telephone booth had no money so he had to make a collect squawk.

Short-answer text

Figure 18: Questions G-H-I utilized in the Error Analysis.

# Annotator's Guidelines

Pun Explanation - Error Analysis

# Task Introduction

*Welcome*! Thank you for participating in this study.

Your responses will contribute to an **error analysis** that examines how Artificial Intelligence (AI) explains the use of puns in short English texts.

Please **read the following sections carefully** before proceeding to the questions.

## Background

*Puns* are a type of wordplay that exploits words with multiple meanings or similar-sounding words for humorous or rhetorical effect. The *pun effect* is created by the greatly accelerated perception of two different senses of an expression.

For example, in the sentence "*You lamb! said Tom sheepishly*", there is a pun because "sheepishly" can be interpreted as "in an embarrassed manner" while also evoking "sheep" referring to the "woolly ruminant mammal".

Another example is "*The plot to his story of the pond was quite shallow*", where the word "shallow" can be interpreted in the sense of "lacking physical depth" and "lacking depth of intellect or knowledge".

The first sentence contains a **heterographic pun**, as "sheepishly" and "sheep" are different words. The second example features a **homographic pun**, where the same word "shallow" is used in two different senses.

In our study, we asked an AI agent to analyze short texts, some containing puns and some not, and to identify the words and senses that create the *pun effect*.

In the following quiz, you will see a selection of non-puns that the AI agent classified as puns. Your job is answering 9 questions to help us evaluate how wrong the AI agent's answers are.

## Definitions

- ***Expression***. An expression can be a single word, such as "car" or "beach", or it can contain multiple words. Examples of these include phrasal verbs such as "put up with", or polywords such as "ice cream", and "run-through".
- ***Sense***. A meaning or interpretation of an expression. By *interpretation* we intend a legitimate reading of an expression that is not necessarily its definition. For example, two interpretations of "parrot" are (1) "*a species of tropical birds*", (2) "*a play on word on 'carrot'*" (this interpretation can be reasonable in certain contexts because the two words are similar). Conversely, the sense "*a colorful dress*" would not be considered a valid interpretation.
- ***Support***. In this evaluation, we discuss the concept of support in relation to expressions and their senses. Polysemous expressions can only be disambiguated if the context supports the intended meaning. We say that a pun is "supported" if some contextual clues in the sentence facilitate the perception of the two senses that create the pun effect. For instance, consider the previous pun "*The plot to his story of the pond was quite shallow*". What would happen if we replace "pond" with "mountain"?

    *The plot to his story of the **mountain** was quite shallow.*

    Now, the sense "*lacking physical depth*" cannot be perceived anymore, as the context does not support it. In the original text, "pond" evokes this specific meaning, but words related to "water", such as "lake" or "river," could also work.
- ***Relationship between words***. You will be asked to identify relationships in terms of *orthography* (spelling) or *phonetics* (sound).
    For example, in the previous heterographic pun, the words "sheepishly" and "sheep" share the same root, indicating an *orthographic and phonetic similarity*. Additionally, the words "sheep" and "ship" can also be considered similar in both regards.
    In homographic examples, you may be asked to rate the relationship/similarity between the same word (*e.g.*, between "shallow" and "shallow"). In such cases, the similarity is trivially present, as the two words are identical.

## Instructions

Each page of this survey will present a short text, which was not intended to contain a pun. You should answer the 7-9 questions related to the text, following these guidelines:

- Most questions require a **Yes** or **No** answer, but some will ask you to provide a rating on a **scale**;
- If you're unsure, you can choose ***Maybe*** if the answer depends on additional information;
- If you do not understand (some parts of) the text or the question, select ***I don't understand***;
- Some questions can be skipped depending on your answers to previous questions. These questions will be clearly marked, and we ask you to **ONLY SKIP** them **if the condition is met**.
However, you can always choose to answer these questions if you wish; skipping them is only intended to help speed up the process;
- We expect you to spend around 2 minutes per page.

Keep in mind that:

- You will find the original text repeated under most questions, allowing you to quickly access it if needed;
- The 7-9 questions are always presented in the same order, which should help you speed up the answering process a little bit;
- You are required to sign in with your Google account in order to save your progress. This way, you can stop the annotation and continue it later, provided that you are **connected to the Internet** (check before closing your browser!). Google will save your progress for up to 30 days;
- You can reach out to [ANONYMOUS] for any question or clarification.

Thank you for your participation!

# Questions with Examples

Here is the list of questions asked for each short text, along with some examples of possible answers.
To help clarify the task, some answers include justifications for the responses in brackets [], but keep in mind that some answers might be subjective.
Please **read these examples carefully**.

---

The first two questions (A, B) are meant to determine if any expression ($w_p'$/$w_a'$) is associated with a legitimate interpretation, <u>regardless</u> of whether each expression is part of a pun.

- **A.** Is $s_p'$ a possible interpretation of $w_p'$ (irrespective of it being a pun or not)?
- **B.** Is $s_a'$ a possible interpretation of $w_a'$ (irrespective of it being a pun or not)?

Examples:
- Is "a species of tropical birds" a possible interpretation of "parrot" (irrespective of it being a pun or not)?
  - Answer: **Yes**
- Is "one who imitates the words or actions of another, especially without understanding them" a possible interpretation of "parrot" (irrespective of it being a pun or not)?
  - Answer: **Yes** [this is a possible figurative interpretation of the word]
- Is "a colorful dress" a possible interpretation of "parrot" (irrespective of it being a pun or not)?
  - Answer: **No**

---

Question C asks to rate how different two senses are. This should be answered <u>regardless</u> of the presence of the two senses in the original text and their correct usage in the pun.

- **C.** **[SKIP THIS QUESTION IF YOU ANSWERED "NO" TO QUESTION A OR B]**
  Consider the word-sense pairs "$w_p'$ —— $s_p'$" and "$w_a'$ —— $s_a'$".
  Rate how **different** the two **senses** are on a scale from 1 (overlapping or very similar) to 5 (very different).

Examples:
- Consider the word-sense pairs "parrot —— a species of tropical birds" and "parrot —— one who imitates the words or actions of another ". Rate how

different the two senses are on a scale from 1 (overlapping or very similar) to 5 (very different).
- ○ Answer: **5** [totally different senses, one literal, one figurative]
● Consider the word-sense pairs "parrot —— a species of tropical birds" and "parrot —— a colorful animal primarily found in subtropical forests". Rate how different the two senses are on a scale from 1 (overlapping or very similar) to 5 (very different).
- ○ Answer: **1** [this is a bit subjective, but birds are animals, so these senses refer to the same thing with different words]

---

The next two questions (D, E) ask to <u>contextualize an expression</u> in the original text and decide if its sense is supported or not, <u>irrespective</u> of the expression being part of a pun.

**D.** **[SKIP THIS QUESTION IF YOU ANSWERED "NO" TO QUESTION A]**

Is there at least one contextual clue in the text (other than the expression itself) that supports the interpretation of **w$_p$'** as **s$_p$'** (irrespective of it being in a pun or not)?

**E.** **[SKIP THIS QUESTION IF YOU ANSWERED "NO" TO QUESTION B]**

Is there at least one contextual clue in the text (other than the expression itself) that supports the interpretation of **w$_a$'** as **s$_a$'** (irrespective of it being in a pun or not)?

Examples:

**TEXT**: *A parrot flies, a coconut rolls.*
● Is there at least one contextual clue in the text (other than the expression itself) that supports the interpretation of "parrot" as "a species of tropical birds"?
- ○ Answer: **Yes** [the verb "flies" suggests we are talking about a bird; "coconut" also suggests a tropical setting, which fits the definition of the bird]
● Is there at least one contextual clue in the text (other than the expression itself) that supports the interpretation of "parrot" as "one who imitates the words or actions of another, especially without understanding them"?
- ○ Answer: **No** [there is nothing in the text that hints at this human behavior]

---

The next two questions (F, G) ask if you can recognize any similarity between two expressions, <u>irrespective</u> of how the expressions are used in the text. Additionally, these questions are not shown for some examples.

**F.** Can you recognize a relationship or similarity between **w$_p$'** and **w$_a$'** in terms of *phonetics* (pronunciation)?

5

**G.** Can you recognize a relationship or similarity between $w_p'$ and $w_a'$ in terms of *orthography* (spelling)?

Examples:
    **TEXT**: *A parrot flies, a coconut rolls*.
- Can you recognize a relationship or similarity between "parrot" and "barrow" in terms of *phonetics* (pronunciation)?
  - Answer: **Yes** [however, this can be subjective]
- Can you recognize a relationship or similarity between "parrot" and "carrot" in terms of *orthography* (spelling)?
  - Answer: **Yes**

---

Finally, the last questions (H, I) ask you to decide if a pun is present in the text or not.

**H.** Do you think the text contains a pun involving $w_p'$ and $w_a'$?
If you think no pun is present, answer 0. If you think there is a pun, rate the quality of the pun from 1 (very poor) to 5 (very effective). If you don't know, skip this question.

**I.** Do you think the text contains any puns involving words or expressions NOT previously mentioned?
If so, write down the words or expressions that produce the pun, separated by commas. Otherwise, skip this question.

Examples:
    **TEXT**: *Time flies like an arrow, fruit lies like a banana*.
- Do you think the text contains a pun involving "lies" and "flies"?
  If you think no pun is present, answer 0. If you think there is a pun, rate the quality of the pun from 1 (very poor) to 5 (very effective). If you don't know, skip this question.
  - Answer: **0** [the word "lies" does not make any sense in the sentence, so the pun does not work]

    **TEXT**: *What do you call a movie about a parrot mocking a topic? A parrot-y*.
- Do you think the text contains a pun involving "parrot-y" and "parody"?
  If you think no pun is present, answer 0. If you think there is a pun, rate the quality of the pun from 1 (very poor) to 5 (very effective). If you don't know, skip this question.
  - Answer: **5** ["pardod-y" and "parroty" both make sense in the sentence, so there is a good pun]



\end{document}